\documentclass[letterpaper]{article}

\usepackage{aaai2027}
\nocopyright

\usepackage[hyphens]{url}
\usepackage{graphicx}
\usepackage{amsmath}
\usepackage{amssymb}
\usepackage{natbib}
\usepackage{caption}
\usepackage{algorithm}
\usepackage{algorithmic}
\usepackage{newfloat}
\usepackage{listings}
\usepackage{booktabs}

\DeclareCaptionStyle{ruled}{
  labelfont=normalfont,
  labelsep=colon,
  strut=off
}

\floatstyle{ruled}
\newfloat{listing}{tb}{lst}{}
\floatname{listing}{Listing}

\title{
FPSGen: Flexible Point Cloud Scene Generation with
BEV-Supported Transport Flows
}

\author{
Wenzhe He\textsuperscript{\rm 1},
Meng Wang\textsuperscript{\rm 1},
JiaWei Qian\textsuperscript{\rm 1},
Jinfeng Xu\textsuperscript{\rm 1},
Ying Liu\textsuperscript{\rm 2},
Ruihui Li\textsuperscript{\rm 1}
}

\affiliations{
\textsuperscript{\rm 1}
College of Computer Science and Electronic Engineering,
Hunan University, Hunan, China\\
\textsuperscript{\rm 2}
College of Information Science and Engineering,
Hunan Normal University, Hunan, China\\
hewenzhe@hnu.edu.cn,
willem@hnu.edu.cn,
qjw18341697566@hnu.edu.cn,\\
jinfengxu@hnu.edu.cn,
liu\_ying@hunnu.edu.cn,
liruihui@hnu.edu.cn
}

\begin{document}

\maketitle

\begin{abstract}
Existing point-based generative methods for outdoor scenes primarily focus on
LiDAR-conditioned completion. During training, noisy point clouds are
constructed by perturbing complete ground-truth scenes, whereas during
inference, they are initialized by adding noise to duplicated partial scans.
This train-inference mismatch inherits the sparsity and visibility bias of
partial scans, leading to sparse distant regions and incomplete geometry in
occluded areas. Moreover, the reliance on partial scans restricts generation
when LiDAR observations are unavailable or replaced by layout cues. We present
FPSGen, a flexible framework that constructs point sources independently of
partial scans. FPSGen first predicts a bird's-eye-view (BEV) prior with density,
height, and mask channels from the active cues. The density map is then sampled
to form a BEV-supported point source, enabling both unconditional and
conditioned initialization. A teacher-student approximate optimal transport
scheme then uses teacher-predicted endpoints to learn a velocity field that
induces straighter transport paths. By integrating BEV point source construction
with path-straightening transport, FPSGen provides a unified framework for
unconditional and flexible cue-conditioned scene generation. Extensive
experiments show that FPSGen achieves state-of-the-art JSD and voxel IoU
performance on SemanticKITTI completion while maintaining strong performance
with a single point transport step. On KITTI-360 unconditional generation, it
also achieves the best Coverage (COV) among the compared methods.
\end{abstract}

\section{Introduction}

\begin{figure}[t]
\centering
\includegraphics[width=\linewidth]{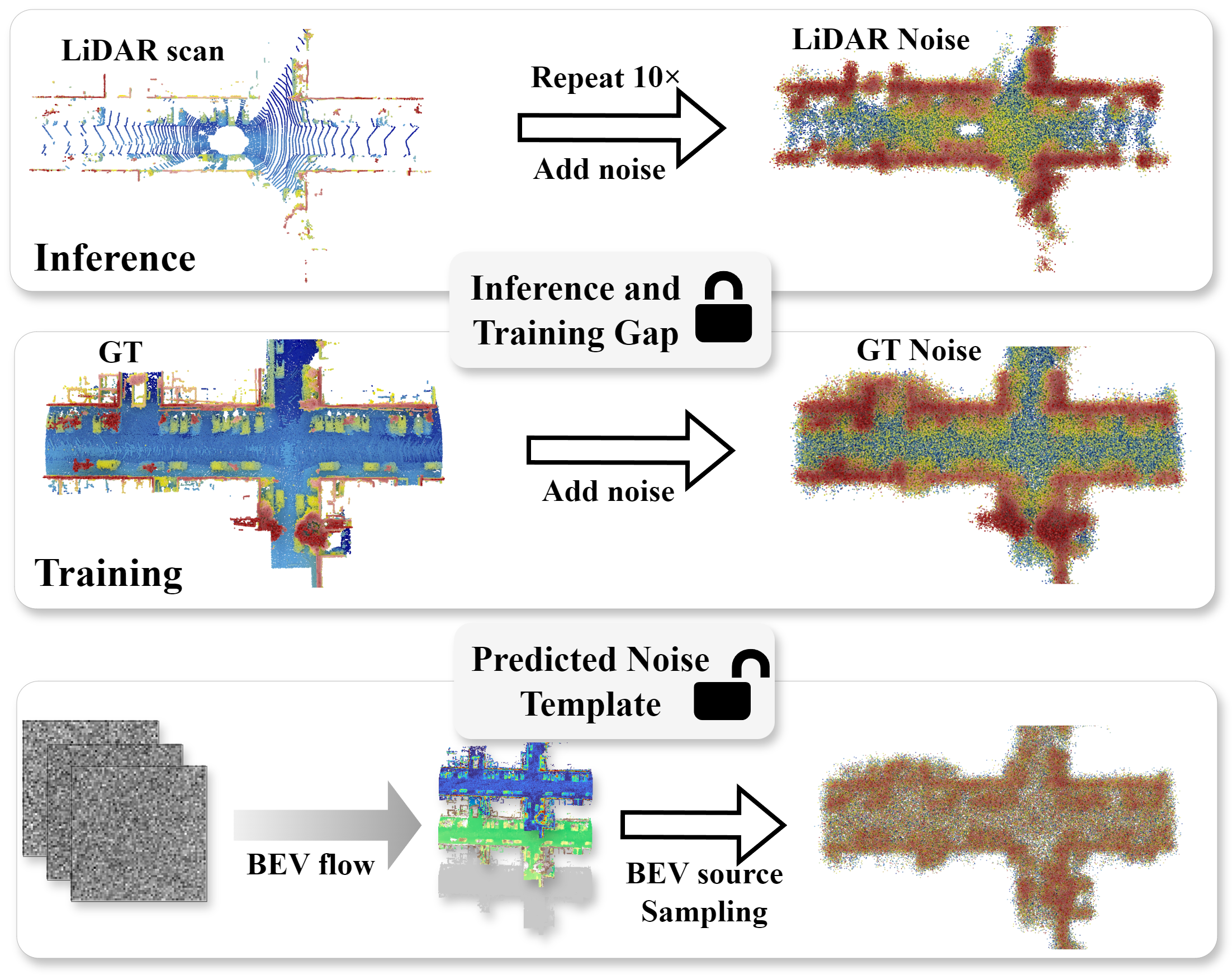}
\caption{Partial scan source mismatch in point diffusion methods for completion and the FPSGen solution. Existing models often train from perturbed complete scenes but infer from duplicated partial LiDAR scans. FPSGen instead applies the same bird's eye view (BEV) source sampler to ground truth priors during training and generated priors during inference.}
\label{fig:teaser}
\end{figure}

Large-scale 3D scene generation is important for autonomous driving simulation, data augmentation, and rare-scene analysis. Existing methods for scene-level generation commonly rely on structured grid-based representations \cite{liu2024PDD,lee2024semcity,nunes2026generatingrealistic3dsemantic}. Although these representations facilitate structured scene modeling, spatial discretization may lose geometric details and incur substantial memory costs at high resolutions. Point clouds instead preserve continuous 3D coordinates and naturally exploit the spatial sparsity of outdoor scenes. However, generating large-scale point cloud scenes remains challenging because outdoor environments span large spatial extents and contain substantially more points than individual objects. Consequently, most point-based generative methods \cite{Wu_2023_CVPR,meng2025pointnspautoregressive3dpoint,zhu2025sealion} focus on single-object generation, while scene-level point cloud generation remains relatively underexplored.

Existing point-based generative methods for outdoor LiDAR scenes primarily focus on LiDAR-conditioned completion. Given a partial scan, methods such as LiDiff \cite{lidiff}, LiDPM \cite{martyniuk2025lidpm}, and ScoreLiDAR \cite{scorelidar} recover complete scenes through iterative point denoising. These approaches are based on point diffusion and completion methods that generate or reconstruct point sets through denoising and geometric refinement \cite{luo2021diffusion,pvd,zeng2022lionlatentpointdiffusion,Lyu2021ACP,li2024diffpmae}. Although LiDPM further discusses unconditional generation, it still requires a predefined noise template for point initialization during the generation process.

As illustrated in Fig.~\ref{fig:teaser}, existing completion methods construct noisy point clouds by perturbing complete ground-truth scenes during training, whereas inference initializes them by adding noise to duplicated partial scans \cite{lidiff,martyniuk2025lidpm,scorelidar}. This train--inference mismatch is further compounded by the sparsity and visibility bias inherited from partial scans, resulting in sparse distant regions and incomplete geometry in occluded areas. Moreover, dependence on partial scans limits generation when LiDAR observations are unavailable or replaced by layout cues. Although a generic Gaussian source avoids such dependence, it provides little scene-specific spatial support over large outdoor environments. The central challenge is therefore to construct an informative point source that covers plausible scene geometry while supporting both unconditional and flexible cue-conditioned generation.

We present FPSGen, a flexible framework that constructs point sources independently of partial scans. FPSGen first predicts a bird's-eye-view (BEV) prior containing density, height, and mask channels from the active cues, and samples the density map to form a BEV-supported point source for both unconditional and conditioned initialization. The complete BEV prior further provides structured scene context for subsequent point transport. However, an informative source alone does not guarantee efficient transport, as independent source--target pairing may induce complex trajectories requiring many integration steps. FPSGen therefore introduces a teacher-student approximate optimal transport scheme, in which the teacher predicts source-indexed clean endpoints that guide the student to learn a velocity field with straighter transport paths. This design avoids explicitly solving a scene-scale optimal transport plan while enabling accurate, high-quality generation with few point transport steps.

By integrating BEV source construction with path-straightening transport, FPSGen provides a unified framework for unconditional and flexible cue-conditioned scene generation. The main contributions are as follows:
\begin{itemize}
\item We introduce a unified framework for unconditional and flexible cue-conditioned point cloud scene generation, covering road, vehicle, LiDAR, and mixed-cue inputs.
\item We propose a BEV prior generator and density-weighted sampler that construct point sources without duplicating partial scans.
\item We develop a teacher-student approximate optimal transport scheme that promotes straighter transport paths without explicit scene-scale matching.
\item Extensive experiments on SemanticKITTI and KITTI-360 demonstrate superior performance across completion and unconditional generation, together with strong accuracy under single-step point transport.
\end{itemize}

\section{Related Work}

\textbf{Scene completion.}
Scene completion reconstructs unobserved geometry from partial measurements. Structured approaches predict occupancy-based or Gaussian scene representations from depth, LiDAR, or camera inputs \cite{song2017semantic,li2023voxformer,huang2024gaussian,huang2024probabilisticgaussiansuperpositionefficient}. Point-based methods instead recover unordered point sets through coarse-to-fine decoding, transformer-based reconstruction \cite{yuan2018pcn,yu2021pointr,xiang2022snowflake,26_aaai_zhou}. At the scale of outdoor LiDAR scenes, LiDiff and LiDPM apply point diffusion to LiDAR scene completion \cite{lidiff,martyniuk2025lidpm}. ScoreLiDAR and Distillation-DPO further accelerate diffusion-based completion through distillation and preference-aligned optimization \cite{scorelidar,zhao2026distillationdpo}. These methods remain observation-dependent because a partial LiDAR scan is required to define the condition and construct the initial point set.

\textbf{Scene generation.}
Large-scale 3D scene generation aims to synthesize coherent geometry and semantics beyond isolated objects. SemCity models outdoor semantic scenes in triplane space \cite{lee2024semcity}, while Pyramid Diffusion generates fine-grained scenes through a coarse-to-fine hierarchy of discrete diffusion models \cite{liu2024PDD}. UniScene adopts semantic occupancy as a unified representation for driving scene generation \cite{li2025uniscene}, and its scaled-up extension further captures the spatial expansion and temporal evolution of 4D scenes \cite{li2025scaling}. For unbounded world generation, WorldGrow extends structured latent blocks through context-aware inpainting \cite{li2026worldgrow}, whereas WorldFlow3D transports volumetric distributions from coarse structure to detailed geometry and appearance \cite{joshi2026worldflow3dflowing3ddistributions}. These methods primarily operate on triplane, occupancy, or latent volumetric representations. In point space, LiDPM explores unconditional scene generation but still relies on a predefined noise template for initialization \cite{martyniuk2025lidpm}. Direct generation of large-scale unordered point sets under flexible input cues therefore remains underexplored.

\textbf{Flow matching.}
Flow matching trains continuous normalizing flows by regressing time-dependent vector fields along prescribed probability paths \cite{lipman2022flowmatching}. Rectified flow and optimal transport conditional flow matching improve source-target coupling and encourage straighter transport paths \cite{liu2022rectifiedflow,tong2023otcfm}. LiFlow introduces flow matching for point-based LiDAR scene completion and constructs consistent training and inference sources from the observed scan \cite{matteazzi2026liflow}. However, its source point set still depends on a LiDAR observation, leaving observation-independent source construction for layout-conditioned and unconditional point generation unresolved.

\section{Method}

\begin{table}[t]
    \centering
    \small
    \begin{tabular}{lp{0.73\linewidth}}
    \toprule
    Symbol & Meaning \\
    \midrule
    \(\mathcal{P}^{l},\mathcal{P}^{gt}\) & Sparse LiDAR condition and complete training point cloud \\
    \(C_m\) & Active condition tuple after masking LiDAR, vehicle, and road cues \\
    \(B_0,B_1\) & Initial Gaussian noise and ground-truth BEV target, respectively \\
    \(\bar{B},\hat{B}\) & Ground-truth BEV prior during training and generated BEV prior \\
    \(\mathcal{R}\) & BEV source sampler \\
    \(\mathcal{P}_{0},\mathcal{P}_{0}^{\mathrm{init}}\) & BEV-supported point source during training and inference \\
    \(\mathcal{P}^{\dagger}_{1},\hat{\mathcal{P}}\) & teacher-estimated source-indexed clean endpoint and inferred point cloud \\
    \(v_\phi,v_\psi\) & BEV velocity field and point velocity field \\
    \(\tau,t\) & BEV-flow time and point-flow time \\
    \(K_B,K_P\) & Numbers of BEV-flow and point-flow Euler steps \\
    \bottomrule
    \end{tabular}
    \caption{Main notation used in the method.}
    \label{tab:notation}
\end{table}

\subsection{Problem Formulation and Overview}

FPSGen decomposes 3D scene generation into BEV support construction and point transport. The BEV stage predicts the spatial distribution of scene points from the active cues, while the point stage transports samples from this support toward complete scene geometry. This formulation replaces duplicated partial-scan initialization with a flexible BEV-supported point source.

We adopt conditional flow matching as the shared framework for both stages \cite{lipman2022flowmatching}. Given paired source and target endpoints \(z_0\) and \(z_1\) under condition \(c\), we define the linear interpolation path and target velocity as
\begin{equation}
z_t=(1-t)z_0+t z_1,\qquad
u_t=z_1-z_0,
\label{eq:fm_prelim_path}
\end{equation}
where \(t\sim\mathcal{U}(0,1)\), with \(\mathcal{U}\) denoting the uniform distribution. A conditional velocity network \(v_\theta(z_t,t,c)\) is trained to approximate the target velocity through
\begin{equation}
\mathcal{L}_{\mathrm{CFM}}
=
\mathbb{E}_{z_0,z_1,t,c}
\left[
\left\|
v_\theta(z_t,t,c)-u_t
\right\|_2^2
\right].
\label{eq:fm_prelim_loss}
\end{equation}
At inference, samples are transported from \(t=0\) to \(t=1\) by integrating the learned ordinary differential equation:
\begin{equation}
\frac{d z_t}{dt}=v_\theta(z_t,t,c),
\label{eq:fm_ode}
\end{equation}
which is numerically solved using Euler updates.

This formulation is instantiated for both BEV generation and point transport in FPSGen, where \(z\) denotes either the BEV prior \(B\) or the point cloud \(\mathcal{P}\). The main notation used throughout the method is summarized in Table~\ref{tab:notation}. For BEV flow training, random noise is paired with the ground-truth BEV prior, while inference transports noise toward a generated BEV prior. For point flow training, a BEV-supported point source is paired with a source-indexed clean endpoint predicted by the teacher, and the student learns the corresponding velocity field. At inference, a point source sampled from the generated BEV prior is transported toward the final scene.

Let \(\mathcal{P}^{l}=\{p_i^l\}_{i=1}^{N_l}\) denote an optional sparse LiDAR observation, and let \(\mathcal{P}^{gt}=\{p_i^{gt}\}_{i=1}^{N}\) denote the complete target point cloud available during training. We encode \(\mathcal{P}^{l}\) as the LiDAR condition \(c_l\), while the other optional cues are the vehicle mask \(c_v\) and road mask \(c_r\). Binary variables \(m_l,m_v,m_r\in\{0,1\}\) indicate whether each cue is active. The active conditions are collected as
\begin{equation}
C_m=(m_l c_l,\;m_v c_v,\;m_r c_r).
\label{eq:condition_tuple}
\end{equation}
During training, both the BEV flow and point flow uniformly sample all eight condition combinations. Inactive cues are replaced with zero tensors before being fed into the networks.

The overall objective is to learn a conditional generator
\begin{equation}
\mathcal{P}\sim p_\theta(\mathcal{P}\mid C_m),
\label{eq:conditional_generation}
\end{equation}
such that completion, layout-conditioned generation, mixed-condition generation, and unconditional generation are handled by a single framework. FPSGen realizes this distribution through a two-stage sampling process:
\begin{equation}
\begin{aligned}
& \hat{B}\sim p_\phi(\hat{B}\mid C_m),\\
& \mathcal{P}_{0}^{\mathrm{init}}=\mathcal{R}(\hat{B};N,\Sigma),\\
& \mathcal{P}\sim p_\psi\!\left(
\mathcal{P}\mid
\mathcal{P}_{0}^{\mathrm{init}},\hat{B},C_m
\right).
\end{aligned}
\label{eq:factorization}
\end{equation}
Here, \(p_\phi\) generates the BEV prior \(\hat{B}\), \(\mathcal{R}\) samples a BEV-supported point source from its density channel, and \(p_\psi\) performs point-level transport. The source contains \(N\) points, and \(\Sigma\) denotes the covariance of the coordinate perturbation. After marginalizing the stochastic BEV prior and source sampling, these components jointly define \(p_\theta(\mathcal{P}\mid C_m)\). We refer to \(\mathcal{R}\) as the BEV source sampler. 

\begin{figure*}[t]
    \centering
    \includegraphics[width=\linewidth]{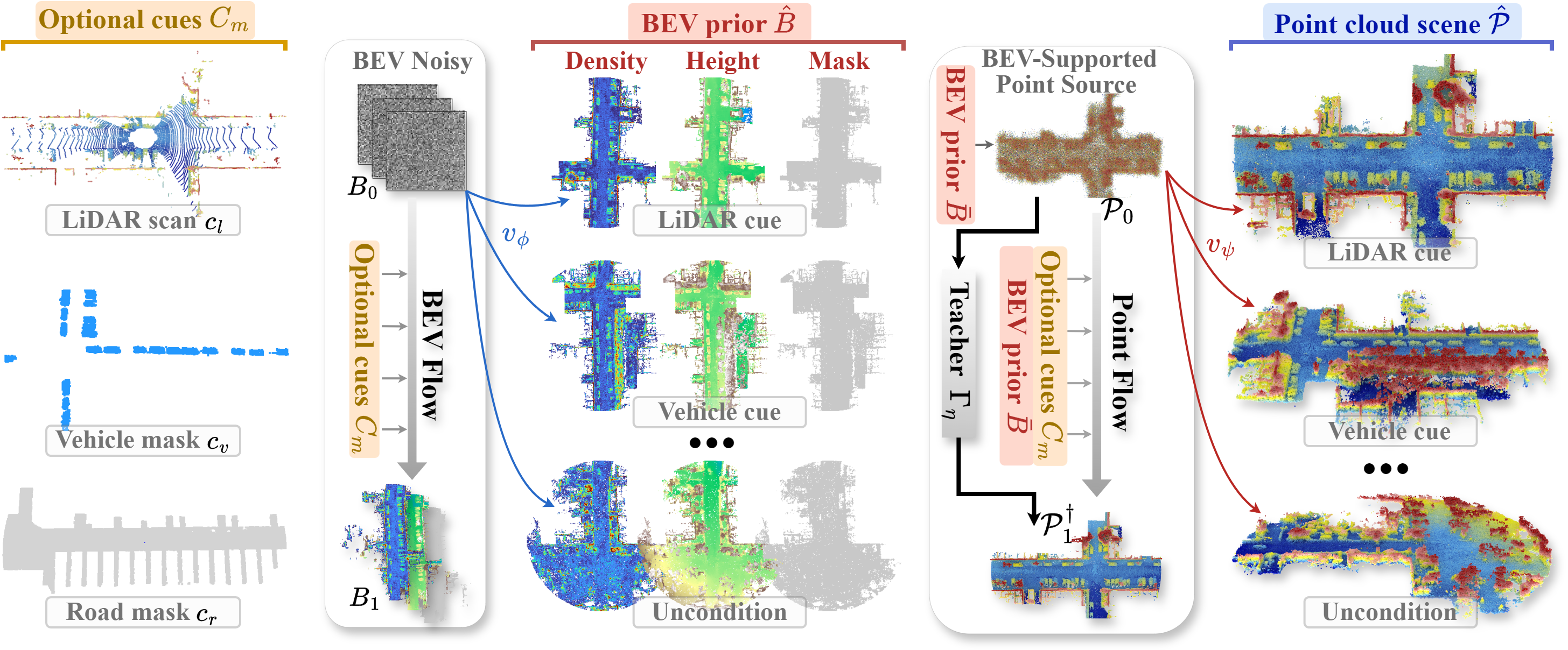}
    \caption{FPSGen training and inference pipeline. Optional cues \(C_m\), including LiDAR, vehicle, and road conditions, drive the BEV flow from noise \(B_0\) to a density, height, and mask prior. The BEV source sampler converts the density channel into a BEV-supported point source \(\mathcal{P}_0\), while the complete BEV prior is provided to the point flow as context. During training, the ground-truth prior \(\bar{B}\) supplies source support, and the teacher estimates the clean endpoint \(\mathcal{P}^{\dagger}_{1}\). During inference, the generated prior \(\hat{B}\) and optional cues guide the student point flow to produce the final scene \(\hat{\mathcal{P}}\).}
    \label{fig:overview}
\end{figure*}

Figure~\ref{fig:overview} summarizes the complete pipeline. During training, the complete point cloud is converted into the ground-truth BEV prior \(\bar{B}\). The BEV source sampler constructs the corresponding point source \(\mathcal{P}_{0}\), and the teacher maps it to a source-indexed clean endpoint \(\mathcal{P}^{\dagger}_{1}\), which supervises the student point flow. During inference, the BEV flow first generates \(\hat{B}\), after which \(\mathcal{R}\) constructs \(\mathcal{P}_{0}^{\mathrm{init}}\), and the student integrates the point velocity field from \(t=0\) to \(t=1\). The same BEV source sampler is used during training and inference, although inference quality depends on the generated prior \(\hat{B}\). Detailed architectures, layout-mask construction, and training schedules are provided in the supplementary material.

\subsection{Flexible Condition BEV Flow Prior}

The BEV prior fuses the LiDAR condition, vehicle mask, and road mask in a common spatial grid. It supplies coarse scene support before unordered points are instantiated. We define a projection operator \(\Phi\) to map the complete ground-truth point cloud \(\mathcal{P}^{gt}\) into a structured BEV target
\begin{equation}
B_1=\Phi(\mathcal{P}^{gt})=[D,H,M]\in \mathbb{R}^{3\times h\times w},
\label{eq:bev_target}
\end{equation}
where \(D\), \(H\), and \(M\) denote normalized density, maximum height, and occupancy mask channels. We use log density, maximum height, and binary occupancy normalization to map the BEV target to the \([-1,1]\) range. The exact construction is given in the supplementary material. The BEV flow uses Gaussian noise \(B_0\sim\mathcal{N}(0,I)\) and the linear path
\begin{equation}
B_\tau=(1-\tau)B_0+\tau B_1,\; u_B=B_1-B_0.
\label{eq:bev_path}
\end{equation}

During training, we sample an active condition tuple \(C_m\) with the same uniform condition mask protocol. The BEV velocity network predicts
\begin{equation}
\hat{u}_B=v_\phi(B_\tau,\tau,C_m),
\label{eq:bev_velocity}
\end{equation}
and is trained with
\begin{equation}
\mathcal{L}_{\mathrm{BEV}}=
\mathrm{E}_{B_0,B_1,\tau,C_m}
\left[\left\|\hat{u}_B-u_B\right\|_2^2\right].
\label{eq:bev_loss}
\end{equation}
Both training-time condition dropout and inference-time available cues are represented by the same tuple \(C_m\). 

To convert a BEV prior into a point source, we define the BEV source sampler \(\mathcal{R}\) using the density channel. Since \(D\) is stored in the normalized BEV scale, we first convert it back to a nonnegative density score, where \(n_{\max}\) is the density clipping constant.
\begin{equation}
\rho_D(q)=
\exp\!\left(\frac{D(q)+1}{2}\log(1+n_{\max})\right)-1 .
\label{eq:density_denorm}
\end{equation}
Given \(B=[D,H,M]\), the BEV cell sampling weight is
\begin{equation}
w_D(q)=
\frac{\max(\rho_D(q),0)+\varepsilon_w}
{\sum_{q'}(\max(\rho_D(q'),0)+\varepsilon_w)}.
\label{eq:density_weight}
\end{equation}
Here \(\varepsilon_w>0\) is a small numerical stabilizer. We independently sample \(N=180{,}000\) cells with replacement according
to the weights \(w_D\), where \(\mathrm{Cat}(w_D)\) denotes the categorical
distribution over all BEV cells parameterized by \(w_D\). Each sampled cell
is mapped to a zero-height metric point and perturbed with Gaussian
coordinate noise.
\begin{equation}
\begin{aligned}
& q_i\sim\mathrm{Cat}(w_D),\;
p_i^b=(x(q_i),y(q_i),0),\\
& \mathcal{R}(B;N,\Sigma)=
\{p_i^b+\epsilon_i\}_{i=1}^{N},
\;
\epsilon_i\sim\mathcal{N}(0,\Sigma).
\end{aligned}
\label{eq:source_operator}
\end{equation}
In our implementation, the BEV grid covers
\([-50,50]\,\mathrm{m}\times[-50,50]\,\mathrm{m}\) with \(256\times256\) cells. The sampled cell centers \(x(q_i)\) and \(y(q_i)\) are converted to metric coordinates when constructing \(\mathcal{R}\).
The resulting metric point source is subsequently processed by the point flow, which is conditioned on the complete BEV prior \(B=[D,H,M]\) with channel values normalized to \([-1,1]\). Thus, \(D\) determines the spatial distribution of the source points, while all three channels provide structured BEV context to the point flow.

\subsection{Teacher Transport Mapping}

Once BEV-supported point source is available, the point stage must map noisy point source to a detailed scene. The BEV-supported point source follows the global road and object layout, but it does not provide reliable point-wise correspondences to \(\mathcal{P}^{gt}\). Direct optimal transport (OT) matching between the noisy point source and \(\mathcal{P}^{gt}\) is computationally expensive at scene scale. We therefore use a learned teacher to approximate the source-to-target matching.

Let \(\bar{B}=[\bar{D},\bar{H},\bar{M}]\) denote the ground-truth BEV prior used during point-stage training. In our implementation, \(\bar{B}=B_1=\Phi(\mathcal{P}^{gt})\). Applying the BEV source sampler in Eq.~(\ref{eq:source_operator}) to this training-time prior yields the noisy source endpoint
\begin{equation}
\mathcal{P}_{0}
=\mathcal{R}(\bar{B};N,\Sigma).
\label{eq:teacher_source}
\end{equation}

Given the noisy source \(\mathcal{P}_{0}\) and the complete target \(\mathcal{P}^{gt}\), the teacher predicts a pointwise displacement toward a clean endpoint:
\begin{equation}
\mathcal{P}^{\dagger}_{1}
=\mathcal{P}_{0}+\Gamma_{\eta}(\mathcal{P}_{0},\mathcal{P}^{gt}),
\label{eq:teacher_mapping}
\end{equation}
where \(\Gamma_{\eta}\) is the teacher network and \(\mathcal{P}^{\dagger}_{1}\) is the teacher-estimated clean endpoint. Since the source point order is preserved by the residual mapping, \(\mathcal{P}^{\dagger}_{1}\) provides a source-indexed clean endpoint for student training.

The teacher is trained using set-level geometric supervision:
\begin{equation}
\mathcal{L}_{T}
=
\mathbb{E}_{\mathcal{P}^{gt},\mathcal{P}_{0}}
\left[
\mathrm{CD}\left(\mathcal{P}^{\dagger}_{1},\mathcal{P}^{gt}\right)
+\lambda_{\mathrm{rep}}\mathcal{L}_{\mathrm{rep}}
\right].
\label{eq:teacher_loss}
\end{equation}
Here, the Chamfer distance aligns the predicted endpoint with the complete scene, while the repulsion term discourages locally collapsed point clusters:
\begin{equation}
\mathcal{L}_{\mathrm{rep}}
=
\frac{1}{N}\sum_{i=1}^{N}
\max\!\left(
0,\,
r_{\mathrm{rep}}
-
\left\|p_i^{\dagger}-p_{\nu(i)}^{\dagger}\right\|_2
\right),
\label{eq:teacher_repulsion}
\end{equation}
where \(p_i^{\dagger}\in\mathcal{P}_{1}^{\dagger}\), \(\nu(i)\) denotes the nearest-neighbor index of \(p_i^{\dagger}\) in \(\mathcal{P}_{1}^{\dagger}\setminus\{p_i^{\dagger}\}\), and \(r_{\mathrm{rep}}\) is the repulsion radius.

We refer to \((\mathcal{P}_{0},\mathcal{P}^{\dagger}_{1})\) as the teacher pair. The student is trained on the BEV-supported point source and its teacher-estimated, source-indexed clean endpoint, avoiding an explicit scene-scale OT plan.

\subsection{Approximate-OT Point Flow}

The teacher-estimated mapping provides a scalable target for point transport. The source-to-target displacement used to train the student is
\begin{equation}
u_P
=
\mathcal{P}^{\dagger}_{1}-\mathcal{P}_{0}
=
\Gamma_{\eta}(\mathcal{P}_{0},\mathcal{P}^{gt}).
\label{eq:student_target_velocity}
\end{equation}
We refer to this construction as Approx OT because the teacher provides an amortized, source-indexed clean endpoint for the BEV-supported point source without explicitly solving optimal transport or superset optimal transport at the \(180{,}000\)-point scene scale. Approx OT therefore denotes a learned surrogate for scene-scale matching rather than an exact optimal transport solution.

For each training scene, the teacher pair \((\mathcal{P}_{0},\mathcal{P}^{\dagger}_{1})\) defines the endpoints of the point transport path. We treat \(\mathcal{P}_{0}\) as the source endpoint at \(t=0\) and \(\mathcal{P}^{\dagger}_{1}\) as the clean endpoint at \(t=1\). We sample \(t\sim\mathcal{U}(0,1)\) and construct
\begin{equation}
\mathcal{P}_t
=
(1-t)\mathcal{P}_{0}
+
t\mathcal{P}^{\dagger}_{1},
\qquad
u_P
=
\mathcal{P}^{\dagger}_{1}
-
\mathcal{P}_{0}.
\label{eq:point_path}
\end{equation}
Given \(\mathcal{P}_t\), \(\bar{B}\), and \(C_m\), the student predicts
\begin{equation}
\hat{u}_P
=
v_\psi(\mathcal{P}_t,t,\bar{B},C_m).
\label{eq:student_velocity}
\end{equation}
Using the same condition-tuple sampling strategy as in the BEV stage, the point transport objective is
\begin{equation}
\mathcal{L}_{\mathrm{Point}}
=
\mathbb{E}_{\mathcal{P}^{gt},\mathcal{P}_{0},t,C_m}
\left[
\left\|
\hat{u}_P-u_P
\right\|_2^2
\right].
\label{eq:point_loss}
\end{equation}
This objective aligns the student velocity with the source-to-target displacement induced by the teacher-estimated mapping. The same condition tuple \(C_m\) is used by both the BEV prior model and the point transport model.

\subsection{Unified Inference}

Inference uses the same BEV-to-point structure as training, but removes all inputs that require complete scene supervision. The teacher is discarded, and the ground-truth BEV prior \(\bar{B}\) is replaced by a generated BEV prior \(\hat{B}\). We use \(\tau\) for BEV flow time and \(t\) for point transport time. For compactness, let \(v_\phi^m\) and \(v_\phi^0\) denote \(v_\phi(B_\tau,\tau,C_m)\) and \(v_\phi(B_\tau,\tau,C_{\emptyset})\), and let \(v_\psi^m\) and \(v_\psi^0\) denote \(v_\psi(\mathcal{P}_t,t,\hat{B},C_m)\) and \(v_\psi(\mathcal{P}_t,t,\hat{B},C_{\emptyset})\). Classifier-free guidance is applied to both flow fields as
\begin{equation}
\begin{array}{l}
v_\phi^{\mathrm{cfg}}=v_\phi^0+s_B(v_\phi^m-v_\phi^0),\\
v_\psi^{\mathrm{cfg}}=v_\psi^0+s_P(v_\psi^m-v_\psi^0).
\end{array}
\label{eq:cfg}
\end{equation}
Here \(s_B\) and \(s_P\) are the BEV and point guidance scales. In the Euler updates below, each guided field is evaluated at the current state and time.

The BEV flow starts from Gaussian noise and evolves forward to a BEV prior.
\begin{equation}
B_{\tau_0}\sim\mathcal{N}(0,I),\;
\frac{dB_\tau}{d\tau}=v_\phi^{\mathrm{cfg}},\;
\hat{B}=B_{\tau=1}.
\label{eq:bev_inference}
\end{equation}
With \(K_B\) forward Euler steps, \(0=\tau_0<\tau_1<\cdots<\tau_{K_B}=1\),
\begin{equation}
B_{\tau_{\ell+1}}
=B_{\tau_\ell}+(\tau_{\ell+1}-\tau_\ell)v_\phi^{\mathrm{cfg}},\;
\hat{B}=B_{\tau_{K_B}}.
\label{eq:bev_euler}
\end{equation}
The generated BEV prior \(\hat{B}\) is then converted into a BEV-supported point source \(\mathcal{P}_{0}^{\mathrm{init}}\) for the point flow.
\begin{equation}
\mathcal{P}_{0}^{\mathrm{init}}=\mathcal{R}(\hat{B};N,\Sigma).
\label{eq:inference_source}
\end{equation}
Let \(0=t_0<t_1<\cdots<t_{K_P}=1\). 
Starting from \(\mathcal{P}_{0}^{\mathrm{init}}\), the point flow transports the source toward the generated scene.
\begin{equation}
\mathcal{P}_{t_0}=\mathcal{P}_{0}^{\mathrm{init}},\;
\frac{d\mathcal{P}_t}{dt}=v_\psi^{\mathrm{cfg}},\;
\hat{\mathcal{P}}=\mathcal{P}_{t_{K_P}}.
\label{eq:point_inference_ode}
\end{equation}
The point Euler updates are
\begin{equation}
\mathcal{P}_{t_{k+1}}
=\mathcal{P}_{t_k}+(t_{k+1}-t_k)
v_\psi^{\mathrm{cfg}}.
\label{eq:euler}
\end{equation}
Thus every conditioning setting follows a single forward inference chain. \(C_m\) drives the BEV flow to \(\hat{B}\), \(\mathcal{R}\) samples \(\mathcal{P}_{0}^{\mathrm{init}}\), and the student point flow produces \(\hat{\mathcal{P}}\) as the generated point cloud scene.

\section{Experiments}

\begin{table*}[t]
\centering
\begin{tabular}{lcccccc}
\toprule
Method & CD \(\downarrow\) & \(JSD_{3D}\) \(\downarrow\) & \(JSD_{BEV}\) \(\downarrow\) & IoU@\(0.5\,\mathrm{m}\) \(\uparrow\) & IoU@\(0.2\,\mathrm{m}\) \(\uparrow\) & IoU@\(0.1\,\mathrm{m}\) \(\uparrow\) \\
\midrule
LMSCNet & 0.641 & -- & 0.431 & 30.83 & 12.09 & 3.65 \\
LODE & 1.029 & -- & 0.451 & 33.81 & 16.39 & 5.00 \\
MID & 0.503 & -- & 0.470 & 31.58 & \underline{22.72} & \underline{13.14} \\
PVD & 1.256 & -- & 0.498 & 15.91 & 3.97 & 0.60 \\
LiDiff & 0.434 & 0.564 & 0.444 & 31.47 & 16.79 & 4.67 \\
LiDPM & 0.446 & \underline{0.532} & 0.440 & \underline{34.09} & 19.45 & 6.27 \\
ScoreLiDAR & \underline{0.406} & -- & 0.425 & -- & -- & -- \\
Distillation-DPO & 0.414 & -- & 0.419 & -- & -- & -- \\
LiFlow & \textbf{0.309} & -- & \underline{0.416} & 31.60 & 13.10 & 3.80 \\
FPSGen (LiDAR) & \underline{0.316} & \textbf{0.492} & \textbf{0.329} & \textbf{44.15} & \textbf{32.26} & \textbf{18.26} \\
\midrule
FPSGen (LiDAR + road) & \underline{0.291} & \underline{0.488} & \underline{0.315} & 44.43 & 32.35 & 18.38 \\
FPSGen (LiDAR + vehicle) & 0.307 & \underline{0.488} & 0.325 & \underline{44.71} & \textbf{32.77} & \textbf{18.53} \\
FPSGen (all conditions) & \textbf{0.288} & \textbf{0.485} & \textbf{0.313} & \textbf{44.87} & \underline{32.71} & \underline{18.49} \\
\bottomrule
\end{tabular}
\caption{LiDAR-conditioned completion on SemanticKITTI sequence 08. Results for other methods follow the published completion protocol.}
\label{tab:completion}
\end{table*}

\begin{table*}[t]
\centering
\setlength{\tabcolsep}{3.2pt}
\begin{tabular}{llcccccc}
\toprule
Condition & Method & COV-CD \(\uparrow\) & MMD-CD \(\downarrow\) & 1-NNA-CD & COV-EMD \(\uparrow\) & MMD-EMD \(\downarrow\) & 1-NNA-EMD \\
\midrule
Unconditional & LiDiff & 39.73 & 8.96 & 87.64 & \underline{43.54} & 5.99 & 83.75 \\
Unconditional & LiDPM & \underline{40.68} & \textbf{7.42} & \textbf{73.57} & 39.35 & \underline{5.92} & \underline{81.84} \\
Unconditional & SemCity & 7.98 & 26.86 & 99.05 & 10.84 & 8.95 & 98.00 \\
Unconditional & FPSGen & \textbf{41.25} & \underline{7.48} & \underline{75.29} & \textbf{44.30} & \textbf{5.85} & \textbf{77.09} \\
\midrule
Vehicle mask & FPSGen & 46.77 & 7.09 & 72.34 & 46.39 & 5.72 & 72.81 \\
Road mask & FPSGen & \underline{63.12} & \underline{5.59} & \textbf{51.24} & \underline{57.98} & \underline{5.26} & \underline{57.51} \\
Road + vehicle & FPSGen & \textbf{65.21} & \textbf{5.43} & \underline{47.72} & \textbf{60.27} & \textbf{5.15} & \textbf{56.08} \\
\bottomrule
\end{tabular}
\caption{Distributional generation quality on KITTI-360. COV and 1-NNA are percentages, and 1-NNA is best when closer to \(50\%\).}
\label{tab:generation_kitti360}
\end{table*}

\subsection{Tasks, Datasets, and Metrics}

We evaluate FPSGen on LiDAR-conditioned completion and flexible conditional scene generation. LiDAR is treated as a strong geometric condition for completion, while unconditional, road mask, and vehicle mask settings test generation from weak or absent conditions.

All FPSGen components are trained once on SemanticKITTI sequences 00--07 and 09--10 \cite{semantickitti}. We evaluate the resulting model on SemanticKITTI sequence 08 and KITTI-360 sequence 00 \cite{kitti360} without retraining. Sequence 08 provides held-out SemanticKITTI evaluation, while KITTI-360 measures transfer to a different dataset. Complete scenes are prepared from aggregated scans following prior LiDAR completion work \cite{lidiff}. Semantic labels are used only to derive road and vehicle masks.

Completion uses Chamfer distance (CD), 3D and BEV Jensen-Shannon divergence (\(JSD_{3D}\) and \(JSD_{BEV}\)), and voxel intersection over union (IoU) at \(0.5\,\mathrm{m}\), \(0.2\,\mathrm{m}\), and \(0.1\,\mathrm{m}\) \cite{chamferdistance,1997JSD}. Generation uses coverage (COV), minimum matching distance (MMD), and 1-nearest-neighbor accuracy (1-NNA) under CD and Earth Mover's Distance (EMD) \cite{achlioptas2018learning,rubner2000earth}. Detailed metric definitions are provided in the supplementary material.

\subsection{Baselines and Main Results}

We compare FPSGen with representative voxel-, point-, and diffusion-based completion methods \cite{lmscnet,li2023lode,vizzo2022makeitdense,pvd,lidiff,martyniuk2025lidpm,scorelidar,zhao2026distillationdpo,matteazzi2026liflow}, and with LiDiff, LiDPM, and SemCity for generation \cite{lidiff,martyniuk2025lidpm,semcity2024}. We also evaluate all completion methods using the same sparse scans generated by LiDM \cite{Lidardiffusioncvpr2024}.

With LiDAR alone, LiFlow achieves the lowest CD of 0.309, while FPSGen obtains a comparable CD of 0.316 and the best \(JSD_{3D}\), \(JSD_{BEV}\), and IoU results in Table~\ref{tab:completion}. As further shown in Table~\ref{tab:init_ablation}, scan-repeated initialization favors CD, whereas the BEV source improves distribution alignment and supports LiDAR-free generation. Adding road and vehicle cues further reduces CD to 0.288 and \(JSD_{BEV}\) to 0.313.
Without retraining on KITTI-360, FPSGen achieves the best unconditional EMD-based results in Table~\ref{tab:generation_kitti360}. Road and vehicle cues further raise COV-CD from 41.25 to 65.21 and bring 1-NNA closer to \(50\%\).
In the LiDM diagnostic, FPSGen raises COV-EMD from 2.85 to 30.80 and reduces MMD-EMD from 13.232 to 6.400, while Distillation-DPO retains the lowest MMD-CD.

\begin{figure*}[t]
\centering
\includegraphics[width=\linewidth]{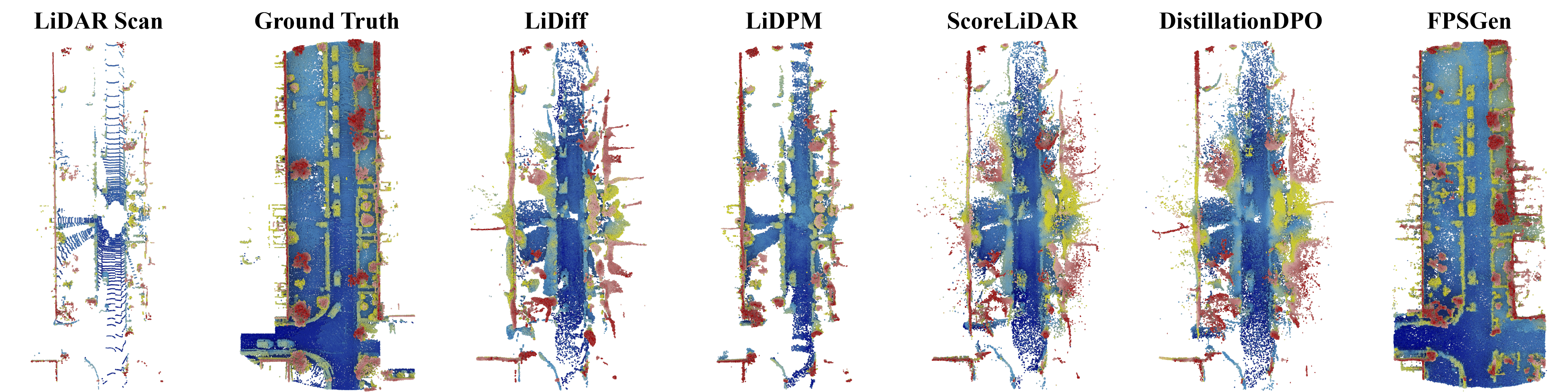}
\caption{Qualitative LiDAR-conditioned completion comparison on SemanticKITTI sequence 08. From left to right, the panels show the input LiDAR scan, ground truth, LiDiff, LiDPM, ScoreLiDAR, Distillation-DPO, and FPSGen.}
\label{fig:completion_qual_main}
\end{figure*}

\begin{table*}[t]
\centering
\setlength{\tabcolsep}{2.0pt}
\begin{tabular}{lcccccc}
\toprule
Method & COV-CD \(\uparrow\) & MMD-CD \(\downarrow\) & 1-NNA-CD & COV-EMD \(\uparrow\) & MMD-EMD \(\downarrow\) & 1-NNA-EMD \\
\midrule
LiDM + LiDiff & 23.57 & 9.816 & 95.15 & \underline{2.85} & 13.518 & \underline{99.90} \\
LiDM + LiDPM & 23.76 & 9.868 & 94.77 & \underline{2.85} & 13.616 & \underline{99.90} \\
LiDM + ScoreLiDAR & 23.38 & \underline{8.013} & 94.20 & 2.28 & 13.366 & \underline{99.90} \\
LiDM + Distillation-DPO & \underline{24.33} & \textbf{7.905} & \underline{93.54} & 2.47 & \underline{13.232} & \underline{99.90} \\
LiDM + FPSGen & \textbf{28.33} & 8.689 & \textbf{87.26} & \textbf{30.80} & \textbf{6.400} & \textbf{85.17} \\
\bottomrule
\end{tabular}
\caption{KITTI-360 LiDM seeded generation diagnostic. LiDM first generates sparse LiDAR scans, and each method completes them into full point clouds. COV and 1-NNA are percentages, and 1-NNA is best when closer to \(50\%\).}
\label{tab:lidm_seed_generation}
\end{table*}

\begin{table}[t]
\centering
\begin{tabular}{lccc}
\toprule
Variant & CD \(\downarrow\) & \(JSD_{3D}\) \(\downarrow\) & \(JSD_{BEV}\) \(\downarrow\) \\
\midrule
w/o density \(D\) & 0.3821 & 0.5653 & 0.3451 \\
w/o height \(H\) & 0.3693 & 0.5510 & 0.3315 \\
w/o mask \(M\) & \underline{0.3163} & \underline{0.4930} & \underline{0.3302} \\
Full \(D{+}H{+}M\) & \textbf{0.3161} & \textbf{0.4924} & \textbf{0.3292} \\
\bottomrule
\end{tabular}
\caption{BEV prior component ablation.}
\label{tab:bev_ablation}
\end{table}

\begin{table}[t]
\centering
\begin{tabular}{lccc}
\toprule
Source & CD & $JSD_{\mathrm{3D}}$ & $JSD_{\mathrm{BEV}}$ \\
\midrule
BEV source sampler & \underline{0.3161} & \textbf{0.4924} & \textbf{0.3292} \\
Gaussian & 0.3337 & 0.6299 & 0.4138 \\
LiDAR-derived & \textbf{0.2562} & \underline{0.5114} & \underline{0.3449} \\
\bottomrule
\end{tabular}
\caption{Point-source initialization ablation with $K_P=32$ point-flow steps. Lower is better for all metrics.}
\label{tab:init_ablation}
\end{table}

\subsection{Ablation Study}

We ablate BEV prior components and point source initialization. Further transport and scalability studies are provided in the supplementary material.

\textbf{BEV prior components.}
Density and height are the main support variables in Table~\ref{tab:bev_ablation}. Removing them raises CD from 0.3161 to 0.3821 and 0.3693, respectively, showing their importance for accurate spatial support, whereas removing the mask has only a minor effect.

\textbf{Source initialization.}
As shown in Table~\ref{tab:init_ablation}, Gaussian initialization degrades all metrics. LiDAR-derived initialization achieves the lowest CD but yields worse JSD scores and requires a partial LiDAR scan as input. In contrast, the BEV source sampler provides better distribution alignment without scan-dependent initialization.

\subsection{Qualitative Analysis and Inference Efficiency}
Figure~\ref{fig:completion_qual_main} further illustrates the role of the source distribution. LiDiff, LiDPM, ScoreLiDAR, and Distillation-DPO construct their initial distributions by perturbing the observed LiDAR scan, so their completions tend to recover regions already supported by visible scan points while leaving heavily occluded areas underfilled. FPSGen first builds a BEV-supported point source over the scene layout, which gives the point flow broader support to transport and better matches the intended completion objective. Additional generation visualizations are included in the supplementary material.

\begin{table}[t]
\centering
\setlength{\tabcolsep}{3.0pt}
\begin{tabular}{lccccc}
\toprule
Method & \(K_B\) & \(K_P\) & CD \(\downarrow\) & Time \(\downarrow\) & Mem. \(\downarrow\) \\
\midrule
LiDiff & -- & 50 & 0.434 & 33.75 & 6.18 \\
LiDPM & -- & 20 & 0.446 & 20.92 & 5.53 \\
ScoreLiDAR & -- & 8 & 0.406 & 7.73 & 6.07 \\
Distillation-DPO & -- & 8 & 0.414 & \underline{5.81} & 6.07 \\
\midrule
FPSGen & 10 & 1 & \underline{0.321} & \textbf{0.95} & \underline{4.05} \\
FPSGen & 10 & 32 & \textbf{0.316} & 15.57 & \textbf{4.04} \\
\bottomrule
\end{tabular}
\caption{Single GPU inference runtime on SemanticKITTI sequence 08. Time is seconds per scene, and memory is peak reserved GPU memory in GB.}
\label{tab:runtime_main}
\end{table}

Table~\ref{tab:runtime_main} reports per-scene inference runtime under the same experimental environment. Here \(K_B\) and \(K_P\) denote BEV-flow and point-flow steps. With \(K_B=10\) and only one point flow step, FPSGen reaches CD 0.321 in 0.95 seconds. Increasing \(K_P\) to 32 reduces CD by 0.005 while remaining faster than LiDiff and LiDPM and using less peak memory. These results demonstrate that the proposed Approx OT construction enables fast scene generation with a single point flow step. The supplementary material provides runtime sweeps over different \(K_B\) and \(K_P\) settings.

\section{Conclusion}

We presented FPSGen, a flexible point cloud scene generator with BEV-supported transport. FPSGen replaces partial scan-dependent initialization with a staged BEV-to-point process. A BEV flow predicts density, height, and mask priors, after which the BEV source sampler constructs a supported point source and a teacher-student point flow transports it into complete scene geometry. This design improves LiDAR-conditioned completion while supporting unconditional and layout-conditioned generation within one framework. The SemanticKITTI-trained model transfers to KITTI-360 without retraining, showing that BEV support bridges structured scene layouts and unordered point clouds. FPSGen still depends on complete point cloud supervision and BEV prior quality, while focusing on single-frame scenes. Future work will explore temporal generation, richer controllable layouts, and broader sensor-aware sampling.

\section{Additional Experimental Results}
\label{sec:additional_results}

\subsection{SemanticKITTI Generation Comparison}
\label{subsec:semkitti_generation_supp}

\begin{table*}[t]
\centering
\setlength{\tabcolsep}{3.2pt}
\begin{tabular}{lcccccc}
\toprule
Method & COV-CD \(\uparrow\) & MMD-CD \(\downarrow\) & 1-NNA-CD & COV-EMD \(\uparrow\) & MMD-EMD \(\downarrow\) & 1-NNA-EMD \\
\midrule
LiDiff & 39.71 & 9.596 & 98.41 & \underline{44.36} & 6.594 & 93.26 \\
LiDPM & \textbf{42.65} & \textbf{6.476} & \underline{87.62} & 43.63 & \underline{6.194} & \underline{88.73} \\
SemCity & 14.71 & 17.795 & 98.04 & 16.91 & 8.083 & 96.45 \\
FPSGen & \underline{41.67} & \underline{6.517} & \textbf{79.04} & \textbf{45.59} & \textbf{6.132} & \textbf{79.53} \\
\bottomrule
\end{tabular}
\caption{Distributional generation quality on SemanticKITTI sequence 08 under a matched LiDAR free protocol. COV and 1-NNA are percentages, and 1-NNA is best when closer to \(50\%\).}
\label{tab:generation_semkitti_supp}
\end{table*}

Table~\ref{tab:generation_semkitti_supp} reports the in-domain SemanticKITTI \cite{semantickitti} generation comparison against LiDiff \cite{lidiff}, LiDPM \cite{martyniuk2025lidpm}, and SemCity \cite{semcity2024}. FPSGen achieves the best EMD side metrics and the 1-NNA-CD value closest to the ideal two sample level, while LiDPM remains slightly lower on MMD-CD.

\subsection{KITTI-360 Completion Comparison}
\label{subsec:kitti360_completion_supp}

Table~\ref{tab:kitti360_completion_supp} provides a KITTI-360 \cite{kitti360} LiDAR-conditioned completion comparison using the same completion metrics as the SemanticKITTI comparison. The published baselines include LMSCNet \cite{lmscnet}, LODE \cite{li2023lode}, MID \cite{vizzo2022makeitdense}, LiDiff \cite{lidiff}, ScoreLiDAR \cite{scorelidar}, and Distillation-DPO \cite{zhao2026distillationdpo}. FPSGen is evaluated under the LiDAR-only condition. For Distillation-DPO, we report the few-step student result and include only the metrics provided in its published KITTI-360 comparison.

\begin{table*}[t]
\centering
\setlength{\tabcolsep}{4.0pt}
\begin{tabular}{lcccccc}
\toprule
Method & CD \(\downarrow\) & \(JSD_{3D}\) \(\downarrow\) & \(JSD_{BEV}\) \(\downarrow\) & IoU@\(0.5\,\mathrm{m}\) \(\uparrow\) & IoU@\(0.2\,\mathrm{m}\) \(\uparrow\) & IoU@\(0.1\,\mathrm{m}\) \(\uparrow\) \\
\midrule
LMSCNet & 0.979 & -- & 0.496 & 26.17 & 9.21 & 2.88 \\
LODE & 1.565 & -- & 0.483 & 33.06 & 15.24 & 4.68 \\
MID & 0.637 & -- & 0.476 & 33.05 & \underline{21.32} & \underline{11.30} \\
LiDiff & 0.564 & -- & 0.459 & \underline{33.23} & 17.55 & 4.88 \\
ScoreLiDAR & \underline{0.472} & -- & 0.444 & -- & -- & -- \\
Distillation-DPO & 0.533 & -- & \underline{0.434} & -- & -- & -- \\
FPSGen & \textbf{0.313} & 0.494 & \textbf{0.337} & \textbf{42.09} & \textbf{31.67} & \textbf{18.66} \\
\bottomrule
\end{tabular}
\caption{KITTI-360 LiDAR conditioned completion comparison. Lower is better for CD, \(JSD_{3D}\), and \(JSD_{BEV}\). Higher is better for IoU.}
\label{tab:kitti360_completion_supp}
\end{table*}

FPSGen is not optimized solely for completion, yet it improves every directly comparable KITTI-360 metric, including CD, \(JSD_{3D}\), \(JSD_{BEV}\), and all three voxel IoU resolutions. FPSGen also reports \(JSD_{3D}=0.494\), although this metric is unavailable for the published baselines and is therefore not used for a best method claim. These results support the claim that BEV-supported point transport preserves scene-level occupancy structure while remaining compatible with non-LiDAR generation modes.

\subsection{Condition Ablations}
\label{subsec:condition_ablations_supp}

Table~\ref{tab:completion_condition_ablation_supp} reports the SemanticKITTI \cite{semantickitti} LiDAR-conditioned completion ablation under different active condition tuples \(C_m=(m_lc_l,m_vc_v,m_rc_r)\). We keep the LiDAR cue active for completion because it is the primary condition that defines the completion task, and then evaluate road and vehicle masks as auxiliary layout cues.

\begin{table*}[t]
\centering
\setlength{\tabcolsep}{3.2pt}
\begin{tabular}{lccccccc}
\toprule
Condition & CD \(\downarrow\) & DCD \(\downarrow\) & \(JSD_{3D}\) \(\downarrow\) & \(JSD_{BEV}\) \(\downarrow\) & IoU@\(0.5\,\mathrm{m}\) \(\uparrow\) & IoU@\(0.2\,\mathrm{m}\) \(\uparrow\) & IoU@\(0.1\,\mathrm{m}\) \(\uparrow\) \\
\midrule
LiDAR & 0.316 & 0.588 & 0.492 & 0.329 & 44.15 & 32.26 & 18.26 \\
LiDAR + road & \underline{0.291} & 0.585 & \underline{0.488} & \underline{0.315} & 44.43 & 32.35 & 18.38 \\
LiDAR + vehicle & 0.307 & \underline{0.584} & \underline{0.488} & 0.325 & \underline{44.71} & \textbf{32.77} & \textbf{18.53} \\
All conditions & \textbf{0.288} & \textbf{0.583} & \textbf{0.485} & \textbf{0.313} & \textbf{44.87} & \underline{32.71} & \underline{18.49} \\
\bottomrule
\end{tabular}
\caption{Completion condition ablation on SemanticKITTI.}
\label{tab:completion_condition_ablation_supp}
\end{table*}

The ablation shows that additional layout cues remain beneficial even when LiDAR observations are available. Among individual auxiliary cues, the road mask provides the largest improvement in BEV distributional alignment, reducing \(JSD_{BEV}\) from 0.329 to 0.315, while the vehicle mask improves local occupancy overlap. Combining all conditions gives the best CD, DCD, \(JSD_{3D}\), \(JSD_{BEV}\), and IoU at \(0.5\,\mathrm{m}\). The small trade-off between IoU at \(0.2\,\mathrm{m}\) and \(0.1\,\mathrm{m}\) suggests that road and vehicle cues mainly improve scene-level support while fine local overlap remains sensitive to point-level geometry.

Table~\ref{tab:generation_condition_ablation_supp} reports LiDAR-free generation under road, vehicle, and joint road+vehicle conditions. Here \(m_l=0\) because generation is intended to test weak condition synthesis rather than sparse scan completion. We use the fixed generation subsets defined in the dataset protocol. SemanticKITTI \cite{semantickitti} is sampled every 10 frames and KITTI-360 \cite{kitti360} every 20 frames.

\begin{table*}[t]
\centering
\setlength{\tabcolsep}{3.0pt}
\begin{tabular}{llcccccc}
\toprule
Dataset & Condition & COV-CD \(\uparrow\) & MMD-CD \(\downarrow\) & 1-NNA-CD & COV-EMD \(\uparrow\) & MMD-EMD \(\downarrow\) & 1-NNA-EMD \\
\midrule
SemanticKITTI & Road & \textbf{60.78} & \underline{5.968} & \textbf{49.51} & \underline{53.92} & \underline{5.960} & \textbf{54.41} \\
SemanticKITTI & Vehicle & 40.20 & 8.425 & 73.77 & 45.59 & 6.656 & 68.38 \\
SemanticKITTI & Road + vehicle & \underline{58.33} & \textbf{5.858} & \underline{48.77} & \textbf{54.41} & \textbf{5.929} & \underline{55.64} \\
\midrule
KITTI-360 & Road & \underline{63.12} & \underline{5.585} & \textbf{51.24} & \underline{57.98} & \underline{5.259} & \underline{57.51} \\
KITTI-360 & Vehicle & 46.77 & 7.086 & 72.34 & 46.39 & 5.721 & 72.81 \\
KITTI-360 & Road + vehicle & \textbf{65.21} & \textbf{5.433} & \underline{47.72} & \textbf{60.27} & \textbf{5.150} & \textbf{56.08} \\
\bottomrule
\end{tabular}
\caption{Generation condition ablation. COV and 1-NNA are percentages. 1-NNA is best when closer to \(50\%\).}
\label{tab:generation_condition_ablation_supp}
\end{table*}

Road conditioning is the strongest single layout cue because it constrains the global drivable support and therefore strongly reduces distributional ambiguity. Vehicle masks alone are more local. They guide object placement, but leave the global road structure less constrained, resulting in lower coverage and higher 1-NNA. Combining road and vehicle cues generally provides the most balanced behavior. On KITTI-360, the joint condition improves COV-CD from 63.12 to 65.21 and MMD-CD from 5.585 to 5.433 compared with road alone. On SemanticKITTI, the road-only and joint settings are close. The road-only condition is closest to the ideal CD side 1-NNA, while the joint condition gives the best CD side MMD.

\subsection{Independent Transport Diagnostic}
\label{subsec:independent_transport_supp}

We also consider an intentionally weaker point-level pairing, denoted as independent transport. This diagnostic uses the same BEV source sampler \(\mathcal{R}\) as FPSGen, but removes the teacher-estimated clean endpoint matching. Let \(\mathcal{P}_{0}=\{p_{0,i}\}_{i=1}^{N}\) be the BEV-supported point source and let \(\mathcal{P}^{gt}=\{p^{gt}_i\}_{i=1}^{N}\) be the complete target point cloud in its stored point order. The independent transport path is formed by directly pairing these two unordered sets by index.
\begin{equation}
\mathcal{P}^{\mathrm{ind}}_t
=(1-t)\mathcal{P}_{0}+t\mathcal{P}^{gt},
\qquad
u^{\mathrm{ind}}_P
=\mathcal{P}^{gt}-\mathcal{P}_{0}.
\label{eq:independent_transport_supp}
\end{equation}
Unlike the Approx OT teacher pair \((\mathcal{P}_{0},\mathcal{P}^{\dagger}_{1})\), this construction does not estimate a source-indexed clean endpoint. Since raw point clouds are unordered, the index wise target \(p^{gt}_i\) is generally unrelated to the BEV anchor that produced \(p_{0,i}\). The resulting velocity field therefore contains many long range and crossing motions, even though the source marginal itself remains BEV-supported.

This diagnostic is different from the \(\beta\)-hybrid perturbation below. Hybrid coupling keeps the teacher-estimated clean endpoint unchanged and only mixes the local perturbation around each sampled BEV anchor. It weakens the source-to-endpoint coupling while preserving the BEV-supported point source marginal. Independent transport instead replaces the teacher-estimated clean endpoint with the original unordered target point set, thereby discarding the teacher's local source-to-target relation. In our experiments this variant performs poorly, which supports the need for a source-indexed teacher mapping rather than a globally independent point set pairing.

\begin{table*}[t]
\centering
\setlength{\tabcolsep}{4.0pt}
\begin{tabular}{cccccccc}
\toprule
\(K_P\) & CD \(\downarrow\) & DCD \(\downarrow\) & \(JSD_{3D}\) \(\downarrow\) & \(JSD_{BEV}\) \(\downarrow\) & IoU@\(0.5\,\mathrm{m}\) \(\uparrow\) & IoU@\(0.2\,\mathrm{m}\) \(\uparrow\) & IoU@\(0.1\,\mathrm{m}\) \(\uparrow\) \\
\midrule
1 & 5.728 & 0.966 & 0.798 & 0.754 & 4.30 & 3.38 & 2.41 \\
2 & 1.925 & 0.855 & 0.722 & 0.576 & 14.81 & 9.59 & 4.99 \\
4 & 0.795 & \underline{0.814} & \underline{0.701} & 0.465 & 18.64 & 10.95 & 5.24 \\
8 & \textbf{0.623} & \textbf{0.804} & \textbf{0.695} & \underline{0.422} & \textbf{19.70} & \textbf{11.32} & \textbf{5.38} \\
16 & \textbf{0.623} & \underline{0.814} & 0.702 & \textbf{0.420} & \underline{19.10} & \underline{10.82} & \underline{5.21} \\
32 & 0.659 & 0.831 & 0.712 & 0.437 & 17.68 & 9.83 & 4.85 \\
50 & 0.680 & 0.838 & 0.717 & 0.447 & 16.94 & 9.37 & 4.70 \\
\bottomrule
\end{tabular}
\caption{Independent transport diagnostic for SemanticKITTI completion. Lower is better for CD, DCD, \(JSD_{3D}\), and \(JSD_{BEV}\). Higher is better for IoU.}
\label{tab:independent_transport_supp}
\end{table*}

The full diagnostic shows a large gap between independent transport and the teacher-estimated pairing. Increasing \(K_P\) improves the independent transport variant from the degenerate one-step result, but it saturates around \(K_P=8\) to \(16\) and remains much worse than the Approx OT teacher-estimated coupling used by FPSGen. This is consistent with the intuition that index-wise pairing between unordered point sets creates unnecessarily long-range and crossing transport trajectories. More Euler steps can reduce discretization error, but they cannot recover the missing source-indexed endpoint relation.

We also compare independent transport against BEV-local hybrid couplings at \(K_P=50\) point-flow steps in Table~\ref{tab:beta_ablation_supp}. Within the BEV-local family, the hybrid perturbation coefficient \(\beta\) weakens the locally teacher-estimated coupling while preserving the BEV-supported point source marginal. The independent row is substantially worse than all BEV-local rows, showing that the source-indexed teacher endpoint is important for stable point transport. Among the BEV local variants, \(\beta=0\) gives the strongest distributional support metrics, while small nonzero perturbations can slightly improve point-level CD.

\begin{table}[t]
\centering
\setlength{\tabcolsep}{4.5pt}
\begin{tabular}{lcccc}
\toprule
Coupling & \(\beta\) & CD \(\downarrow\) & \(JSD_{3D}\) \(\downarrow\) & \(JSD_{BEV}\) \(\downarrow\) \\
\midrule
Independent & -- & 0.6804 & 0.7165 & 0.4475 \\
\midrule
BEV local & 0.00 & 0.3162 & \textbf{0.4922} & \textbf{0.3287} \\
BEV local & 0.01 & 0.3166 & 0.4928 & \underline{0.3289} \\
BEV local & 0.05 & 0.3156 & 0.4925 & 0.3303 \\
BEV local & 0.20 & \textbf{0.3151} & \underline{0.4923} & 0.3306 \\
BEV local & 1.00 & \underline{0.3155} & 0.4928 & 0.3312 \\
\bottomrule
\end{tabular}
\caption{Point-stage source-to-target coupling ablation on SemanticKITTI with \(K_P=50\).}
\label{tab:beta_ablation_supp}
\end{table}

\subsection{Sinkhorn Approx OT Scalability Diagnostic}
\label{subsec:sinkhorn_approx_ot_supp}

Not-So-Optimal Transport Flows for 3D Point Cloud Generation studies an OT approximation motivated by the law of large numbers, where a much larger offline candidate set is sampled and the final point set is then drawn from it \cite{hui2025notsooptimal}. In their object-scale setting, the final point cloud has \(2048\) points while the offline pool contains \(100{,}000\) candidates, giving a ratio of about \(48.8\times\). We evaluate the same idea with a Sinkhorn-based approximation in our large scene setting using every 100th frame of SemanticKITTI sequence 08. Since our target point cloud contains \(180{,}000\) points, we measure candidate pools up to \(1.8\) million points, corresponding to a \(10\times\) pool. The analogous \(50\times\) setting would require about \(9.0\) million candidates per scene.

\begin{table*}[t]
\centering
\setlength{\tabcolsep}{5.0pt}
\begin{tabular}{llccccc}
\toprule
Method & Ratio & Points & Time \(\downarrow\) & Disp. \(\downarrow\) & CD \(\downarrow\) & DCD \(\downarrow\) \\
\midrule
Teacher-estimated clean endpoint & -- & \(180{,}000\) & \textbf{0.31} & \textbf{1.231} & \textbf{0.143} & 0.468 \\
\midrule
Sinkhorn Approx OT & \(1\times\) & \(180{,}000\) & \underline{27.57} & 1.851 & \underline{0.146} & \textbf{0.374} \\
Sinkhorn Approx OT & \(2\times\) & \(360{,}000\) & 62.04 & 1.856 & 0.151 & \underline{0.417} \\
Sinkhorn Approx OT & \(4\times\) & \(720{,}000\) & 161.10 & 1.850 & 0.153 & 0.435 \\
Sinkhorn Approx OT & \(6\times\) & \(1{,}080{,}000\) & 309.05 & 1.832 & 0.153 & 0.440 \\
Sinkhorn Approx OT & \(10\times\) & \(1{,}800{,}000\) & 762.97 & \underline{1.815} & 0.154 & 0.445 \\
\bottomrule
\end{tabular}
\caption{Sinkhorn Approx OT scalability diagnostic on SemanticKITTI sequence 08, averaged over scans sampled every 100 frames. Time is seconds per scene. Disp. is the mean transport displacement. CD denotes \(cd_p\). Ratio and points describe the Sinkhorn candidate pool size relative to the \(180{,}000\) point target scene.}
\label{tab:sinkhorn_approx_ot_supp}
\end{table*}

\begin{figure*}[t]
\centering
\includegraphics[width=\linewidth]{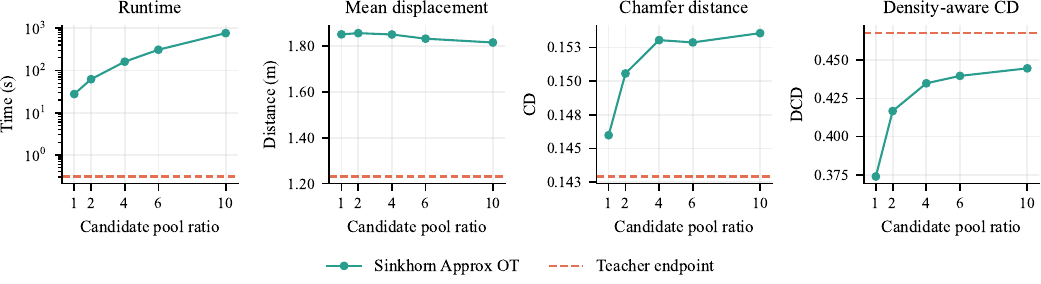}
\caption{Scaling behavior of the Sinkhorn Approx OT diagnostic. The teacher estimated clean endpoint is shown as a horizontal reference because it is measured only at the \(180{,}000\) point scene scale and does not vary with the candidate pool ratio. Runtime grows rapidly with the candidate pool size, while the random candidate sampling protocol used by the large pool approximation slightly worsens CD and DCD in this scene scale setting.}
\label{fig:approx_ot_scaling_supp}
\end{figure*}

Table~\ref{tab:sinkhorn_approx_ot_supp} and Figure~\ref{fig:approx_ot_scaling_supp} show that the Sinkhorn approximation is difficult to use as an online scene-scale pairing module. Even at the native \(180{,}000\) point scale, the measured runtime is \(27.57\) seconds per scene, about \(89\times\) slower than the amortized teacher-estimated clean endpoint construction. It also achieves CD comparable to the teacher-estimated endpoint construction rather than a clear geometric improvement, although the \(1\times\) Sinkhorn setting gives lower DCD. As the candidate pool grows from \(1\times\) to \(10\times\), the runtime increases from \(27.57\) to \(762.97\) seconds. Meanwhile, the random sampling strategy used in the large candidate-pool approximation introduces additional sampling variation, and the measured CD and DCD become slightly worse rather than improving with more candidates.

The offline candidate pool strategy is also impractical at SemanticKITTI scale. The measured \(10\times\) case already takes \(762.97\) seconds, or \(12.72\) minutes, for a single scene. The SemanticKITTI training split contains \(19{,}130\) samples, so precomputing such approximate OT pairings would require roughly \(4{,}054\) hours under the same measured setup. A \(50\times\) pool would require about \(9.0\) million candidates per scene and would further increase this cost. This is disproportionate for a pairing step and motivates our teacher-based Approx OT design, which amortizes the scene-scale correspondence problem while avoiding explicit Sinkhorn matching over millions of candidate points.

\subsection{Marginal Preservation of Hybrid Coupling}
\label{subsec:hybrid_coupling_supp}

We briefly justify the hybrid perturbation used in the \(\beta\)-ablation. Endpoint coupling choices are central in flow matching and conditional flow matching \cite{lipman2022flowmatching,tong2023otcfm}. Hybrid and deliberately non-optimal source target couplings have recently been studied for point cloud flow models, where relaxing an exact OT coupling can make the learned transport easier to model \cite{hui2025notsooptimal}. Our setting differs because the source is not sampled directly from an unconditional Gaussian. FPSGen first samples BEV grid anchors according to the normalized density channel \(D\), and then adds coordinate noise. Thus, \(\beta\) is used here as a controlled diagnostic for weakening the local teacher-estimated coupling while preserving the BEV-supported point source marginal.

For completeness, we restate the source distribution used in this analysis. Given a BEV prior \(B=[D,H,M]\), the normalized density value \(D(q)\in[-1,1]\) at BEV cell \(q\) is converted to a nonnegative score
\begin{equation}
\rho_D(q)=
\exp\!\left(\frac{D(q)+1}{2}\log(1+n_{\max})\right)-1 ,
\label{eq:hybrid_density_denorm_supp}
\end{equation}
where \(n_{\max}\) is the density clipping constant. Using the same BEV source sampler notation, the cell sampling probability is
\begin{equation}
w_D(q)=
\frac{\max(\rho_D(q),0)+\varepsilon_w}
{\sum_{q'}(\max(\rho_D(q'),0)+\varepsilon_w)} .
\label{eq:hybrid_density_weight_supp}
\end{equation}
Here \(\varepsilon_w>0\) is a small numerical stabilizer. A source point is obtained by sampling \(q_i\sim\mathrm{Cat}(w_D)\), mapping the cell to its zero-height metric anchor \(a_i=p^b(q_i)=(x(q_i),y(q_i),0)\), and adding Gaussian perturbation \(\zeta_i\sim\mathcal{N}(0,\Sigma)\), so that \(p_{0,i}=a_i+\zeta_i\). The covariance \(\Sigma\) is the coordinate noise covariance used by the BEV source sampler \(\mathcal{R}\).

For hybrid coupling, we keep the same sampled anchor and replace the perturbation by
\begin{equation}
\tilde{p}_{0,i}^{(\beta)}=a_i+\tilde{\zeta}_{i}^{(\beta)},\qquad
\tilde{\zeta}_{i}^{(\beta)}
=\sqrt{1-\beta}\,\zeta_i+\sqrt{\beta}\,\epsilon_i,
\label{eq:hybrid_anchor_noise}
\end{equation}
where \(\epsilon_i\sim\mathcal{N}(0,\Sigma)\) is an independent Gaussian perturbation and \(\beta\in[0,1]\). Conditioned on the anchor \(a_i\), the mixed perturbation remains Gaussian.
\begin{equation}
\begin{array}{l}
\mathrm{E}[\tilde{\zeta}_{i}^{(\beta)}]=0,\\
\mathrm{Cov}[\tilde{\zeta}_{i}^{(\beta)}]
=(1-\beta)\Sigma+\beta\Sigma=\Sigma .
\end{array}
\label{eq:hybrid_noise_moments}
\end{equation}
Therefore \(\tilde{p}_{0,i}^{(\beta)}\mid q_i\sim\mathcal{N}(a_i,\Sigma)\). Marginalizing over the density-sampled BEV cell gives
\begin{equation}
p_{\beta}(x\mid B)
=\sum_{q}w_D(q)\,
\mathcal{N}\!\left(x;p^b(q),\Sigma\right),
\label{eq:hybrid_source_marginal}
\end{equation}
which is exactly the single-point marginal of the BEV-supported point source construction. Since the sampled cells and perturbations are independent across points, the same conclusion holds for the full point set under the product construction conditioned on \(B\).

It remains to verify that the clean-endpoint marginal is unchanged. Let \(Y=\mathcal{P}^{\dagger}_{1}\) denote the teacher-estimated clean endpoint paired with the original BEV-supported point source, and let \(A\) and \(Z\) collect the sampled anchors and original perturbations. The hybrid source \(\tilde{\mathcal{P}}_{0}\) is sampled from a conditional transition \(p_{\beta}(\tilde{\mathcal{P}}_{0}\mid A,Z)\) that integrates to one. Hence
\begin{equation}
\begin{array}{l}
\displaystyle
\int p_{\beta}(\tilde{\mathcal{P}}_{0},Y\mid \bar{B},\mathcal{P}^{gt})
\,d\tilde{\mathcal{P}}_{0}\\
\displaystyle\quad =
\int p(Y,A,Z\mid \bar{B},\mathcal{P}^{gt})\\
\displaystyle\quad \cdot
\left[\int p_{\beta}(\tilde{\mathcal{P}}_{0}\mid A,Z)
d\tilde{\mathcal{P}}_{0}\right]dA\,dZ\\
\displaystyle\quad =
p(Y\mid \bar{B},\mathcal{P}^{gt}).
\end{array}
\label{eq:hybrid_target_marginal}
\end{equation}
Thus, the hybrid perturbation changes the coupling between the BEV-supported point source and the teacher-estimated clean endpoint, while preserving the clean-endpoint marginal. The source remains the same BEV density mixture defined by \(w_D\), and the target remains the same teacher-estimated clean endpoint distribution. The coefficient \(\beta\) therefore provides a controlled way to move from the teacher-aligned perturbation toward a less coupled perturbation while keeping the training marginals fixed.

\subsection{Hybrid Coupling Results}
\label{subsec:jacobian_analysis_supp}

\begin{figure}[t]
\centering
\includegraphics[width=\linewidth]{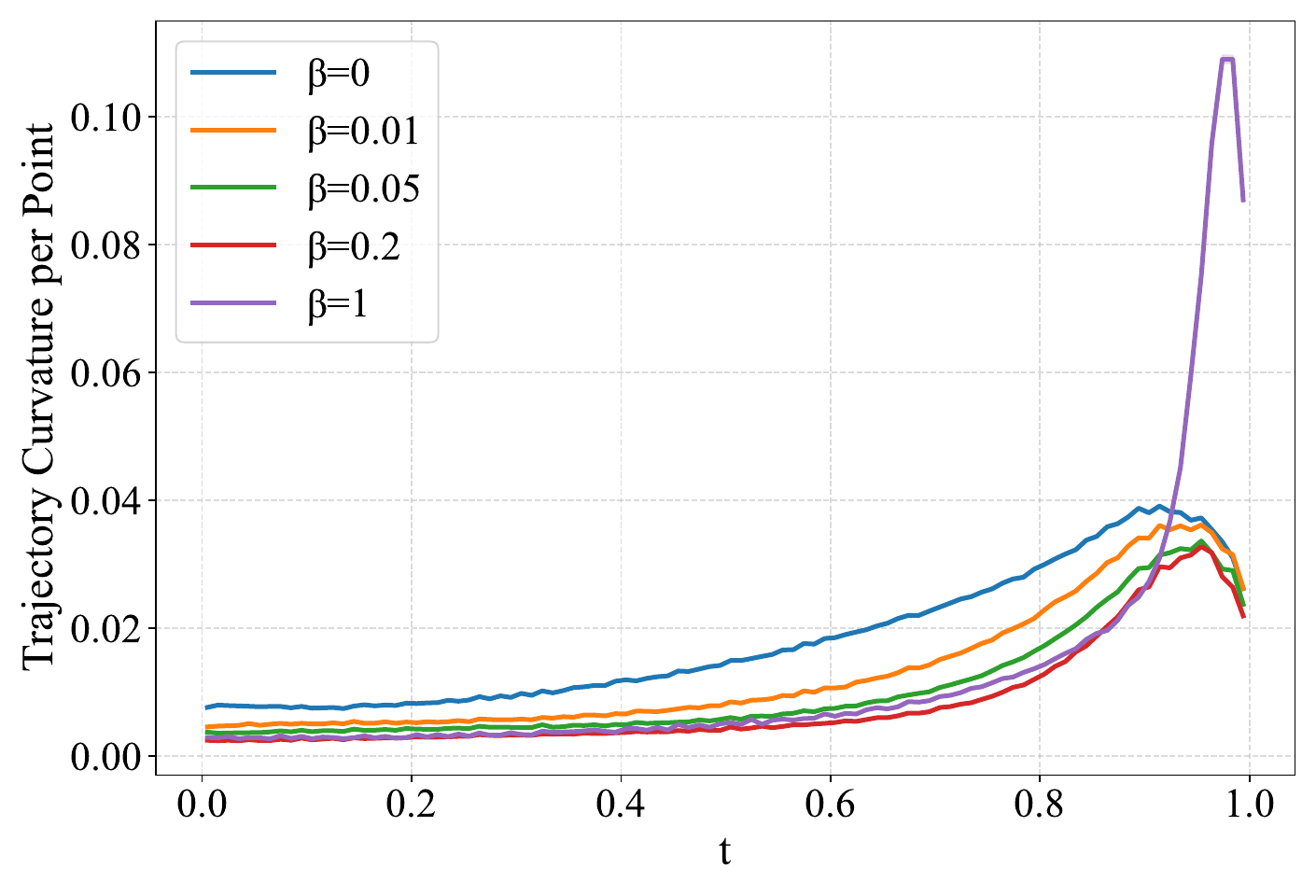}
\caption{Discrete trajectory curvature proxy per point along point transport paths for different \(\beta\) values under the generated BEV prior \(\hat{B}\) and the full condition state \(m_lm_vm_r=111\). Curves are averaged over the 408 SemanticKITTI sequence 08 samples obtained by taking every 10th frame. Each trajectory uses 100 uniform integration steps, and each value is assigned to the midpoint between two adjacent recorded times.}
\label{fig:curvature_beta_supp}
\end{figure}

\begin{figure}[t]
\centering
\includegraphics[width=\linewidth]{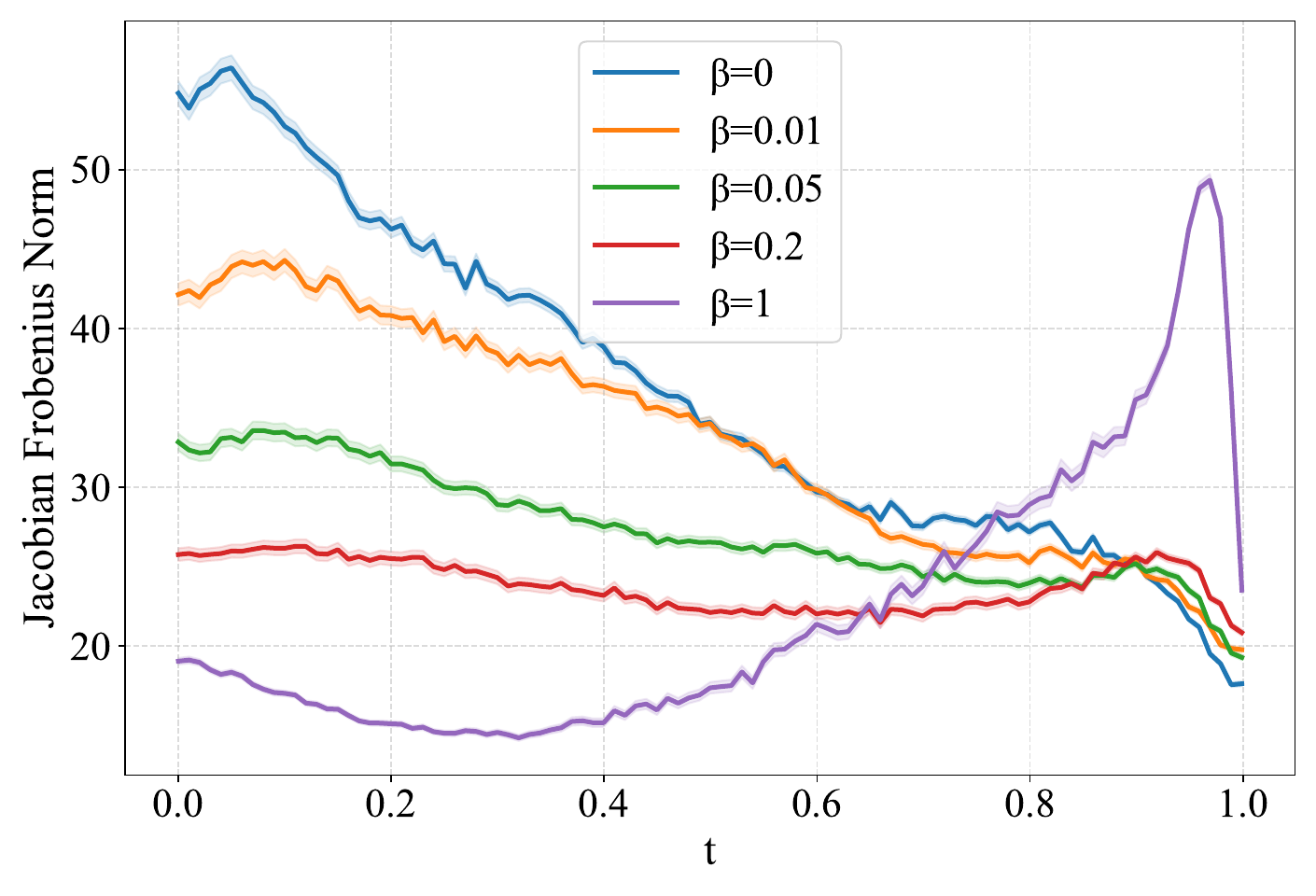}
\caption{Jacobian Frobenius norm along point transport paths for different \(\beta\) values under the generated BEV prior \(\hat{B}\) and the full condition state \(m_lm_vm_r=111\). Curves are averaged over the 408 SemanticKITTI sequence 08 samples obtained by taking every 10th frame, using 100 uniform integration steps and 4 Rademacher probes at each recorded state. Shaded regions show the measured variation across samples.}
\label{fig:jacobian_beta_supp}
\end{figure}

We further analyze the point flow matching paths induced by different hybrid perturbation coefficients \(\beta\in\{0.0,0.01,0.05,0.2,1.0\}\). This experiment complements the non-optimal coupling perspective of \cite{hui2025notsooptimal}. Instead of assuming that a less coupled path is always preferable, we test how much coupling should be retained when the source already follows BEV scene support. We evaluate the fixed subset of 408 samples obtained by taking every 10th frame of SemanticKITTI sequence 08. For every sample, the point flow receives the generated BEV prior \(\hat{B}\) and the full condition state \(m_l=m_v=m_r=1\), so \(C_m=(c_l,c_v,c_r)\). Each trajectory uses 100 uniform integration steps from the BEV-supported source endpoint \(t=0\) to the teacher-estimated clean endpoint \(t=1\). At every recorded state, we reevaluate the student velocity while keeping \(\hat{B}\) and \(C_m\) fixed. The Jacobian estimate uses \(K_{\mathrm{probe}}=4\) independently sampled Rademacher probes and differentiates only with respect to the point cloud state.

Let \(t_i\) and \(t_{i+1}\) be two adjacent recorded times, and let \(v_i=v_\psi(\mathcal{P}_{t_i},t_i,\hat{B},C_m)\) denote the predicted point velocity. Because the trajectory uses a uniform 100-step discretization, we use the squared finite difference of adjacent velocities as a discrete trajectory curvature proxy per point.
\begin{equation}
\kappa_i^{\mathrm{point}}
=\frac{1}{N}
\left\|v_{i+1}-v_i\right\|_F^2 .
\label{eq:trajectory_curvature_supp}
\end{equation}
Here \(N\) is the number of points and \(\kappa_i^{\mathrm{point}}\) is plotted at the midpoint \((t_i+t_{i+1})/2\) in Figure~\ref{fig:curvature_beta_supp}. This finite-difference quantity measures velocity variation under the fixed uniform discretization. It is a trajectory curvature proxy rather than a parameterization-invariant geometric curvature. Let \(\operatorname{vec}(\cdot)\) stack all point coordinates into a vector in \(\mathbb{R}^{3N}\). We also estimate the Jacobian of the vectorized point velocity field with respect to the vectorized point cloud state,
\begin{equation}
J_t=
\frac{
\partial\,\operatorname{vec}
\left(v_\psi(\mathcal{P}_t,t,\hat{B},C_m)\right)}
{\partial\,\operatorname{vec}(\mathcal{P}_t)} .
\label{eq:jacobian_def_supp}
\end{equation}
Following the Hutchinson estimator \cite{hutchinson1989stochastic}, we use \(K_{\mathrm{probe}}\) independent Rademacher probes \(z_k\in\{-1,+1\}^{3N}\) and define
\begin{equation}
\ell_k(t)=
z_k^\top
\operatorname{vec}\left(v_\psi(\mathcal{P}_t,t,\hat{B},C_m)\right).
\label{eq:jacobian_probe_scalar_supp}
\end{equation}
Automatic differentiation with respect to \(\operatorname{vec}(\mathcal{P}_t)\) gives
\(\nabla_{\operatorname{vec}(\mathcal{P}_t)}\ell_k(t)=J_t^\top z_k\).
The Jacobian Frobenius norm used in Figure~\ref{fig:jacobian_beta_supp} is then estimated by
\begin{equation}
\|\hat{J}_t\|_F
=
\left[
\frac{1}{K_{\mathrm{probe}}}
\sum_{k=1}^{K_{\mathrm{probe}}}
\left\|
\nabla_{\operatorname{vec}(\mathcal{P}_t)}
\ell_k(t)
\right\|_2^2
\right]^{1/2}.
\label{eq:jacobian_probe_supp}
\end{equation}

Figure~\ref{fig:curvature_beta_supp} shows that \(\beta=0.0\) has a larger curvature proxy than the other locally perturbed variants before approximately \(t=0.8\). Near the clean endpoint, the values for \(\beta=0.01\), \(\beta=0.05\), and \(\beta=0.2\) become close to that of \(\beta=0.0\). The fully independent perturbation \(\beta=1.0\), however, increases sharply near \(t=1\). This indicates that making the source perturbation too independent creates a terminal segment with rapidly changing velocity, even though its early trajectory has smaller velocity variation.

Figure~\ref{fig:jacobian_beta_supp} provides a complementary view of vector field complexity. The \(\beta=0.0\) field has the largest Jacobian Frobenius norm at the beginning of point transport, namely near \(t=0\), and decreases overall along the path. A high initial Jacobian means that the learned velocity is sensitive to small changes in the BEV supported point source. For an Euler update with interval \(\Delta t\), let \(\delta_i\) denote a small state error and let \(e_i\) denote the local velocity prediction error. Their first-order propagation is
\begin{equation}
\delta_{i+1}
\approx
(I+\Delta t J_{t_i})\delta_i+\Delta t e_i.
\label{eq:jacobian_error_propagation_supp}
\end{equation}
A large Jacobian can therefore amplify an existing state error, while the additive term accumulates local velocity error across updates. This sensitivity-based explanation is consistent with the \(K_P=2,4,8\) samplers in Table~\ref{tab:forward_step_supp} underperforming the one-step endpoint update, although the Jacobian norm alone does not establish causality. The \(K_P=16\) sampler recovers performance comparable to the one-step setting. The \(K_P=32\) and \(K_P=50\) samplers better resolve the path and improve CD and \(JSD_{3D}\), while the one-step setting retains the best \(JSD_{BEV}\).

The \(\beta=1.0\) Jacobian curve follows the opposite pattern. It is low near the source but grows rapidly near the endpoint, matching the terminal increase of the curvature proxy in Figure~\ref{fig:curvature_beta_supp}. The intermediate \(\beta\) values reduce the initial Jacobian relative to \(\beta=0.0\), but they do not improve the final completion metrics consistently. Thus, in our BEV-supported scene setting, weakening the local teacher-estimated coupling can smooth part of the trajectory, yet it also introduces less favorable endpoint behavior or weaker scene support. This supports the use of the most strongly coupled local construction, especially when one step or sufficiently resolved integration is used.

\begin{figure*}[t]
\centering
\includegraphics[width=\linewidth]{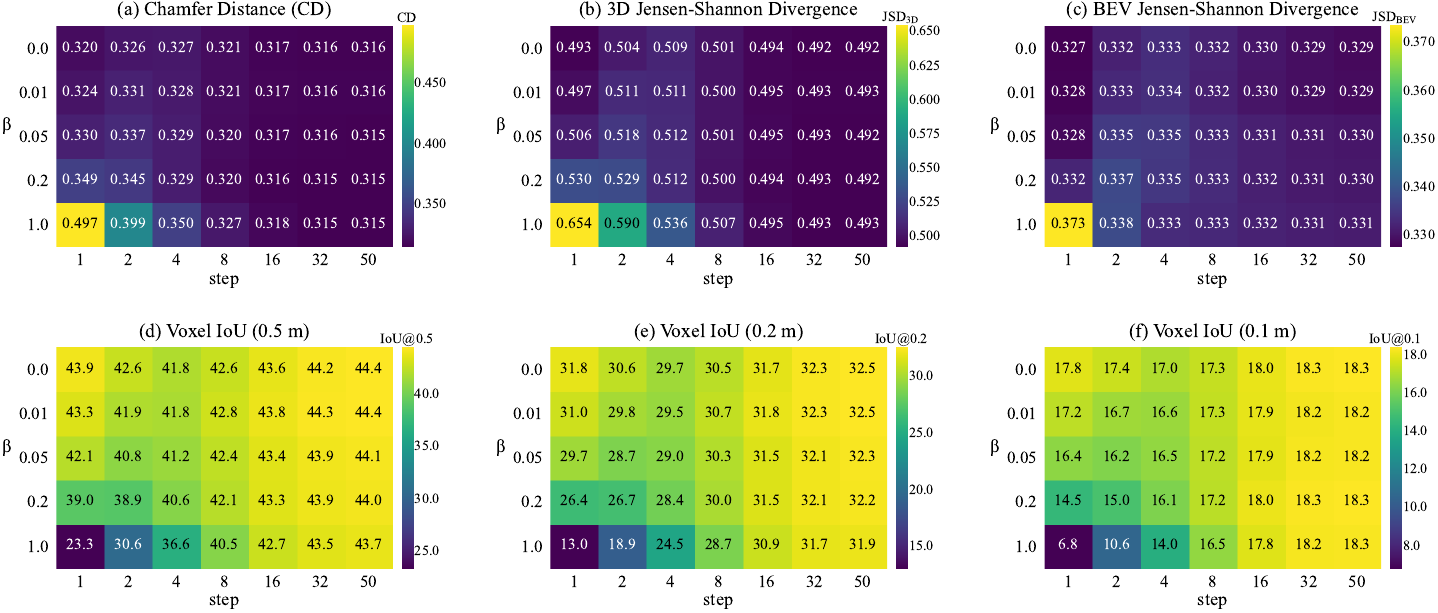}
\caption{Completion quality heatmaps over \(\beta\) and point-flow steps \(K_P\). Panels (a)--(c) show CD, \(JSD_{3D}\), and \(JSD_{BEV}\). Panels (d)--(f) show voxel IoU at \(0.5\,\mathrm{m}\), \(0.2\,\mathrm{m}\), and \(0.1\,\mathrm{m}\). Several \(\beta\) values become competitive with enough point flow steps, while \(\beta=0.0\) provides an unusually strong one step result and recovers when the sampler uses larger \(K_P\).}
\label{fig:beta_step_heatmaps_supp}
\end{figure*}

Figure~\ref{fig:beta_step_heatmaps_supp} summarizes the corresponding completion quality across \(\beta\) and point-flow step counts \(K_P\). Panels (a)--(c) report CD, \(JSD_{3D}\), and \(JSD_{BEV}\). Panels (d)--(f) report voxel IoU at \(0.5\,\mathrm{m}\), \(0.2\,\mathrm{m}\), and \(0.1\,\mathrm{m}\). The heatmaps show that different \(\beta\) values can converge to competitive final completion quality when the point sampler is sufficiently resolved. Nevertheless, \(\beta=0.0\) is distinctive because it achieves a strong one-step result while also recovering at larger \(K_P\). For this SemanticKITTI completion setting, the most coupled Approx OT variant performs better overall than deliberately independent transport. The BEV density prior already places source points near plausible scene support, so preserving the local teacher-estimated pairing reduces unnecessary long-range motion and gives a simpler one-step source-to-endpoint mapping.

\subsection{Endpoint CD Dynamics}
\label{subsec:endpoint_cd_dynamics_supp}

\begin{figure}[t]
\centering
\includegraphics[width=\linewidth]{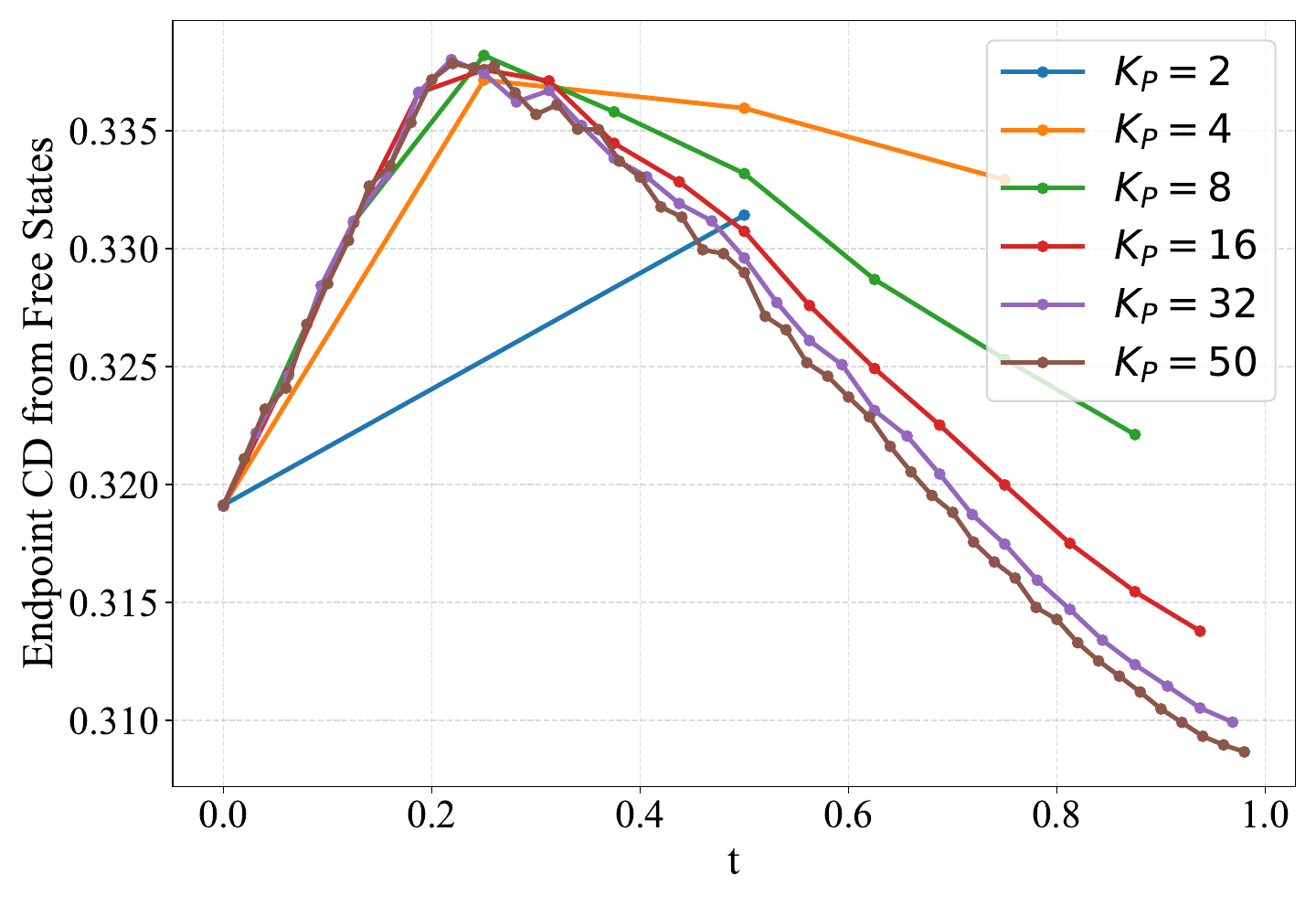}
\caption{Endpoint CD estimated from free point flow states under the generated BEV prior \(\hat{B}\), the full condition state \(m_lm_vm_r=111\), and \(\beta=0.0\). At each recorded state, the remaining interval is directly extrapolated with the local point velocity, and the resulting endpoint is compared with the fixed teacher-estimated clean endpoint \(\mathcal{P}^{\dagger}_{1}\). The curves show an early increase in teacher endpoint error followed by recovery when the trajectory is resolved with enough point-flow steps.}
\label{fig:endpoint_cd_dynamics_supp}
\end{figure}

We analyze why a small number of point-flow steps can perform worse than either a direct one step update or a more finely resolved trajectory. This diagnostic uses the same fixed subset of 408 SemanticKITTI sequence 08 samples obtained by taking every 10th frame. As in the preceding path analysis, the point flow receives the generated BEV prior \(\hat{B}\) and the full condition state \(m_l=m_v=m_r=1\), so \(C_m=(c_l,c_v,c_r)\). For each point step count \(K_P\in\{2,4,8,16,32,50\}\), let \(\mathcal{P}^{\mathrm{free}}_t\) denote a state reached by free Euler integration from the same BEV-supported point source. The teacher-estimated clean endpoint \(\mathcal{P}^{\dagger}_{1}\) associated with that source pair is held fixed as the reference. At every recorded state, we use the local velocity to directly predict the remaining displacement to the endpoint,
\begin{equation}
\begin{array}{l}
\widetilde{\mathcal{P}}^{\mathrm{free}}_1(t)
=\mathcal{P}^{\mathrm{free}}_t
+(1-t)\\[1pt]
\displaystyle\qquad v^{\mathrm{cfg}}_\psi
\left(\mathcal{P}^{\mathrm{free}}_t,t,\hat{B},C_m\right).
\end{array}
\label{eq:free_endpoint_prediction_supp}
\end{equation}
We then measure the permutation-invariant endpoint error against the teacher estimated clean endpoint,
\begin{equation}
E_{\mathrm{end}}^{\mathrm{CD}}(t)
=\mathrm{CD}\left(
\widetilde{\mathcal{P}}^{\mathrm{free}}_1(t),
\mathcal{P}^{\dagger}_{1}
\right).
\label{eq:free_endpoint_cd_supp}
\end{equation}
This quantity evaluates how accurately the vector field at the current free state recovers the endpoint assigned by the teacher-estimated pairing. It neither measures the geometric distance between the intermediate state and the transport path nor directly replaces the final CD to \(\mathcal{P}^{gt}\) reported in the task evaluation.

Figure~\ref{fig:endpoint_cd_dynamics_supp} shows that all trajectories begin from the same one-step endpoint estimate, with CD to \(\mathcal{P}^{\dagger}_{1}\) close to \(0.319\). For \(K_P=4,8,16,32,50\), this teacher-endpoint error first increases and reaches approximately \(0.338\) near \(t=0.2\) to \(0.25\). The early free states therefore imply less accurate direct recovery of the paired teacher endpoint than the original source state. After this transient growth, the curves decrease as integration proceeds. The \(K_P=2\) and \(K_P=4\) samplers provide too few evaluations through this correction region, and their final recorded states remain above the common source estimate. The \(K_P=8\) trajectory only partially recovers. At \(K_P=16\), endpoint CD falls below the one-step estimate, while \(K_P=32\) and \(K_P=50\) continue through the correction phase and reach the lowest errors. Their increasingly similar curves also indicate that the trajectory becomes better resolved as \(K_P\) grows.

The source-side Jacobian behavior in Figure~\ref{fig:jacobian_beta_supp} is consistent with this two-phase pattern. The large Jacobian near \(t=0\) indicates that the predicted velocity is locally sensitive to changes in the initial point state, which can make the first few free updates enter a region with higher teacher-endpoint error. The subsequent decrease in endpoint CD provides direct set-level evidence that the learned field progressively recovers the teacher assigned endpoint rather than remaining uniformly unstable. This relationship should be interpreted as complementary evidence rather than a causal conclusion from the Jacobian alone. Overall, the diagnostic explains the non-monotonic step ablation as an interaction between an initially sensitive region and the number of updates available to complete the later trajectory correction.

\subsection{Runtime and Memory Benchmark}
\label{subsec:runtime_supp}

Table~\ref{tab:runtime_supp} reports the single-GPU inference benchmark used in the scalability discussion. All methods are evaluated on SemanticKITTI \cite{semantickitti} sequence 08 with the same \(50\mathrm{m}\) range, after 5 warmup samples and over 20 measured samples. The benchmark records wall clock inference time per scene and peak reserved GPU memory on a single RTX 3090 GPU. FPSGen runs the BEV flow and the student point flow at inference, so we report the two step counts separately. The other methods use a single point-stage sampler. The teacher network is used only during training.

\begin{table*}[t]
\centering
\setlength{\tabcolsep}{2.2pt}
\begin{tabular}{lcccccc}
\toprule
Method & \(K_B\) & \(K_P\) & CD \(\downarrow\) & Inference Params (M) & Time (s) \(\downarrow\) & Mem. (GB) \(\downarrow\) \\
\midrule
LiDiff & -- & 50 & 0.434 & 32.67 & 33.75 & 6.18 \\
LiDPM & -- & 20 & 0.446 & 32.67 & 20.92 & 5.53 \\
ScoreLiDAR & -- & 8 & 0.406 & 32.67 & 7.73 & 6.07 \\
Distillation-DPO & -- & 8 & 0.414 & 32.67 & \underline{5.81} & 6.07 \\
\midrule
FPSGen & 10 & 1 & \underline{0.321} & 28.55 & \textbf{0.95} & \underline{4.05} \\
FPSGen & 10 & 32 & \textbf{0.316} & 28.55 & 15.57 & \textbf{4.04} \\
\bottomrule
\end{tabular}
\caption{Runtime and memory benchmark on SemanticKITTI sequence 08. Time is wall clock seconds per scene averaged over 20 measured samples after 5 warmup samples. Inference parameters are reported in millions, and memory is peak reserved GPU memory.}
\label{tab:runtime_supp}
\end{table*}

FPSGen with \(K_B=10\) and \(K_P=1\) already obtains a CD close to the \(K_P=32\) setting at substantially lower runtime. FPSGen is faster than LiDiff \cite{lidiff} and LiDPM \cite{martyniuk2025lidpm} under the measured \(K_B=10, K_P=32\) setting, while also using less peak reserved memory. ScoreLiDAR \cite{scorelidar} and Distillation-DPO \cite{zhao2026distillationdpo} are faster in this benchmark because they use \(K_P=8\), whereas FPSGen uses \(K_B=10, K_P=32\). The memory comparison is nevertheless favorable to FPSGen, with measured peak reserved GPU memory of about \(4.04\) GB compared with \(5.53\)--\(6.18\) GB for the completion baselines.

\begin{figure}[t]
\centering
\includegraphics[width=\linewidth]{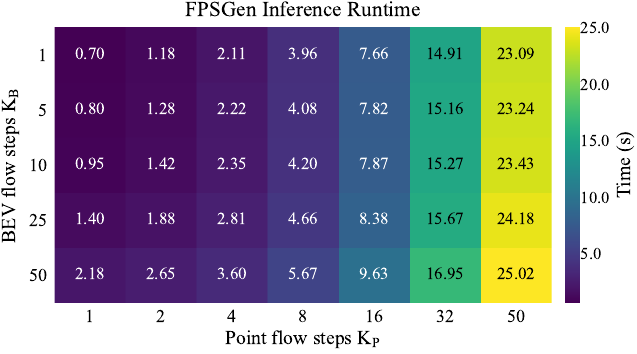}
\caption{Separate FPSGen runtime sweep over BEV-flow steps \(K_B\) and point flow steps \(K_P\). Each cell reports average wall clock seconds per scene on one RTX 3090.}
\label{fig:runtime_fpsgen_steps_supp}
\end{figure}

We additionally measure FPSGen under different BEV-flow and point-flow step counts in a separate sweep. Figure~\ref{fig:runtime_fpsgen_steps_supp} reports the average runtime per scene for this FPSGen-only evaluation. Runtime grows almost linearly with \(K_P\), while increasing \(K_B\) adds a smaller but visible overhead. Across all tested FPSGen settings, peak reserved GPU memory ranges from \(3.53\) GB to \(4.14\) GB.

\subsection{Generation Results with DCD}
\label{subsec:generation_dcd_supp}

Table~\ref{tab:generation_dcd_supp} supplements the main generation tables with DCD-based distributional metrics \cite{wu2021densityaware}. DCD highlights whether a generated scene matches not only the nearest-neighbor geometry but also the density allocation of the reference point clouds.

\begin{table*}[t]
\centering
\setlength{\tabcolsep}{5.0pt}
\begin{tabular}{lllccc}
\toprule
Dataset & Protocol & Method & COV-DCD \(\uparrow\) & MMD-DCD \(\downarrow\) & 1-NNA-DCD \\
\midrule
SemanticKITTI & Unconditional & LiDiff & 13.24 & 0.916 & 100.00 \\
SemanticKITTI & Unconditional & LiDPM & 16.91 & 0.825 & 99.88 \\
SemanticKITTI & Unconditional & SemCity & \underline{19.36} & \underline{0.816} & \underline{99.63} \\
SemanticKITTI & Unconditional & FPSGen & \textbf{24.75} & \textbf{0.765} & \textbf{87.87} \\
\midrule
KITTI-360 & Unconditional & LiDiff & 19.39 & 0.823 & 99.62 \\
KITTI-360 & Unconditional & LiDPM & \textbf{31.56} & \underline{0.773} & \underline{80.61} \\
KITTI-360 & Unconditional & SemCity & 16.16 & 0.816 & 99.71 \\
KITTI-360 & Unconditional & FPSGen & \underline{24.90} & \textbf{0.764} & \textbf{73.76} \\
\midrule
KITTI-360 & LiDM-seeded & ScoreLiDAR & \underline{18.82} & \underline{0.834} & \underline{98.95} \\
KITTI-360 & LiDM seeded & Distillation-DPO & 16.73 & 0.835 & 99.14 \\
KITTI-360 & LiDM seeded & LiDiff & 17.49 & 0.846 & 99.71 \\
KITTI-360 & LiDM seeded & LiDPM & 18.44 & 0.846 & 99.71 \\
KITTI-360 & LiDM seeded & FPSGen & \textbf{30.04} & \textbf{0.781} & \textbf{85.93} \\
\bottomrule
\end{tabular}
\caption{Additional generation metrics using DCD. COV and 1-NNA are percentages. 1-NNA is best when closer to \(50\%\).}
\label{tab:generation_dcd_supp}
\end{table*}

On SemanticKITTI \cite{semantickitti}, FPSGen improves all DCD-side metrics over LiDiff \cite{lidiff}, LiDPM \cite{martyniuk2025lidpm}, and SemCity \cite{semcity2024}, consistent with the CD/EMD trends in Table~\ref{tab:generation_semkitti_supp}. On KITTI-360 \cite{kitti360}, LiDPM obtains higher DCD coverage in the unconditional setting, while FPSGen obtains the best DCD MMD and the 1-NNA value closest to the ideal \(50\%\). SemCity is less competitive after converting its structured semantic output to voxel-center point clouds under this evaluation protocol. This representation mismatch should be considered when interpreting point-level distances, especially on KITTI-360 transfer.

Under the KITTI-360 LiDM seeded protocol, FPSGen improves DCD coverage, DCD MMD, and DCD 1-NNA over the available completion baselines, indicating that the BEV-supported point transport better preserves balanced scene density after LiDM sparse scan seeding. We use LiDM \cite{Lidardiffusioncvpr2024} as an established scene level LiDAR generation baseline. Its sparse scan output provides a common input for testing how different completion pipelines expand generated observations into full scenes.

\subsection{Initialization Noise and DCD}
\label{subsec:init_noise_dcd_supp}

Table~\ref{tab:init_noise_dcd_supp} reports the completion-side diagnostic for different initialization modes. The BEV source sampler denotes samples drawn from \(\mathcal{R}(\bar{B};N,\Sigma)\) as defined in the main paper. The LiDAR-derived initialization reduces CD because it starts from a scan-like support that is already close to the target surface. However, its DCD \cite{wu2021densityaware} is consistently worse than the BEV source sampler at matched step counts, showing that lower CD does not necessarily imply better density allocation. This supports our use of DCD as a complementary diagnostic. The BEV source sampler better preserves balanced scene density, while the LiDAR-derived initialization can concentrate points excessively around scan like structures.

\begin{table*}[t]
\centering
\setlength{\tabcolsep}{6.0pt}
\begin{tabular}{llcccc}
\toprule
Init. & \(K_P\) & CD \(\downarrow\) & DCD \(\downarrow\) & \(JSD_{3D}\) \(\downarrow\) & \(JSD_{BEV}\) \(\downarrow\) \\
\midrule
BEV source sampler & 1 & \underline{0.3209} & \textbf{0.5933} & \textbf{0.4925} & \textbf{0.3272} \\
LiDAR-derived & 1 & \textbf{0.2699} & \underline{0.6375} & \underline{0.5144} & \underline{0.3442} \\
Gaussian & 1 & 0.4899 & 0.7736 & 0.6360 & 0.4407 \\
\midrule
BEV source sampler & 8 & \underline{0.3213} & \textbf{0.6001} & \textbf{0.5006} & \textbf{0.3320} \\
LiDAR-derived & 8 & \textbf{0.2561} & \underline{0.6343} & \underline{0.5098} & \underline{0.3444} \\
Gaussian & 8 & 0.3607 & 0.7807 & 0.6407 & 0.4240 \\
\midrule
BEV source sampler & 32 & \underline{0.3162} & \textbf{0.5878} & \textbf{0.4924} & \textbf{0.3292} \\
LiDAR-derived & 32 & \textbf{0.2562} & \underline{0.6353} & \underline{0.5114} & \underline{0.3449} \\
Gaussian & 32 & 0.3337 & 0.7651 & 0.6299 & 0.4138 \\
\bottomrule
\end{tabular}
\caption{Initialization noise diagnostic for SemanticKITTI completion. Lower is better for CD, DCD, \(JSD_{3D}\), and \(JSD_{BEV}\).}
\label{tab:init_noise_dcd_supp}
\end{table*}

\subsection{Generation BEV Flow Step Ablation}
\label{subsec:generation_bev_step_supp}

Table~\ref{tab:generation_bev_step_supp} reports the generation side ablation of BEV flow sampling steps \(K_B\) while keeping the point flow step count fixed at \(K_P=32\). This experiment is different from the completion side BEV step diagnostic in Table~\ref{tab:bev_step_supp}. Here the entire LiDAR-free generation pipeline is evaluated with distributional metrics \cite{achlioptas2018learning,yang2019pointflow} on the fixed generation subsets.

\begin{table*}[t]
\centering
\setlength{\tabcolsep}{3.0pt}
\begin{tabular}{llcccccc}
\toprule
Dataset & \(K_B\) & COV-CD \(\uparrow\) & MMD-CD \(\downarrow\) & 1-NNA-CD  & COV-EMD \(\uparrow\) & MMD-EMD \(\downarrow\) & 1-NNA-EMD \\
\midrule
SemanticKITTI & 10 & 19.61 & 7.637 & 87.50 & 19.85 & 7.694 & 92.03 \\
SemanticKITTI & 30 & \underline{36.52} & \underline{6.834} & \underline{83.95} & \underline{38.24} & \underline{6.364} & \underline{82.60} \\
SemanticKITTI & 50 & \textbf{41.67} & \textbf{6.517} & \textbf{79.04} & \textbf{45.59} & \textbf{6.132} & \textbf{79.53} \\
\midrule
KITTI-360 & 10 & 17.68 & 8.806 & 81.27 & 15.02 & 6.998 & 91.25 \\
KITTI-360 & 30 & \underline{35.17} & \underline{7.850} & \underline{76.71} & \underline{37.64} & \underline{6.045} & \underline{80.13} \\
KITTI-360 & 50 & \textbf{41.25} & \textbf{7.480} & \textbf{75.29} & \textbf{44.30} & \textbf{5.847} & \textbf{77.09} \\
\bottomrule
\end{tabular}
\caption{Generation ablation over BEV flow steps \(K_B\) with \(K_P=32\). COV and 1-NNA are percentages. 1-NNA is best when closer to \(50\%\).}
\label{tab:generation_bev_step_supp}
\end{table*}

Increasing \(K_B\) from 10 to 50 consistently improves generation quality on both datasets. The effect is especially visible in coverage and 1-NNA, suggesting that the BEV prior needs enough integration steps to form a reliable global scene support before point level transport. On KITTI-360, COV-CD increases from 17.68 to 41.25 and 1-NNA-EMD improves from 91.25 to 77.09. On SemanticKITTI, COV-CD increases from 19.61 to 41.67 and 1-NNA-EMD improves from 92.03 to 79.53.

\subsection{Generation Point Flow Step Ablation}
\label{subsec:generation_point_step_supp}

Table~\ref{tab:generation_point_step_supp} reports the SemanticKITTI \cite{semantickitti} generation-side point-flow step ablation using the same COV, MMD, and 1-NNA protocol \cite{achlioptas2018learning,yang2019pointflow}. In this experiment, the BEV-flow step count is fixed at \(K_B=50\), and only the point flow integration steps \(K_P\) are varied in \(\{1,4,16,32,50\}\). This isolates how much multi-step point transport is needed after the BEV prior has already formed the global scene support.

\begin{table*}[t]
\centering
\setlength{\tabcolsep}{3.0pt}
\begin{tabular}{llcccccc}
\toprule
Dataset & \(K_P\) & COV-CD \(\uparrow\) & MMD-CD \(\downarrow\) & 1-NNA-CD & COV-EMD \(\uparrow\) & MMD-EMD \(\downarrow\) & 1-NNA-EMD \\
\midrule
SemanticKITTI & 1 & \underline{41.42} & 6.587 & 80.88 & \textbf{45.59} & \underline{6.132} & \underline{80.02} \\
SemanticKITTI & 4 & \underline{41.42} & 6.595 & 80.51 & \textbf{45.59} & 6.134 & 80.15 \\
SemanticKITTI & 16 & \textbf{41.67} & 6.529 & 79.29 & \textbf{45.59} & \textbf{6.131} & \textbf{79.53} \\
SemanticKITTI & 32 & \textbf{41.67} & \underline{6.517} & \textbf{79.04} & \textbf{45.59} & \underline{6.132} & \textbf{79.53} \\
SemanticKITTI & 50 & \textbf{41.67} & \textbf{6.512} & \underline{79.17} & \textbf{45.59} & \underline{6.132} & \textbf{79.53} \\
\bottomrule
\end{tabular}
\caption{Generation ablation over point flow steps \(K_P\) with \(K_B=50\) on SemanticKITTI. COV and 1-NNA are percentages. 1-NNA is best when closer to \(50\%\).}
\label{tab:generation_point_step_supp}
\end{table*}

This ablation complements Table~\ref{tab:generation_bev_step_supp}. The BEV step ablation studies whether the global support prior is sufficiently sampled, whereas the point-step ablation studies whether the point velocity field benefits from finer integration after the source points are already initialized from the generated BEV prior. Most coverage metrics saturate early once the BEV prior is fixed at 50 steps, while MMD-CD changes only slightly and reaches its lowest value at \(K_P=50\). This suggests that the BEV stage dominates global support formation, and additional point flow steps mainly improve distributional alignment rather than changing the covered support.

\subsection{Completion BEV Flow Step Ablation}
\label{subsec:completion_bev_flow_supp}

Table~\ref{tab:bev_step_supp} studies the number of BEV-flow steps \(K_B\) while keeping the point-flow step count fixed at \(K_P=32\). With one BEV step, CD, DCD, and \(JSD_{3D}\) are substantially worse than at larger step counts. Increasing \(K_B\) to 5--10 sharply improves these metrics. Further increasing \(K_B\) to 25 or 50 yields only marginal changes. We therefore use a moderate BEV step count in the default configuration to balance generation quality and sampling cost.

\begin{table}[t]
\centering
\setlength{\tabcolsep}{5.5pt}
\begin{tabular}{ccccc}
\toprule
\(K_B\) & CD \(\downarrow\) & DCD \(\downarrow\) & \(JSD_{3D}\) \(\downarrow\) & \(JSD_{BEV}\) \(\downarrow\) \\
\midrule
1 & 0.4397 & 0.6694 & 0.5966 & 0.3319 \\
5 & \textbf{0.3159} & 0.5889 & 0.4941 & \textbf{0.3264} \\
10 & \underline{0.3162} & \textbf{0.5878} & \textbf{0.4924} & \underline{0.3292} \\
25 & 0.3196 & \underline{0.5887} & \underline{0.4931} & 0.3329 \\
50 & 0.3219 & 0.5897 & 0.4941 & 0.3344 \\
\bottomrule
\end{tabular}
\caption{SemanticKITTI completion ablation over BEV flow steps with \(K_P=32\). Lower is better.}
\label{tab:bev_step_supp}
\end{table}

\subsection{Completion Point Flow Step Ablation}
\label{subsec:completion_point_flow_supp}

\begin{table}[t]
\centering
\setlength{\tabcolsep}{5.5pt}
\begin{tabular}{cccc}
\toprule
\(K_P\) & CD \(\downarrow\) & \(JSD_{3D}\) \(\downarrow\) & \(JSD_{BEV}\) \(\downarrow\) \\
\midrule
1  & 0.3207 & 0.4929 & \textbf{0.3275} \\
2  & 0.3263 & 0.5039 & 0.3316 \\
4  & 0.3271 & 0.5090 & 0.3336 \\
8  & 0.3209 & 0.5009 & 0.3320 \\
16 & 0.3172 & 0.4945 & 0.3303 \\
32 & \textbf{0.3161} & \underline{0.4924} & 0.3292 \\
50 & \underline{0.3162} & \textbf{0.4922} & \underline{0.3287} \\
\bottomrule
\end{tabular}
\caption{SemanticKITTI completion ablation over point flow steps under the default Approx OT teacher estimated coupling. Lower is better.}
\label{tab:forward_step_supp}
\end{table}

Table~\ref{tab:forward_step_supp} reports the completion-side point flow step ablation for the default Approx OT teacher-estimated coupling. The one step sampler is already strong because the BEV-supported point source and teacher-estimated clean endpoint define a simple local endpoint map. More point steps are not monotonically better at small counts, but 16--50 steps recover stable CD and JSD values as the sampler better resolves the full transport path.

\section{Implementation Details}
\label{sec:implementation}

To support reproducibility, we will release the training code, evaluation scripts, and configuration files upon acceptance.

\subsection{Dataset Processing}
\label{subsec:dataset_processing}

\textbf{Training split.}
FPSGen is trained once on SemanticKITTI \cite{semantickitti} sequences 00--07 and 09--10, following the split used by prior LiDAR completion work such as LiDiff and LiDPM \cite{lidiff,martyniuk2025lidpm}. This training split contains \(19{,}130\) samples. The BEV flow, teacher, and student point flow are all trained on this split, and the same trained model is used for SemanticKITTI completion, SemanticKITTI generation, and cross-dataset evaluation on KITTI-360 \cite{kitti360}. No component is retrained on sequence 08 or KITTI-360.

\textbf{Completion evaluation.}
SemanticKITTI sequence 08 contains \(4{,}071\) test samples \cite{semantickitti}. For LiDAR conditioned completion, the raw LiDAR scan is the sparse input, and the complete target is prepared by aggregating neighboring scans into a dense scene under the same completion protocol as prior LiDAR completion work \cite{lidiff,martyniuk2025lidpm}. To make the comparison with LiDiff and related completion baselines fair, we keep the same range cropping, coordinate convention, and evaluation preparation as the LiDiff setting. The full completion comparison evaluates all frames in sequence 08. The KITTI-360 validation set contains \(10{,}515\) samples \cite{kitti360}, and we use it as a cross-dataset evaluation setting under the same metric definitions. For completion ablations on SemanticKITTI and KITTI-360, we evaluate one frame every 10 frames. Adjacent LiDAR scans in driving sequences are highly similar, and the 10 frame interval provides a representative estimate while reducing repeated evaluation on near-duplicate scans.

\textbf{Generation evaluation.}
Evaluation is performed on fixed held-out subsets to avoid adjacent frame redundancy. On SemanticKITTI \cite{semantickitti} sequence 08, we select one frame every 10 frames, yielding 408 reference samples. We use this subset for the in-domain generation comparison and the SemanticKITTI generation ablations.

On KITTI-360 \cite{kitti360}, we select one frame every 20 frames, yielding 526 reference samples, and use this as the cross-dataset generation evaluation. All generation metrics use the same spatial range and evaluation configuration. For the LiDiff and LiDPM unconditional diagnostics, we set the condition state to empty and replace their partial-scan repetition source with a LiDAR-free BEV Gaussian source. Their trained denoisers and point sampling schedules remain unchanged. The SemanticKITTI generation comparison uses these matched LiDAR-free logs together with FPSGen and the SemCity baseline processed as described below. The KITTI-360 comparison uses the corresponding formal KITTI-360 logs under the same protocol.

In the KITTI-360 LiDM seeded diagnostic, LiDM \cite{Lidardiffusioncvpr2024} first generates sparse LiDAR scans, and completion baselines complete these scans into full point clouds over the same 526 frame KITTI-360 subset.

\textbf{SemCity baseline processing.}
For SemCity \cite{semcity2024}, we follow its triplane autoencoder and diffusion training pipeline. The autoencoder is trained first, triplane features are extracted with the last numerically stable checkpoint, and the diffusion model is then trained on the saved triplane representation. SemCity natively generates a local semantic occupancy scene of approximately \(51.2\mathrm{m}\times51.2\mathrm{m}\), corresponding to a \(256\times256\times32\) voxel grid with \(0.2\mathrm{m}\) voxel size. To match our \(100\mathrm{m}\times100\mathrm{m}\) evaluation region, we first generate a seed scene, expand it to about \(102.4\mathrm{m}\times102.4\mathrm{m}\) using the model's outpainting capability in multiple directions, and then center-crop the result to \(100\mathrm{m}\times100\mathrm{m}\). The final sample keeps only voxels inside a radius of \(50\mathrm{m}\) around the scene origin.

We convert SemCity semantic occupancy outputs to point clouds by using voxel centers. The horizontal voxel size is \(0.2\mathrm{m}\). For the vertical coordinate, directly using the raw voxel index gives a poor alignment with SemanticKITTI style point clouds, so we apply a vertical offset and use \(z=0.2\,i_z-2.0\), where \(i_z\) is the vertical voxel index. This substantially improves the generated \(z\) distribution, although the final point cloud metrics remain weak. Since SemCity outputs voxel-center point clouds while the references are raw LiDAR point clouds, CD, EMD, DCD, and related point metrics can be harsher than semantic occupancy quality alone would suggest. KITTI-360 results should be interpreted as cross-dataset transfer, because SemCity is primarily designed for SemanticKITTI style scenes.

\textbf{Condition protocol.}
We separate conditions by their strength and intended task. The sparse LiDAR scan is a strong geometric condition because it directly anchors observed surfaces and is therefore used for completion and completion side ablations. Road and vehicle masks are weaker layout conditions because they constrain scene support, drivable regions, and object locations, but do not specify complete 3D geometry. We therefore use them to evaluate flexible generation without LiDAR. During BEV flow and point flow training, each sample independently draws one of the eight possible mask states with equal probability, i.e., each LiDAR/vehicle/road combination has probability \(12.5\%\). Any inactive condition is replaced by a zero tensor. In the notation \(C_m=(m_lc_l,m_vc_v,m_rc_r)\), completion ablations compare LiDAR-active settings with or without auxiliary layout cues, while generation ablations evaluate LiDAR-free settings with road, vehicle, or joint layout cues. This protocol avoids conflating LiDAR conditioned completion with LiDAR free scene synthesis while ensuring that both networks see unconditional, layout-only, LiDAR-only, and mixed condition cases during training.
\begin{equation}
(m_l,m_v,m_r)\sim
\operatorname{Unif}
\!\left(\{0,1\}^{3}\right).
\label{eq:condition_sampling}
\end{equation}

\textbf{Layout mask construction.} The layout conditions \(c_v\) and \(c_r\) are derived from ground-truth semantic labels and are used only as coarse BEV constraints. For each scene, we first select points whose semantic labels belong to the target layout group, discard their \(z\)-coordinates, and project their \(xy\)-coordinates onto the same \(256\times256\) BEV grid used by the BEV flow. A grid cell is marked as occupied for a layout channel if at least one selected point falls into that cell; otherwise, it is set to zero. This gives a multi-channel binary layout tensor \(c_{\mathrm{layout}}\in\{0,1\}^{C\times256\times256}\), where the default channels correspond to vehicle and road/ground.

\begin{equation}
c_k(q)=\mathbb{I}\!\left[\exists i:\ell_i\in\mathcal{Y}_k,\;\Pi(x_i,y_i)=q\right],\qquad k\in\{v,r\}.
\label{eq:layout_mask_construction}
\end{equation}

Here, \(\mathbb{I}[\cdot]\) denotes the indicator function, which equals \(1\) when the condition inside the brackets is true and \(0\) otherwise. The variable \(\ell_i\) is the semantic label of point \(i\), \(\mathcal{Y}_k\) is the label set for layout class \(k\), and \(\Pi\) maps metric \(xy\)-coordinates within a \(50\,\mathrm{m}\) radius of the scene origin to a BEV cell \(q\). For SemanticKITTI \cite{semantickitti}, the vehicle channel includes moving and static vehicle-related labels, while the road/ground channel includes road, other ground, sidewalk, and parking labels. For KITTI-360 \cite{kitti360}, the vehicle channel includes car, truck, bus, caravan, trailer, and train labels, while the road/ground channel includes road, sidewalk, and parking. During network input preparation, inactive layout channels are replaced by zeros according to the sampled mask variables, while active layout masks are linearly mapped to the same \([-1,1]\) range as the BEV target channels.

\textbf{Guidance scales.}
We apply classifier-free guidance (CFG) \cite{ho2022classifierfree} to both flow stages. For all non-empty condition settings, including LiDAR-conditioned completion, layout conditioned generation, and mixed condition generation, we use scales \(s_B=s_P=2\) for the BEV flow and point flow. For unconditional generation, both scales are set to \(0\), so sampling follows the unconditional branch without amplifying a nonexistent conditioning signal.

\textbf{Integration step notation.}
Throughout the paper, \(K_B\) and \(K_P\) denote the number of forward Euler steps used by the BEV flow and point flow, respectively. This stage-specific notation avoids ambiguity in the ablation and runtime comparisons.

\textbf{BEV target normalization.}
For BEV flow training, each complete point cloud is rasterized into density, maximum-height, and occupancy mask channels, following the common use of BEV representations for autonomous driving perception \cite{PointPillars}. Density is computed by counting points in each BEV cell and applying log normalization after clipping by the maximum density. The height channel stores the maximum \(z\) value in each occupied cell, rather than the mean height. Empty cells are assigned height \(-1\), and the occupancy mask is mapped from \(\{0,1\}\) to \(\{-1,1\}\). Consequently, all three BEV target channels are in the \([-1,1]\) range used by flow matching. During sampling with the BEV source sampler \(\mathcal{R}\), sampled grid cells are converted back to physical \(xy\) coordinates in meters, while the BEV tensor provided to the point flow remains in the normalized \([-1,1]\) scale.
For a BEV cell \(q\), let \(n(q)\) be its point count, \(z^{\max}(q)\) be its maximum height, and \(m(q)=\mathbf{1}[n(q)>0]\) be its occupancy indicator. With density clipping constant \(n_{\max}\), minimum height \(z_{\min}\), and height range \(\Delta z\), the target channels are
\begin{equation}
\begin{aligned}
D(q) &= 2\frac{\log\!\left(1+\min\{n(q),n_{\max}\}\right)}{\log(1+n_{\max})}-1,\\
H(q) &= m(q)\left(2\frac{z^{\max}(q)-z_{\min}}{\Delta z}-1\right)-\left(1-m(q)\right),\\
M(q) &= 2m(q)-1.
\end{aligned}
\label{eq:bev_target_norm}
\end{equation}

\subsection{Network Details}
\label{subsec:network_details}

\begin{figure*}[t]
\centering
\includegraphics[width=\linewidth]{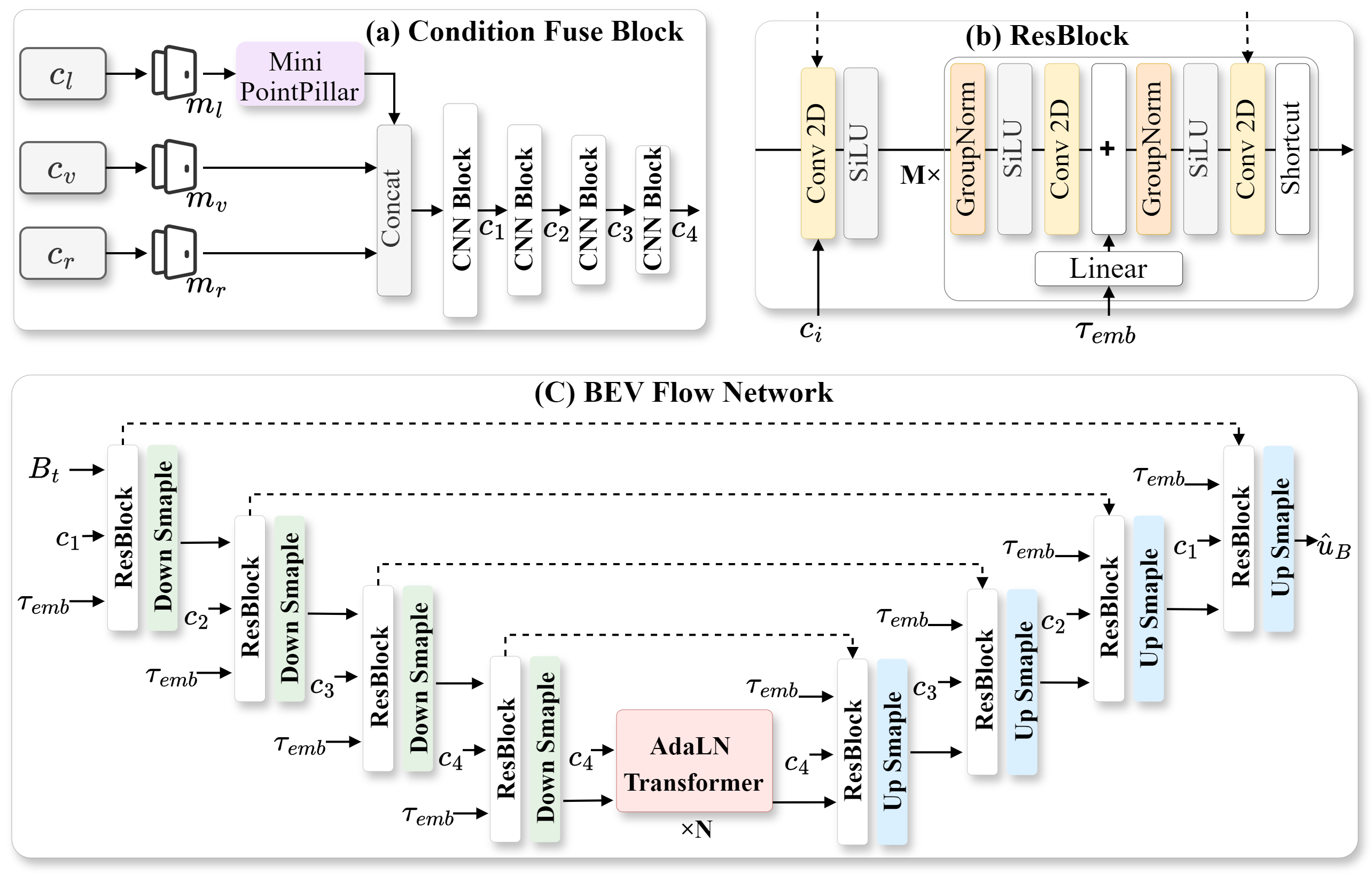}
\caption{Architecture of the BEV flow model. (a) The condition fusion block converts LiDAR and layout inputs into multi-scale BEV condition features. (b) The ResBlock used in the U-Net is a stage-level conditional block that first fuses the scale-matched condition feature by channel concatenation and \(1\times1\) projection, then applies time-conditioned residual convolutional updates. (c) The BEV flow pipeline stacks these conditional ResBlocks with downsampling, upsampling, skip connections, and spatial AdaLN transformer blocks to predict the BEV velocity field.}
\label{fig:bev_flow_architecture}
\end{figure*}

Both network stages are implemented as conditional flow matching models, following the simulation-free velocity regression formulation of flow matching and rectified flows \cite{lipman2022flowmatching,liu2022rectifiedflow}. Architecturally, the BEV flow follows the image generation practice of combining convolutional U-Nets, time-conditioned residual blocks, and attention-based bottlenecks \cite{ronneberger2015unet,ho2020denoising,vaswani2017attention,peebles2023scalable}. Unlike latent diffusion models that first compress images into a learned VAE latent space \cite{rombach2022high}, our BEV representation is a low-resolution \(256\times256\) tensor with density, height, and mask channels. We therefore generate directly in BEV pixel space without introducing an additional learned reconstruction objective.

\textbf{BEV flow model.}
The main paper defines the BEV flow objective and its role in the two-stage sampler. Here we specify the implementation used to predict \(\hat{u}_B=v_{\phi}(B_{\tau},\tau,C_m)\). In this expression, \(B_{\tau}\) is the intermediate BEV state at flow time \(\tau\), and \(v_{\phi}\) is the BEV velocity network. The network receives the active condition tuple \(C_m\), builds multi-scale BEV condition features, and predicts the velocity field of the density, height, and occupancy-mask channels.

Fig.~\ref{fig:bev_flow_architecture} summarizes the BEV flow model. We use three components, including a condition fusion block that builds multi-scale BEV features, a time-conditioned convolutional U-Net that preserves local spatial detail, and an adaptive Layer Normalization (AdaLN) transformer bottleneck that models long-range interaction across road and object regions. The condition fusion block first maps the optional LiDAR condition into a \(256\times256\) pseudo-image with a Mini-PointPillar encoder, inspired by the BEV pillar representation used in PointPillars \cite{PointPillars}. The encoder uses EdgeConv style \(k\)-nearest neighbor (kNN) aggregation \cite{wang2019dynamic} with \(k=16\) local neighbors and outputs a 32-channel BEV feature aligned with the layout masks. The masks \(m_l,m_v,m_r\) are applied before this fusion. At training time each mask triplet is sampled with probability \(12.5\%\), and inactive branches are replaced by zeros. Therefore the same BEV architecture learns unconditional, LiDAR-only, layout-only, and mixed-condition priors without changing its input interface.
\begin{equation}
\begin{gathered}
e_{ij}=[p_j-p_i,p_i],
\qquad j\in\mathcal{N}_k(i),\\
\ell_i=\max_{j\in\mathcal{N}_k(i)}
\operatorname{MLP}_{e}(e_{ij}),\\
a_i=\left[
p_i,\;
p_i-\bar{p}_{q_i},\;
(x_i,y_i)-g(q_i),\;
\ell_i
\right],\\
u_i=\operatorname{PFN}(a_i),\\
E_{\mathrm{pc}}(\mathcal{P}^{l})_{q,c}
=\max_{i:q_i=q} u_{i,c}.
\end{gathered}
\label{eq:minipointpillar_impl}
\end{equation}
Here \(p_i=(x_i,y_i,z_i)\) is a LiDAR point, \(\mathcal{N}_k(i)\) is its \(k\)-nearest neighbor set, \(e_{ij}\) is an edge feature, and \(\ell_i\) is the local geometry feature obtained by max pooling over neighboring features. The grid index \(q_i\) is obtained by projecting \((x_i,y_i)\) onto the BEV grid, \(\bar{p}_{q_i}\) is the mean point in the same pillar, and \(g(q_i)\) is the metric center of that BEV cell. The augmented point feature \(a_i\) combines absolute coordinates, pillar-relative offsets, cell-center offsets, and kNN geometry. A point feature network (PFN) maps \(a_i\) to \(u_i\), and dynamic max scatter over points in each grid cell forms the LiDAR BEV feature \(E_{\mathrm{pc}}(\mathcal{P}^{l})\). This Mini-PointPillar encoder keeps the projection lightweight while preserving local 3D neighborhood information that would be lost by simple point counting.

The LiDAR BEV feature is then concatenated with the active layout channels and passed through a four-level convolutional encoder. We denote its outputs by \(F_1,\ldots,F_4\), corresponding to the \(c_1,\ldots,c_4\) features in Fig.~\ref{fig:bev_flow_architecture}.
\begin{equation}
\begin{gathered}
F_0=\operatorname{concat}
\!\left(m_lE_{\mathrm{pc}}(\mathcal{P}^{l}),m_vc_v,m_rc_r\right),\\
F_i=\operatorname{CNN}_i(F_{i-1}),\\
\operatorname{CNN}_i(x)=
\mathrm{Conv}_{3\times3}\!\left(
\mathrm{SiLU}\!\left(\mathrm{GN}(h_i)\right)\right),\\
h_i=\mathrm{Conv}_{3\times3}^{s_i}(x).
\end{gathered}
\label{eq:bev_condition_fpn_impl}
\end{equation}
Here \(F_0\) is the concatenated condition map, \(F_i\) is the \(i\)-th multi-scale condition feature, \(E_{\mathrm{pc}}\) denotes the Mini-PointPillar encoder, \(h_i\) is an intermediate feature inside the convolutional neural network (CNN) block, \(s_1=1\), and \(s_i=2\) for \(i>1\). \(\mathrm{Conv}\), \(\mathrm{GN}\), and \(\mathrm{SiLU}\) denote convolution, Group Normalization, and the SiLU activation function. When LiDAR is unavailable, the LiDAR branch is inactive according to \(C_m\), and the condition encoder uses the available layout channels. The feature hierarchy has channel widths \([32,64,128,256]\) and resolutions \(256^2,128^2,64^2,\) and \(32^2\). This hierarchy matches the different spatial roles of the conditions. Road masks provide global drivable support, vehicle masks provide localized object priors, and LiDAR features provide measured geometric anchors.

The flow time \(\tau\) is encoded before being injected into the residual and transformer blocks. We use a sinusoidal embedding with frequencies on the standard \(10000\) based log scale, following the positional encoding convention used in transformer models \cite{vaswani2017attention}, after scaling \(\tau\) to the training time interval.
\begin{equation}
\begin{gathered}
\gamma(\tau)=
\left[\sin(1000\tau\omega_k),\cos(1000\tau\omega_k)\right]_{k=0}^{127},\\
\omega_k=\exp\!\left(-\frac{k\log 10000}{127}\right),\\
\tau_{\mathrm{emb}}=W_2\,\mathrm{SiLU}\!\left(W_1\gamma(\tau)\right).
\end{gathered}
\label{eq:bev_time_embedding_impl}
\end{equation}
Here \(\gamma(\tau)\) is the sinusoidal time embedding, \(\omega_k\) is the \(k\)-th frequency, and \(W_1,W_2\) are learned linear projections. The projected 256-dimensional vector \(\tau_{\mathrm{emb}}\) is the shared temporal code used by all BEV residual blocks and the AdaLN transformer bottleneck.

The BEV velocity predictor is a U-Net style encoder--decoder \cite{ronneberger2015unet}. The noisy BEV state \(B_{\tau}\) is first projected by an input convolution. In Fig.~\ref{fig:bev_flow_architecture}, each ResBlock denotes a stage-level conditional residual block. It contains a condition fusion layer and two time-conditioned residual updates. Let \(\operatorname{Fuse}_i\) denote the condition fusion layer at scale \(i\). It concatenates the current U-Net feature \(h\) with the scale-matched condition feature \(F_i\) along the channel dimension, then projects the concatenated feature back to the U-Net feature width.
\begin{equation}
\operatorname{Fuse}_i(h,F_i)=
\mathrm{SiLU}\!\left(\mathrm{Conv}_{1\times1}\!\left(\operatorname{concat}(h,F_i)\right)\right),
\label{eq:bev_spatial_fusion_impl}
\end{equation}
The fused feature is then processed by time-conditioned residual units. For a feature map \(x\), each residual unit is implemented as
\begin{equation}
\begin{gathered}
h_1=\mathrm{Conv}_1\!\left(\mathrm{SiLU}(\mathrm{GN}(x))\right),\\
h_2=h_1+W_{\tau}(\tau_{\mathrm{emb}}),\\
\operatorname{Res}(x,\tau_{\mathrm{emb}})
=\mathrm{Conv}_2\!\left(\mathrm{SiLU}(\mathrm{GN}(h_2))\right)
+\operatorname{Shortcut}(x).
\end{gathered}
\label{eq:bev_resblock_impl}
\end{equation}
Here \(W_{\tau}\) is a learned linear projection from the time embedding to the block width, and \(\operatorname{Shortcut}\) is either the identity mapping or a \(1\times1\) convolution when the channel width changes. Thus, a stage-level ResBlock in the figure can be written compactly as
\begin{equation}
\begin{gathered}
r_{i,1}=\operatorname{Res}_{i,1}\!\left(\operatorname{Fuse}_i(h,F_i),\tau_{\mathrm{emb}}\right),\\
r_{i,2}=\operatorname{Res}_{i,2}\!\left(r_{i,1},\tau_{\mathrm{emb}}\right).
\end{gathered}
\label{eq:bev_stage_resblock_impl}
\end{equation}
The encoder uses channel widths \([32,64,128,256]\). Both \(r_{i,1}\) and \(r_{i,2}\) are stored as down to up skip features at the same resolution. During decoding, the feature at the corresponding resolution is concatenated with these stored skip features before the two residual updates.
\begin{equation}
\begin{gathered}
\tilde{h}_{i,1}=\operatorname{concat}(h_i^{\uparrow},r_{i,2}),\\
\tilde{r}_{i,1}=\operatorname{Res}_{i,1}^{\uparrow}
\!\left(\operatorname{Fuse}_i(\tilde{h}_{i,1},F_i),\tau_{\mathrm{emb}}\right),\\
\tilde{h}_{i,2}=\operatorname{concat}(\tilde{r}_{i,1},r_{i,1}),\\
\tilde{r}_{i,2}=\operatorname{Res}_{i,2}^{\uparrow}
\!\left(\tilde{h}_{i,2},\tau_{\mathrm{emb}}\right).
\end{gathered}
\label{eq:bev_skip_impl}
\end{equation}
Here \(h_i^{\uparrow}\) is the decoder feature at the matching resolution, and \(\tilde{r}_{i,2}\) is followed by nearest neighbor upsampling and a convolution except at the final resolution. This explicitly links the downsampling and upsampling paths and preserves local BEV detail while the bottleneck handles long range interaction.

At the \(32\times32\) bottleneck, we add a learned 2D positional embedding and apply eight spatial AdaLN transformer blocks. This design follows the transformer attention mechanism \cite{vaswani2017attention} and the AdaLN style conditioning used in diffusion transformers \cite{peebles2023scalable}, while keeping the surrounding encoder--decoder convolutional structure for local BEV detail. Let \(z\) be the flattened bottleneck feature with size \(N_b\times L_b\times C_b\), where \(N_b\), \(L_b\), and \(C_b\) denote batch size, spatial token length, and channel width. Let \(z_c\) be the flattened condition feature from \(F_4\). A time multilayer perceptron (MLP) predicts adaptive Layer Normalization (LN) parameters and residual gates.
\begin{equation}
(\delta_a,\alpha_a,g_a,\delta_m,\alpha_m,g_m)
=\operatorname{chunk}\!\left(W_d\,\mathrm{SiLU}(\tau_{\mathrm{emb}})\right).
\label{eq:bev_adaln_params_impl}
\end{equation}
Here \(W_d\) is a learned linear projection, \(\operatorname{chunk}\) splits its output into six equal width vectors, \(\delta\), \(\alpha\), and \(g\) denote shift, scale, and gate parameters, and the subscripts \(a\) and \(m\) correspond to the attention and MLP branches. Each AdaLN transformer block then performs multi-head self-attention (MSA), multi-head cross-attention (MCA), and an MLP update.
\begin{equation}
\begin{gathered}
\tilde{z}_a=\mathrm{LN}(z)(1+\alpha_a)+\delta_a,\\
z' = z+g_a\,\mathrm{MSA}(\tilde{z}_a),\\
z''=z'+\mathrm{MCA}\!\left(\mathrm{LN}(z'),z_c,z_c\right),\\
\tilde{z}_m=\mathrm{LN}(z'')(1+\alpha_m)+\delta_m,\\
z_{\mathrm{out}}=z''+g_m\,\mathrm{MLP}(\tilde{z}_m).
\end{gathered}
\label{eq:bev_adaln_transformer_impl}
\end{equation}
In Eq.~(\ref{eq:bev_adaln_transformer_impl}), \(\tilde{z}_a\) and \(\tilde{z}_m\) are the modulated attention and MLP inputs, and \(z'\), \(z''\), and \(z_{\mathrm{out}}\) denote the intermediate and output token features. We use eight attention heads and a four-times expansion ratio in the MLP. The AdaLN parameters make the bottleneck explicitly time-dependent, while cross-attention keeps the long range BEV reasoning anchored to the condition hierarchy.

At the model interface, the BEV velocity predictor uses the variables defined for the BEV flow.
\begin{equation}
\hat{u}_B=v_{\phi}\!\left(B_{\tau},\tau,C_m\right).
\label{eq:bev_net_impl}
\end{equation}
The internal features \(F_1,\ldots,F_4\) are not additional external conditions. They are implementation-level representations of \(C_m\) used by the condition fusion layers and AdaLN transformer blocks. The final convolutional head predicts the BEV velocity field, which is integrated to obtain the generated BEV prior \(\hat{B}\).

\textbf{Point flow network.}
The main paper defines the point flow with the training-time BEV prior \(\bar{B}\) and the inference-time generated BEV prior \(\hat{B}\). This subsection gives the deployed sparse U-Net and condition modulation details. We use a sparse point backbone similar in spirit to the completion networks used by LiDiff, LiDPM, and ScoreLiDAR \cite{lidiff,martyniuk2025lidpm,scorelidar}, and extend it with BEV and layout aware modulation so that the same point flow can operate under LiDAR, layout, mixed, and unconditional settings.

The backbone is an instance-normalized sparse 3D U-Net. Sparse convolution is used because outdoor driving scenes are mostly empty in 3D, making dense volumetric convolution inefficient. This follows the sparse convolutional design commonly used for large-scale 3D point scenes \cite{choy20194d}. The input point state \(\mathcal{P}_t\) is voxelized into a sparse tensor \(S_0\), then processed by a sparse convolutional stem.
\begin{equation}
S_0=\operatorname{Stem}(\mathcal{P}_t).
\label{eq:point_stem_impl}
\end{equation}
The encoder has four downsampling stages, and the decoder has four transposed convolution upsampling stages with skip connections. We use sparse residual blocks with instance normalization in both directions. For an encoder level \(l\), with the stage-wise gate \(w_l\) defined below, the sparse feature update is
\begin{equation}
\begin{gathered}
\tilde{S}_l=S_l\odot w_l,\\
S_{l+1}=\operatorname{Down}_l(\tilde{S}_l).
\end{gathered}
\label{eq:point_encoder_impl}
\end{equation}
For the decoder, the upsampled sparse feature is concatenated with the encoder feature at the matching resolution.
\begin{equation}
\begin{gathered}
\tilde{Y}_l=Y_l\odot w_l^{\uparrow},\\
Y_{l-1}^{\uparrow}=\operatorname{Up}_l(\tilde{Y}_l),\\
Y_{l-1}=\operatorname{Res}_l^{\uparrow}
\!\left(Y_{l-1}^{\uparrow}\oplus S_{l-1}\right).
\end{gathered}
\label{eq:point_decoder_impl}
\end{equation}
Here \(S_l\) and \(Y_l\) are encoder and decoder sparse tensors, \(w_l^{\uparrow}\) is the decoder-side gate at level \(l\), \(\odot\) denotes channel-wise multiplication on sparse features, and \(\oplus\) denotes sparse tensor concatenation at matching coordinates. The channel layout is a reduced-width variant of the LiDiff/LiDPM sparse U-Net family \cite{lidiff,martyniuk2025lidpm}. While the original completion backbones commonly use \([32,32,64,128,256,256,128,96,96]\), our student point flow uses \([16,16,32,64,128,64,64,48,48]\). Thus the widest bottleneck is reduced from 256 to 128 channels and the decoder/output feature widths are also narrowed. This reduction is important because the student is evaluated repeatedly under different condition modes and sampling steps. A lighter sparse backbone keeps flexible generation practical while preserving the same encoder--decoder topology. The final multilayer perceptron (MLP) head maps \(48\rightarrow32\rightarrow3\) to predict a 3D velocity.

The main architectural difference from a LiDAR completion sparse denoiser is the stage-wise condition modulation. This modulation is driven by three condition streams, namely a sparse point anchor encoder, a BEV support encoder, and a layout mask encoder. The point anchor encoder processes the active sparse point condition. In LiDAR conditioned modes, the anchor is the sparse LiDAR scan. In LiDAR-free modes, the LiDAR branch is zeroed by the mask. During teacher training, the teacher uses the clean target point cloud to construct the teacher-estimated endpoint pairing, as described below. The student anchor encoder is a reduced width instance-normalized sparse encoder.
\begin{equation}
\begin{gathered}
S^{a}_0=\operatorname{Stem}(\mathcal{P}^{a}),\\
S^{a}_{i+1}=\operatorname{Enc}^{a}_i(S^{a}_i),
\qquad i=0,\ldots,3,\\
S^{a}=S^{a}_4 .
\end{gathered}
\label{eq:point_anchor_encoder_impl}
\end{equation}
Here \(\mathcal{P}^{a}\) is the active sparse point anchor after mask application, \(\operatorname{Stem}\) contains two sparse convolutions with instance-normalization and ReLU activations, and each \(\operatorname{Enc}^{a}_i\) contains a stride 2 sparse convolution block followed by two instance normalized residual blocks. The encoder uses the same reduced channel sequence \([16,16,32,64,128]\) as the point flow encoder, so the anchor representation is lightweight but spatially aligned with the sparse U-Net.

For compact implementation notation, let \(B_{\star}=\bar{B}\) during student training and \(B_{\star}=\hat{B}\) during inference. This alias is used only in this implementation description, and it refers to the ground truth BEV prior or the generated BEV prior defined in the main paper. We encode the density, height, and mask BEV tensor and the layout mask tensor with two lightweight feature pyramid encoders.
\begin{equation}
\begin{gathered}
G^{B}=E_B(B_{\star}),\\
G^{M}=E_M([m_vc_v,m_rc_r]).
\end{gathered}
\label{eq:point_bev_streams_impl}
\end{equation}
Both \(E_B\) and \(E_M\) are convolutional feature pyramid encoders with strides \((1,2,2,2)\). The main BEV stream \(G^{B}\) has 64 channels and carries density, height, and occupancy information. The mask stream \(G^{M}\) has 16 channels and carries road and vehicle layout cues. Each encoder first builds bottom-up features and then uses lateral \(1\times1\) projections with top down upsample add fusion.
\begin{equation}
\begin{gathered}
X^{B}=B_{\star},\qquad
X^{M}=[m_vc_v,m_rc_r],\\
C_0^{s}=\operatorname{ConvStage}_0^{s}(X^{s}),\\
C_i^{s}=\operatorname{ConvStage}_i^{s}(C_{i-1}^{s}),
\qquad i\in\{1,2,3\},\\
Q_3^{s}=\operatorname{Lat}_3^{s}(C_3^{s}),\\
Q_i^{s}=\operatorname{Smooth}_i^{s}
\!\left(\operatorname{Lat}_i^{s}(C_i^{s})
+\operatorname{Up}(Q_{i+1}^{s})\right),\\
i\in\{2,1,0\},\qquad s\in\{B,M\}.
\end{gathered}
\label{eq:point_bev_fpn_impl}
\end{equation}
Here \(X^s\) is the input tensor of stream \(s\), \(C_i^{s}\) is its bottom-up feature, \(Q_i^{s}\) is the corresponding top-down feature, \(\operatorname{Lat}_i^{s}\) is a lateral \(1\times1\) projection, and \(\operatorname{Up}\) denotes bilinear upsampling. The deployed point flow uses the full-resolution outputs \(G^{B}=Q_0^{B}\) and \(G^{M}=Q_0^{M}\). At every sparse U-Net stage, these shared maps are sampled at the current sparse coordinates. To avoid overloading the layout projection operator \(\Pi\), we denote this sparse-aligned BEV sampling operation by \(\Pi_{\mathrm{sp}}(S_l,G)\). Then
\begin{equation}
\begin{gathered}
b_l^{B}=\operatorname{Proj}_l^{B}
\!\left(\Pi_{\mathrm{sp}}(S_l,G^{B})\right),\\
b_l^{M}=\operatorname{Proj}_l^{M}
\!\left(\Pi_{\mathrm{sp}}(S_l,G^{M})\right).
\end{gathered}
\label{eq:point_bev_sample_impl}
\end{equation}
Here \(\operatorname{Proj}_l^{B}\) and \(\operatorname{Proj}_l^{M}\) are level-specific linear projections. The operator \(\Pi_{\mathrm{sp}}\) converts sparse voxel coordinates back to metric \(xy\) locations and indexes the corresponding BEV cell, so the BEV condition is spatially aligned with the active sparse support instead of being pooled into a global vector.

The point flow also uses a time embedding and a matched sparse anchor feature. It uses the same sinusoidal construction as the BEV flow with a separate 48-dimensional code that matches the sparse U-Net feature scale. The code is projected independently at each level.
\begin{equation}
\begin{gathered}
\gamma_P(t)=
\left[\sin(1000t\omega_k^P),\cos(1000t\omega_k^P)\right]_{k=0}^{23},\\
\omega_k^P=\exp\!\left(-\frac{k\log 10000}{23}\right),\\
e_l^{t}=T_l(\gamma_P(t)).
\end{gathered}
\label{eq:point_time_impl}
\end{equation}
Here \(\gamma_P(t)\) is the 48-dimensional point flow time code, \(\omega_k^P\) is its \(k\)-th frequency, and \(T_l\) is the learned projection at sparse U-Net level \(l\).
The matched anchor feature \(a_l\) is obtained by nearest neighbor matching between the current sparse coordinates and the sparse point anchor feature \(S^{a}\).
\begin{equation}
a_l=A_l\!\left(\operatorname{Match}(S_l,S^{a})\right).
\label{eq:point_anchor_impl}
\end{equation}
Here \(S^{a}\) denotes the sparse anchor tensor used by the point network, \(\operatorname{Match}\) returns the nearest anchor feature for each active sparse coordinate, and \(A_l\) is a level-specific projection.

The stage-wise gate is then predicted from four aligned signals, namely the matched sparse anchor, the transport time, the BEV support feature, and the layout feature.
\begin{equation}
\begin{gathered}
w_l=W_l\!\left([a_l,e_l^{t},b_l^{B},b_l^{M}]\right),\\
\tilde{S}_l=S_l\odot w_l.
\end{gathered}
\label{eq:point_gate_impl}
\end{equation}
Here \(W_l\) is the level-specific gating network. The same modulation form is applied at the four encoder levels and the four decoder levels. This gives every sparse convolutional stage access to condition information at the correct spatial support and temporal position. In compact form, the point velocity predictor is
\begin{equation}
\begin{gathered}
\hat{u}_P
=v_{\psi}(\mathcal{P}_t,t,B_{\star},C_m),\\
\hat{u}_P
=\operatorname{Head}\!\left(
\operatorname{SparseUNet}_{\psi}
(\mathcal{P}_t;\{w_l\}_{l})
\right).
\end{gathered}
\label{eq:point_net_impl}
\end{equation}

This formulation makes the role of the point flow complementary to the BEV flow. The sparse U-Net and its skip connections preserve local 3D geometry, the BEV streams provide global support and layout alignment, and the time embedding specifies the current location along the flow path. We use the instance-normalized variant because the point network sees heterogeneous condition mixtures within training batches. Instance normalization is less sensitive to batch composition than batch normalization in this setting. This is an implementation choice for flexible condition generation rather than a limitation of LiDiff or ScoreLiDAR, which remain strong LiDAR completion architectures.

\textbf{Teacher network.}
The teacher network is used only during training to construct the source-to-target supervision for the student point flow. Architecturally, it follows a LiDiff-style sparse network \cite{lidiff} but removes the diffusion time encoding branch. Because the teacher is not used at inference, we keep it as a higher capacity full width sparse network, while the deployed student uses the reduced width point flow described above. Given a noisy source point set and a complete target scene, the teacher estimates a direct displacement toward the target.

In teacher training, the complete point cloud is provided as the sparse anchor input rather than a partial scan. Let \(\mathcal{P}_{0}\) denote the noisy BEV-supported point source and \(\mathcal{P}^{gt}\) the complete target point cloud. Following the notation used throughout the paper, the teacher predicts a displacement field toward the clean endpoint.
\begin{equation}
\mathcal{P}^{\dagger}_{1}
=\mathcal{P}_{0}+\Gamma_{\eta}
\!\left(\mathcal{P}_{0},\mathcal{P}^{gt}\right).
\label{eq:teacher_mapping_impl}
\end{equation}
Here \(\Gamma_{\eta}\) is the teacher displacement network with parameters \(\eta\). The target side anchor \(\mathcal{P}^{gt}\) gives the teacher direct access to the destination geometry, so the teacher learns an approximate source-to-target correspondence rather than a completion model conditioned on a sparse observation. Using the same notation, the teacher objective is Chamfer distance plus a local repulsion term on the teacher-estimated clean endpoint \(\mathcal{P}^{\dagger}_{1}\).
\begin{equation}
\begin{array}{l}
\mathcal{L}_{T}
=\mathrm{E}_{\bar{B},\mathcal{P}^{gt},\mathcal{P}_{0}}
\\[1pt]
\displaystyle\quad\left[
\mathrm{CD}(\mathcal{P}^{\dagger}_{1},\mathcal{P}^{gt})
+\lambda_{\mathrm{rep}}\mathcal{L}_{\mathrm{rep}}
\right].
\end{array}
\label{eq:teacher_loss_impl}
\end{equation}
The local repulsion term is applied to the teacher-estimated clean endpoint.
\begin{equation}
\mathcal{L}_{\mathrm{rep}}
=\frac{1}{N}\sum_{i=1}^{N}
\max\!\left(0,\,
r_{\mathrm{rep}}-
\left\|p_i^{\dagger}-p_{\nu(i)}^{\dagger}\right\|_2
\right).
\label{eq:teacher_repulsion_impl}
\end{equation}
Here \(p_i^{\dagger}\in\mathcal{P}^{\dagger}_{1}\), \(\nu(i)\) is the nearest neighbor index of \(p_i^{\dagger}\) within \(\mathcal{P}^{\dagger}_{1}\setminus\{p_i^{\dagger}\}\), \(r_{\mathrm{rep}}=0.2\,\mathrm{m}\) is the repulsion radius, and \(\lambda_{\mathrm{rep}}=0.5\) is the loss weight. We also monitor a displacement excess diagnostic that compares the predicted displacement magnitude with the nearest target distance, but the teacher objective used for the reported model combines CD with the repulsion regularizer. After training, the teacher is not used at inference. It only provides the endpoint displacement or paired target direction used to train the time-dependent student velocity field.

\subsection{Training Schedule}
\label{subsec:training_schedule}

FPSGen is trained in three stages that match the factorized generation pipeline. We first train the BEV flow for 500 epochs to learn the density, height, and mask prior under the sampled condition tuple \(C_m\). We then train the teacher transport network for 5 epochs, using the ground truth BEV prior \(\bar{B}\) and the complete scene \(\mathcal{P}^{gt}\) to produce source-to-target training pairs. Finally, we train the student point flow for 10 epochs on these teacher-estimated pairs. This staged schedule separates global BEV support learning, correspondence construction, and time-dependent point transport so that each component is optimized with its own objective.

\begin{table}[t]
\centering
\setlength{\tabcolsep}{3.0pt}
\begin{tabular}{lccc}
\toprule
Stage & Epochs & Batch & Optimizer and schedule \\
\midrule
BEV flow & 500 & 8 & AdamW, \(10^{-4}\), warmup+cosine \\
Teacher & 5 & 2 & Adam, \(10^{-3}\), ExpLR \(\gamma=0.8\) \\
Student & 10 & 2 & Adam, \(10^{-4}\), ExpLR \(\gamma=0.8\) \\
\bottomrule
\end{tabular}
\caption{Optimization settings for the three training stages.}
\label{tab:optimization_details}
\end{table}

For the BEV flow, we use AdamW with betas \((0.9,0.999)\), weight decay \(10^{-4}\), and an initial learning rate of \(10^{-4}\). The learning rate is updated every optimization step, with a linear warmup for the first 1000 steps followed by cosine decay. For the teacher and student point networks, we use Adam with betas \((0.9,0.999)\). The teacher uses learning rate \(10^{-3}\) and is trained for 5 epochs. The student uses learning rate \(10^{-4}\), sets \(\mathrm{max\_epoch}=10\), and is trained for 10 epochs. Both point networks use an exponential learning rate scheduler with multiplicative factor \(\gamma=0.8\) applied once per epoch.

\textbf{Hardware and timing.}
All training experiments are conducted on two NVIDIA RTX 3090 GPUs. Testing and metric computation are run on a single RTX 3090 GPU for all methods and ablations. For timing related measurements, we keep the evaluation machine otherwise idle and disable unrelated user programs, so the reported runtime numbers are measured under a consistent single GPU environment.

\section{Evaluation Metrics}
\label{sec:metrics}

This section specifies the metrics used in the completion and generation experiments. The completion metrics follow recent LiDAR scene completion evaluations \cite{lidiff,martyniuk2025lidpm,scorelidar,zhao2026distillationdpo}. The generation metrics follow point cloud distributional evaluation with matched generated and reference sets \cite{achlioptas2018learning,yang2019pointflow}.

\subsection{Completion Metrics}
\label{subsec:completion_metrics}

\textbf{Chamfer Distance.}
Given a predicted complete point cloud \(\mathcal{P}^{\dagger}=\{p_i^{\dagger}\}_{i=1}^{N_{\mathrm{pred}}}\) and a ground-truth complete point cloud \(\mathcal{P}^{gt}=\{p_j^{gt}\}_{j=1}^{N_{\mathrm{gt}}}\), Chamfer Distance (CD) \cite{chamferdistance,achlioptas2018learning} measures their bidirectional nearest-neighbor discrepancy:
\begin{equation}
\begin{aligned}
\mathrm{CD}(\mathcal{P}^{\dagger},\mathcal{P}^{gt})
&=
\frac{1}{N_{\mathrm{pred}}}
\sum_{i=1}^{N_{\mathrm{pred}}}
\min_{1\leq j\leq N_{\mathrm{gt}}}
\left\|p_i^{\dagger}-p_j^{gt}\right\|_2 \\
&\quad+
\frac{1}{N_{\mathrm{gt}}}
\sum_{j=1}^{N_{\mathrm{gt}}}
\min_{1\leq i\leq N_{\mathrm{pred}}}
\left\|p_j^{gt}-p_i^{\dagger}\right\|_2.
\end{aligned}
\label{eq:metric_cd}
\end{equation}
Lower CD indicates better point-level geometric alignment. For completion evaluation, \(\mathcal{P}^{\dagger}\) denotes the completed scene, while \(\mathcal{P}^{gt}\) denotes the aggregated ground-truth complete scene.

\textbf{Density-aware Chamfer Distance.}
CD can be dominated by nearest neighbor proximity and may under-penalize uneven point density. We therefore also report Density-aware Chamfer Distance (DCD) \cite{wu2021densityaware} in the supplementary diagnostics. For two point sets \(X=\{x_i\}_{i=1}^{N_X}\) and \(Y=\{y_j\}_{j=1}^{N_Y}\), let
\begin{equation}
\begin{aligned}
\pi_{XY}(i)
&=
\operatorname*{arg\,min}_{j}
\left\|x_i-y_j\right\|_2^2, \\
\pi_{YX}(j)
&=
\operatorname*{arg\,min}_{i}
\left\|y_j-x_i\right\|_2^2.
\end{aligned}
\label{eq:dcd_nn_assign}
\end{equation}
be nearest neighbor assignments from \(X\) to \(Y\) and from \(Y\) to \(X\), respectively. Let \(d_i^{X\rightarrow Y}=\|x_i-y_{\pi_{XY}(i)}\|_2^2\) and \(d_j^{Y\rightarrow X}=\|y_j-x_{\pi_{YX}(j)}\|_2^2\). Define the assignment counts
\begin{equation}
\begin{aligned}
c_j^Y &= \bigl|\{i \mid \pi_{XY}(i)=j\}\bigr|, \\
c_i^X &= \bigl|\{j \mid \pi_{YX}(j)=i\}\bigr|.
\end{aligned}
\label{eq:dcd_counts}
\end{equation}
With exponent \(\lambda=1\), numerical stabilizer \(\varepsilon_{\mathrm{dcd}}=10^{-6}\), and size factors \(\gamma_{X\rightarrow Y}=N_Y/N_X\) and \(\gamma_{Y\rightarrow X}=N_X/N_Y\), the density weights are
\begin{equation}
\begin{aligned}
w_i^{X\rightarrow Y} &= \frac{\gamma_{X\rightarrow Y}}{(c_{\pi_{XY}(i)}^Y)^\lambda+\varepsilon_{\mathrm{dcd}}}, \\
w_j^{Y\rightarrow X} &= \frac{\gamma_{Y\rightarrow X}}{(c_{\pi_{YX}(j)}^X)^\lambda+\varepsilon_{\mathrm{dcd}}}.
\end{aligned}
\label{eq:dcd_weights}
\end{equation}
The DCD with scale \(\alpha\) is
\begin{equation}
\begin{array}{l}
\mathrm{DCD}_{\alpha}(X,Y)
=\frac{1}{2}\Bigg[
\displaystyle
\frac{1}{N_X}\sum_i
\left(1-e^{-\alpha d_i^{X\rightarrow Y}}w_i^{X\rightarrow Y}\right)\\
\displaystyle\quad+
\frac{1}{N_Y}\sum_j
\left(1-e^{-\alpha d_j^{Y\rightarrow X}}w_j^{Y\rightarrow X}\right)
\Bigg].
\end{array}
\label{eq:dcd}
\end{equation}
Lower DCD indicates better joint agreement in geometry and density. We use \(\alpha=1\) for all reported DCD evaluations, matching the evaluation implementation.

\textbf{3D and BEV Jensen-Shannon Divergence.}
Jensen-Shannon Divergence (JSD) measures the discrepancy between occupancy distributions \cite{1997JSD}. For \(JSD_{3D}\), predicted and ground truth scenes are voxelized in 3D and aggregated into normalized occupancy histograms. For \(JSD_{BEV}\), points are projected to the ground plane before constructing BEV occupancy histograms. For normalized histograms \(p\) and \(q\),
\begin{equation}
\begin{array}{l}
\mathrm{JSD}(p,q)=\frac{1}{2}\mathrm{KL}(p\,\|\,m)+
\frac{1}{2}\mathrm{KL}(q\,\|\,m),\quad
m=\frac{1}{2}(p+q).
\end{array}
\label{eq:metric_jsd}
\end{equation}
Here \(\mathrm{KL}\) denotes Kullback--Leibler divergence. Lower JSD indicates better spatial distribution matching.

\textbf{Voxel IoU.}
Voxel intersection over union (IoU) is commonly used for evaluating occupancy prediction and LiDAR scene completion \cite{lmscnet,lidiff,martyniuk2025lidpm}. We compute voxel IoU at resolutions \(r\in\{0.5,0.2,0.1\}\,\mathrm{m}\). Let \(V_r(\hat{\mathcal{P}})\) and \(V_r(\mathcal{P}^{gt})\) denote the occupied voxel sets obtained by voxelizing the prediction and ground truth at resolution \(r\), respectively. Then,
\begin{equation}
\mathrm{IoU}_{r}=
\frac{|V_r(\hat{\mathcal{P}})\cap V_r(\mathcal{P}^{gt})|}
{|V_r(\hat{\mathcal{P}})\cup V_r(\mathcal{P}^{gt})|}.
\label{eq:metric_iou}
\end{equation}
Following OPUS \cite{NEURIPS2024_d8ca28a3}, we apply 3D morphological closing, implemented as max-pool dilation followed by erosion, to each occupancy grid before computing IoU. Specifically, we use kernel sizes \(\{1,3,5\}\) for voxel sizes \(\{0.5,0.2,0.1\}\,\mathrm{m}\), respectively. Thus, no morphological operation is applied at \(0.5\,\mathrm{m}\). Larger IoU indicates better spatial overlap. Voxel IoU at \(0.5\,\mathrm{m}\) emphasizes scene-level coverage, whereas voxel IoU at \(0.1\,\mathrm{m}\) is more sensitive to fine-grained geometric alignment.

\subsection{Generation Metrics}
\label{subsec:generation_metrics}

For generation evaluation, let \(\mathbb{G}=\{X_i\}_{i=1}^{N_{\mathrm{set}}}\) and \(\mathbb{T}=\{Y_j\}_{j=1}^{N_{\mathrm{set}}}\) denote the generated and reference point-cloud sets, respectively. Each \(X_i\) or \(Y_j\) is a point cloud in \(\mathbb{R}^{3}\). Before computing pairwise distances, all generated and reference scenes are processed using identical sampling and coordinate-normalization procedures to obtain the same fixed point budget \(N_{\mathrm{point}}\), such that \(|X_i|=|Y_j|=N_{\mathrm{point}}\).

We use \(d(\cdot,\cdot)\) to denote a distance between two point clouds, instantiated by Chamfer Distance (CD), Earth Mover's Distance (EMD), or Density-aware Chamfer Distance (DCD). Following the standard point-cloud generation evaluation protocol, we report Coverage (COV) and Minimum Matching Distance (MMD) \cite{achlioptas2018learning}, together with 1-nearest neighbor accuracy (1-NNA) \cite{yang2019pointflow}. All methods are evaluated using the same generated-set size, reference set, point-cloud preprocessing, accelerated CD implementation, and approximate EMD solver.

\textbf{Earth Mover's Distance.}
For two equal-cardinality point clouds \(X=\{x_i\}_{i=1}^{N_{\mathrm{point}}}\) and \(Y=\{y_j\}_{j=1}^{N_{\mathrm{point}}}\), the normalized Earth Mover's Distance is theoretically defined as the minimum average transportation cost over all bijective correspondences \cite{rubner2000earth,achlioptas2018learning}:

\begin{equation}
\mathrm{EMD}(X,Y)
=
\min_{\varphi\in\operatorname{Bij}(X,Y)}
\frac{1}{N_{\mathrm{point}}}
\sum_{x\in X}
\left\|x-\varphi(x)\right\|_{2},
\label{eq:metric_emd}
\end{equation}

where \(\operatorname{Bij}(X,Y)\) denotes the set of all bijections from \(X\) to \(Y\), and \(\|\cdot\|_{2}\) denotes the Euclidean norm.

In practice, we approximate the optimal correspondence using a GPU-accelerated soft matching solver. To improve numerical consistency, the input coordinates are first divided by a fixed normalization range \(R_{\max}\):

\begin{equation}
\widetilde{x}_i=\frac{x_i}{R_{\max}},
\qquad
\widetilde{y}_j=\frac{y_j}{R_{\max}}.
\label{eq:metric_emd_normalization}
\end{equation}

The approximate matching solver produces a nonnegative soft matching matrix \(\widehat{\mathbf{M}}=[\widehat{m}_{ij}]\in\mathbb{R}^{N_{\mathrm{point}}\times N_{\mathrm{point}}}\), where \(\widehat{m}_{ij}\) denotes the matching weight assigned between \(\widetilde{x}_i\) and \(\widetilde{y}_j\). The total matching cost returned by the solver is

\begin{equation}
\operatorname{MatchCost}
\left(\widetilde{X},\widetilde{Y}\right)
=
\sum_{i=1}^{N_{\mathrm{point}}}
\sum_{j=1}^{N_{\mathrm{point}}}
\widehat{m}_{ij}
\left\|
\widetilde{x}_i-\widetilde{y}_j
\right\|_{2}.
\label{eq:metric_match_cost}
\end{equation}

Accordingly, the approximate normalized EMD reported in our experiments is computed as

\begin{align}
\widehat{\mathrm{EMD}}(X,Y)
&=
\frac{R_{\max}}{N_{\mathrm{point}}}
\operatorname{MatchCost}
\left(\widetilde{X},\widetilde{Y}\right)
\nonumber\\
&=
\frac{R_{\max}}{N_{\mathrm{point}}}
\sum_{i=1}^{N_{\mathrm{point}}}
\sum_{j=1}^{N_{\mathrm{point}}}
\widehat{m}_{ij}
\left\|
\widetilde{x}_i-\widetilde{y}_j
\right\|_{2}
\nonumber\\
&=
\frac{1}{N_{\mathrm{point}}}
\sum_{i=1}^{N_{\mathrm{point}}}
\sum_{j=1}^{N_{\mathrm{point}}}
\widehat{m}_{ij}
\left\|
x_i-y_j
\right\|_{2}.
\label{eq:metric_emd_implementation}
\end{align}

We set \(R_{\max}=50\) in all experiments. Multiplication by \(R_{\max}\) restores the matching cost to the original coordinate scale, while division by \(N_{\mathrm{point}}\) yields the average matching cost per point. Lower EMD indicates a smaller average transportation distance between the two point clouds and therefore better globally matched geometry. The same approximate matching solver and solver settings are used for all evaluated methods.

\textbf{Coverage.}
Coverage (COV) measures the fraction of reference point clouds that are selected as the nearest neighbor of at least one generated point cloud \cite{achlioptas2018learning}. For each generated point cloud \(X\in\mathbb{G}\), we define its nearest reference point cloud under distance \(d\) as

\begin{equation}
\operatorname{NN}_{d}(X;\mathbb{T})
\in
\arg\min_{Y\in\mathbb{T}} d(X,Y),
\label{eq:metric_cov_nn}
\end{equation}

where ties, if any, are resolved using a fixed deterministic rule. Coverage is then computed as

\begin{equation}
\mathrm{COV}_{d}(\mathbb{G},\mathbb{T})
=
\frac{
\left|
\left\{
\operatorname{NN}_{d}(X;\mathbb{T})
\,:\,
X\in\mathbb{G}
\right\}
\right|
}{
|\mathbb{T}|
}.
\label{eq:metric_cov}
\end{equation}

The set operation in the numerator removes repeated selections, so a reference point cloud selected by multiple generated samples is counted only once. Higher COV indicates that the generated set covers a larger fraction of the reference set and generally reflects greater generation diversity. However, COV alone does not measure the geometric quality of the corresponding nearest-neighbor matches. When reported as a percentage, the value in Eq.~\eqref{eq:metric_cov} is multiplied by \(100\).

\textbf{Minimum Matching Distance.}
Minimum Matching Distance (MMD) measures how closely the reference point clouds are approximated by the generated set \cite{achlioptas2018learning}. For each reference point cloud, it finds the nearest generated point cloud and then averages the corresponding distances:

\begin{equation}
\mathrm{MMD}_{d}(\mathbb{G},\mathbb{T})
=
\frac{1}{|\mathbb{T}|}
\sum_{Y\in\mathbb{T}}
\min_{X\in\mathbb{G}}
d(Y,X).
\label{eq:metric_mmd}
\end{equation}

Lower MMD indicates that samples in the reference set are more closely approximated by samples in the generated set. Since the matching is performed from the reference set to the generated set, generated samples that are not selected as the nearest neighbor of any reference sample are not directly penalized by MMD. Here, MMD refers to Minimum Matching Distance rather than Maximum Mean Discrepancy.

\textbf{1-Nearest Neighbor Accuracy.}
The 1-nearest neighbor accuracy (1-NNA) is a classifier-based two-sample test commonly used for point-cloud generation evaluation \cite{yang2019pointflow}. It measures whether generated and reference point clouds can be distinguished according to their local neighborhoods.

Let \(\mathbb{S}=\mathbb{G}\sqcup\mathbb{T}\) denote the labeled disjoint union of the generated and reference sets, where \(\sqcup\) preserves the source label of each sample. We define the binary source label as
\begin{equation}
y(Z)=
\begin{cases}
0, & Z\in\mathbb{G},\\
1, & Z\in\mathbb{T}.
\end{cases}
\label{eq:metric_1nna_label}
\end{equation}

For each point cloud \(Z\in\mathbb{S}\), its leave-one-out nearest neighbor is defined as
\begin{equation}
\operatorname{NN}_{d}(Z):=\underset{Z'\in\mathbb{S}\setminus\{Z\}}{\arg\min}\, d(Z,Z').
\label{eq:metric_1nna_nn}
\end{equation}

The predicted label of \(Z\) is given by the source label of its nearest neighbor. The 1-NNA score is then computed as
\begin{equation}
\mathrm{1\mbox{-}NNA}_{d}(\mathbb{G},\mathbb{T})
=
\frac{1}{|\mathbb{G}|+|\mathbb{T}|}
\sum_{Z\in\mathbb{S}}
\mathbb{I}\!\left[y(Z)=y\!\left(\operatorname{NN}_{d}(Z)\right)\right].
\label{eq:metric_1nna}
\end{equation}
Here, \(\mathbb{I}[\cdot]\) denotes the indicator function, which equals \(1\) when the condition inside the brackets is true and \(0\) otherwise.

Because the generated and reference sets are balanced, \(|\mathbb{G}|=|\mathbb{T}|=N_{\mathrm{set}}\), a 1-NNA score close to \(50\%\) indicates that generated and reference samples are difficult to distinguish. In contrast, a score substantially above \(50\%\), approaching \(100\%\), indicates a larger distributional discrepancy. Therefore, a score closer to \(50\%\) is preferred.

\section{Additional Qualitative Comparisons}
\label{sec:qualitative}

This section collects additional qualitative comparisons under matched camera, range, point size, and color settings. The visualizations complement CD \cite{achlioptas2018learning}, DCD \cite{wu2021densityaware}, JSD \cite{1997JSD}, and IoU \cite{lmscnet,lidiff} by exposing the geometric alignment, point density, and scene support patterns that contribute to these aggregate metrics.

\subsection{Flow Matching Trajectory Visualization}
\label{subsec:flow_matching_trajectory_supp}

\begin{figure*}[t]
\centering
\IfFileExists{fig/flowshow.png}{%
\includegraphics[width=\linewidth]{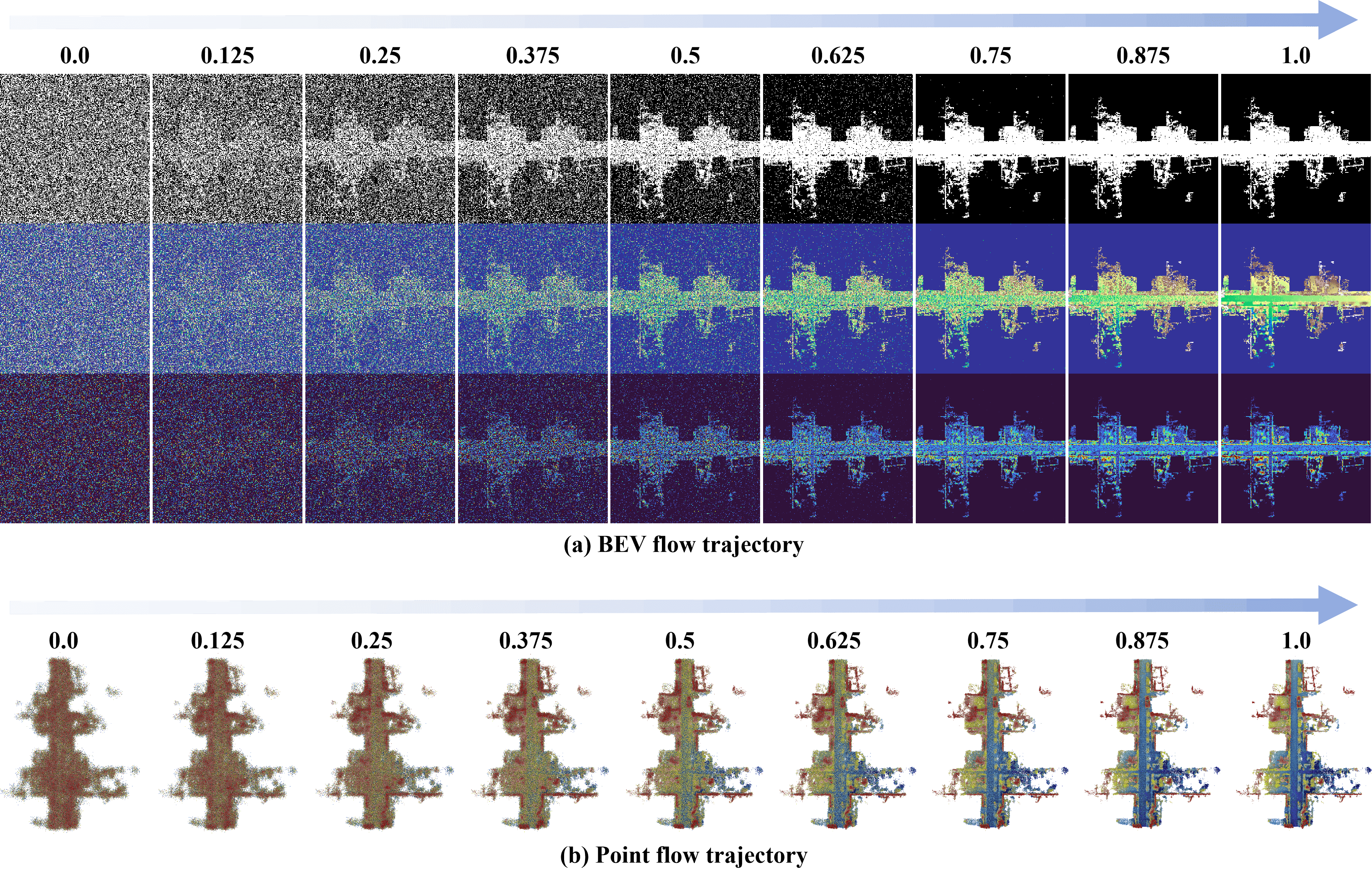}%
}{%
\fbox{\begin{minipage}[c][0.56\textheight][c]{0.96\linewidth}
\centering
\textbf{(a) BEV Flow Trajectory}\\[4em]
\textbf{(b) Point Flow Trajectory}
\end{minipage}}%
}
\caption{Flow matching trajectory visualization. Panel (a), BEV Flow Trajectory, illustrates transport from \(B_{\tau=0}\) to the generated BEV prior \(\hat{B}\). Panel (b), Point Flow Trajectory, illustrates the subsequent transport from \(\mathcal{P}_{0}^{\mathrm{init}}\) to the generated point cloud \(\hat{\mathcal{P}}\).}
\label{fig:flow_matching_trajectory_supp}
\end{figure*}

Figure~\ref{fig:flow_matching_trajectory_supp} provides a qualitative demonstration of the two flow matching stages \cite{lipman2022flowmatching,liu2022rectifiedflow} in FPSGen. The upper panel shows the BEV flow trajectory from the noisy state \(B_{\tau=0}\) to the generated BEV prior \(\hat{B}\). The lower panel shows the point flow trajectory from the inference source \(\mathcal{P}_{0}^{\mathrm{init}}=\mathcal{R}(\hat{B};N,\Sigma)\) to the generated point cloud \(\hat{\mathcal{P}}\). This visualization complements the architecture and quantitative ablations by showing that FPSGen first builds scene-level support in BEV space and then forms detailed geometry through point-level transport.

\subsection{Flexible-Condition FPSGen Visualizations}
\label{subsec:flexible_condition_visualization_supp}

FPSGen supports all eight combinations of LiDAR, vehicle, and road conditions through the active condition tuple \(C_m=(m_lc_l,m_vc_v,m_rc_r)\). We denote a condition code by the binary triplet \(m_lm_vm_r\), where the first digit indicates whether the LiDAR condition is active, the second digit indicates the vehicle mask, and the third digit indicates the road mask. Table~\ref{tab:flex_condition_codes_supp} summarizes the resulting condition settings.

\begin{table}[t]
\centering
\setlength{\tabcolsep}{4pt}
\begin{tabular}{ccccc}
\toprule
Code & LiDAR & Vehicle & Road & Setting \\
\midrule
000 & -- & -- & -- & Unconditional \\
001 & -- & -- & Yes & Road only \\
010 & -- & Yes & -- & Vehicle only \\
011 & -- & Yes & Yes & Layout only \\
100 & Yes & -- & -- & LiDAR only \\
101 & Yes & -- & Yes & LiDAR + road \\
110 & Yes & Yes & -- & LiDAR + vehicle \\
111 & Yes & Yes & Yes & All conditions \\
\bottomrule
\end{tabular}
\caption{Condition code convention for flexible FPSGen visualization. The three digits in \(m_lm_vm_r\) correspond to LiDAR, vehicle, and road conditions.}
\label{tab:flex_condition_codes_supp}
\end{table}

\begin{figure*}[t]
\centering
\includegraphics[width=\linewidth]{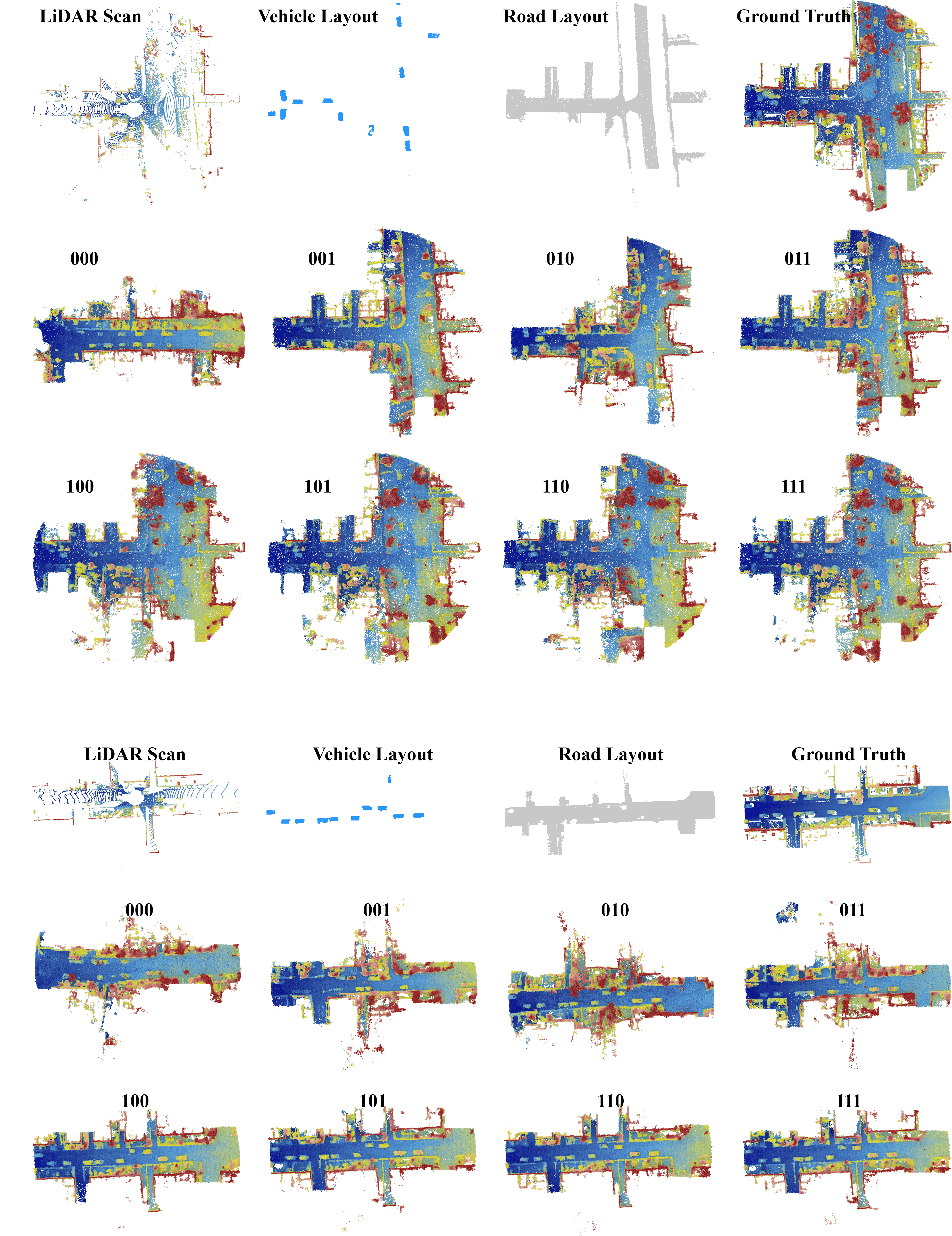}
\caption{Flexible condition FPSGen visualization on KITTI-360 \cite{kitti360}, part I. Each panel corresponds to one condition code \(m_lm_vm_r\), where the digits denote LiDAR, vehicle, and road conditions. The comparison shows how weak layout cues guide scene-level generation even without a LiDAR input.}
\label{fig:qual_condition_grid_supp_a}
\end{figure*}

\begin{figure*}[t]
\centering
\includegraphics[width=\linewidth]{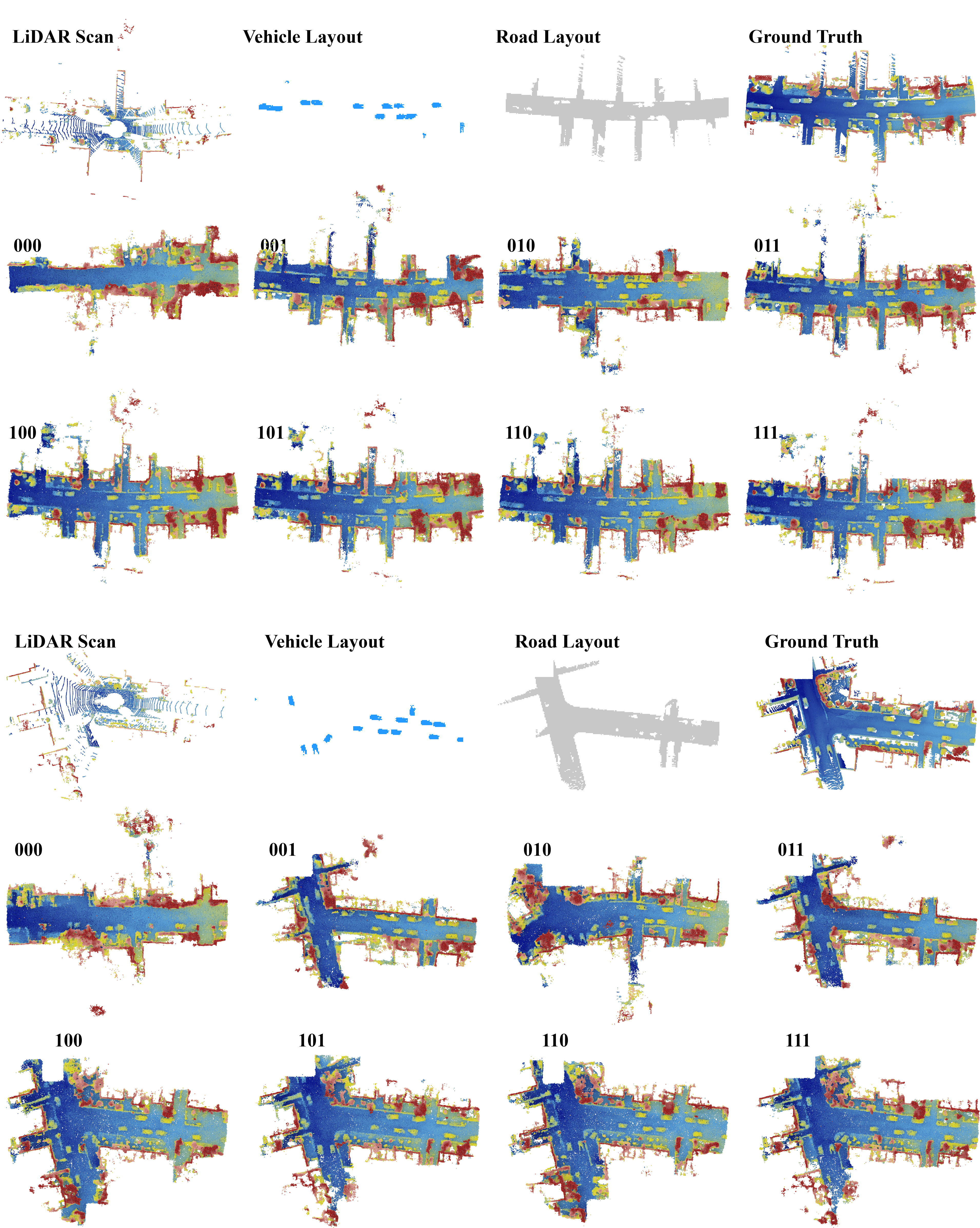}
\caption{Flexible condition FPSGen visualization on KITTI-360 \cite{kitti360}, part II. Adding LiDAR changes the task from weakly conditioned generation toward completion, while road and vehicle masks provide complementary layout control over drivable regions and object placement.}
\label{fig:qual_condition_grid_supp_b}
\end{figure*}

Figures~\ref{fig:qual_condition_grid_supp_a} and~\ref{fig:qual_condition_grid_supp_b} show representative flexible condition generation results on KITTI-360 \cite{kitti360}. The LiDAR-free cases \(000\), \(001\), \(010\), and \(011\) illustrate flexible generation from weak or absent conditions. The LiDAR-active cases \(100\), \(101\), \(110\), and \(111\) show how the same model transitions back to completion-like behavior when a strong geometric condition is available. Road masks mainly constrain global drivable support, vehicle masks localize object regions, and LiDAR anchors observed geometry.

\subsection{A Toy Example of Layout Control}
\label{subsec:text_layout_visualization_supp}

\begin{figure*}[t]
\centering
\includegraphics[width=\linewidth]{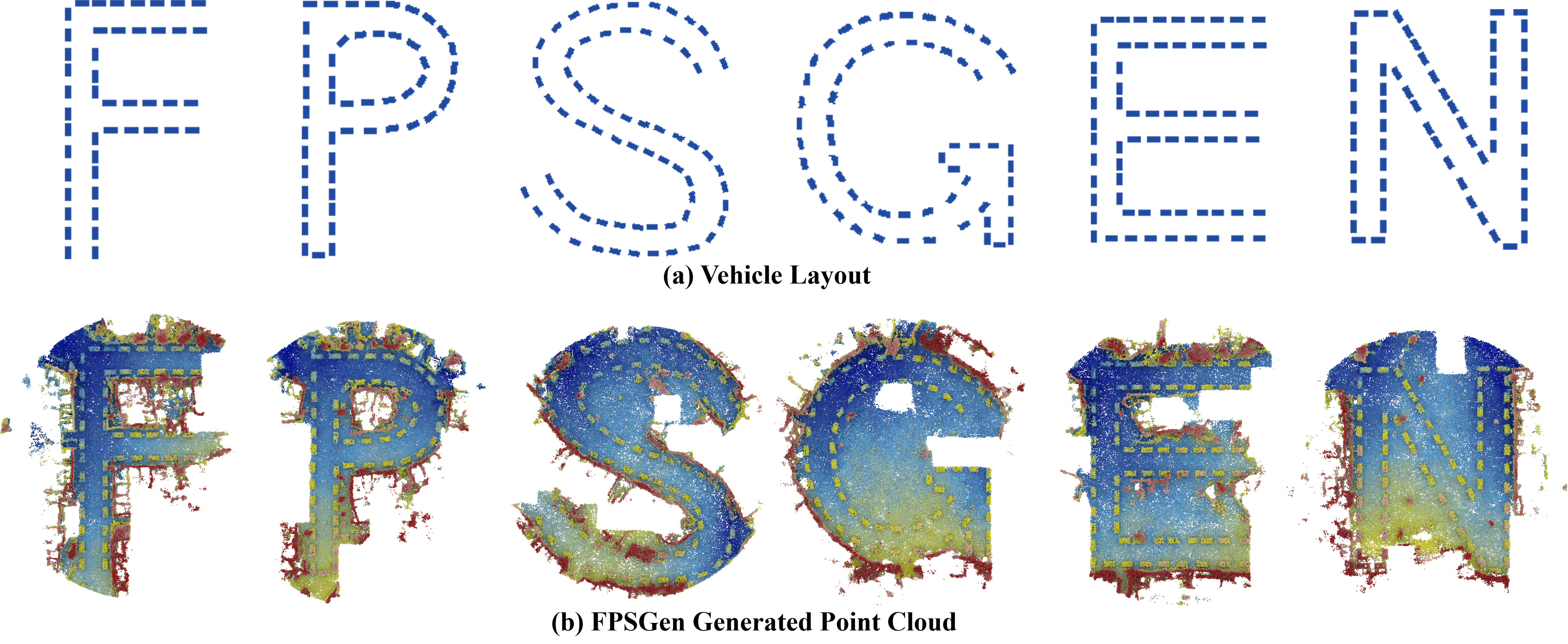}
\caption{A toy example of vehicle layout control. The upper row shows (a) FPSGen Vehicle Layout, where the vehicle-only condition \(010\) forms the six letters ``FPSGEN.'' The lower row shows (b) FPSGen Generated Point Cloud. The generated scenes preserve the prescribed letter-level support while synthesizing three dimensional road surfaces, objects, and vertical geometry without LiDAR or road cues.}
\label{fig:text_layout_visualization_supp}
\end{figure*}

We construct an artificial vehicle layout in which occupied regions form the six letters ``FPSGEN.'' This example uses the vehicle only condition code \(010\), equivalently \(C_m=(0,c_v,0)\), without LiDAR or road cues. Unlike the dataset-derived layouts above, this text-shaped mask deliberately departs from typical road scene configurations and therefore serves as a qualitative stress test of spatial controllability.

Figure~\ref{fig:text_layout_visualization_supp} shows that the generated point clouds retain recognizable letter silhouettes, including curved boundaries, diagonal strokes. At the same time, the outputs are not flat copies of the binary masks. FPSGen expands the two-dimensional layouts into three-dimensional scenes containing road-like surfaces, object clusters, and vertical structures. The result indicates that the vehicle cue can influence global scene support rather than only local object placement, while the learned BEV and point priors supply plausible geometry around the imposed arrangement. This toy example is intended as a controllability demonstration rather than a text generation benchmark or evidence of exact mask reconstruction.

\subsection{Completion Visualizations}
\label{subsec:completion_visualization_supp}

\begin{figure*}[t]
\centering
\includegraphics[width=\linewidth]{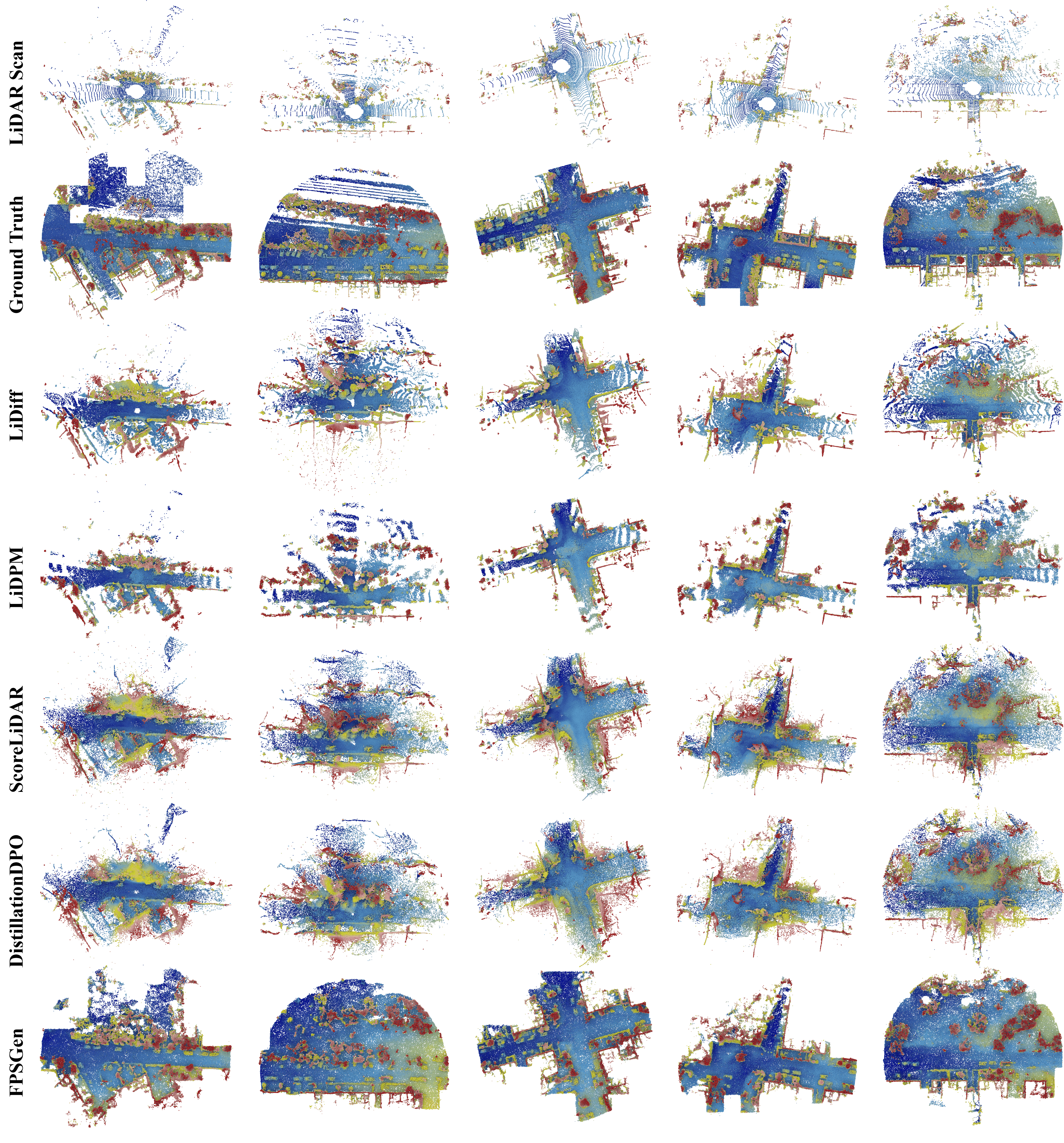}
\caption{Additional LiDAR-conditioned completion visualization on SemanticKITTI \cite{semantickitti} sequence 08. The comparison uses the same viewpoint and rendering setup as the SemanticKITTI completion visualization in the paper and further illustrates how different source constructions affect occluded scene recovery and global point support.}
\label{fig:qual_completion_supp}
\end{figure*}

Figure~\ref{fig:qual_completion_supp} compares the input LiDAR scan, the ground truth, and available completion outputs from LiDiff \cite{lidiff}, LiDPM \cite{martyniuk2025lidpm}, ScoreLiDAR \cite{scorelidar}, Distillation-DPO \cite{zhao2026distillationdpo}, and FPSGen. The matched renderings expose whether each method recovers distant occluded regions, preserves road-level support, and avoids concentrating completed points around the observed scan.

\subsection{LiDM-Seeded Visualizations}
\label{subsec:lidm_seeded_visualization_supp}

\begin{figure*}[t]
\centering
\includegraphics[width=\linewidth]{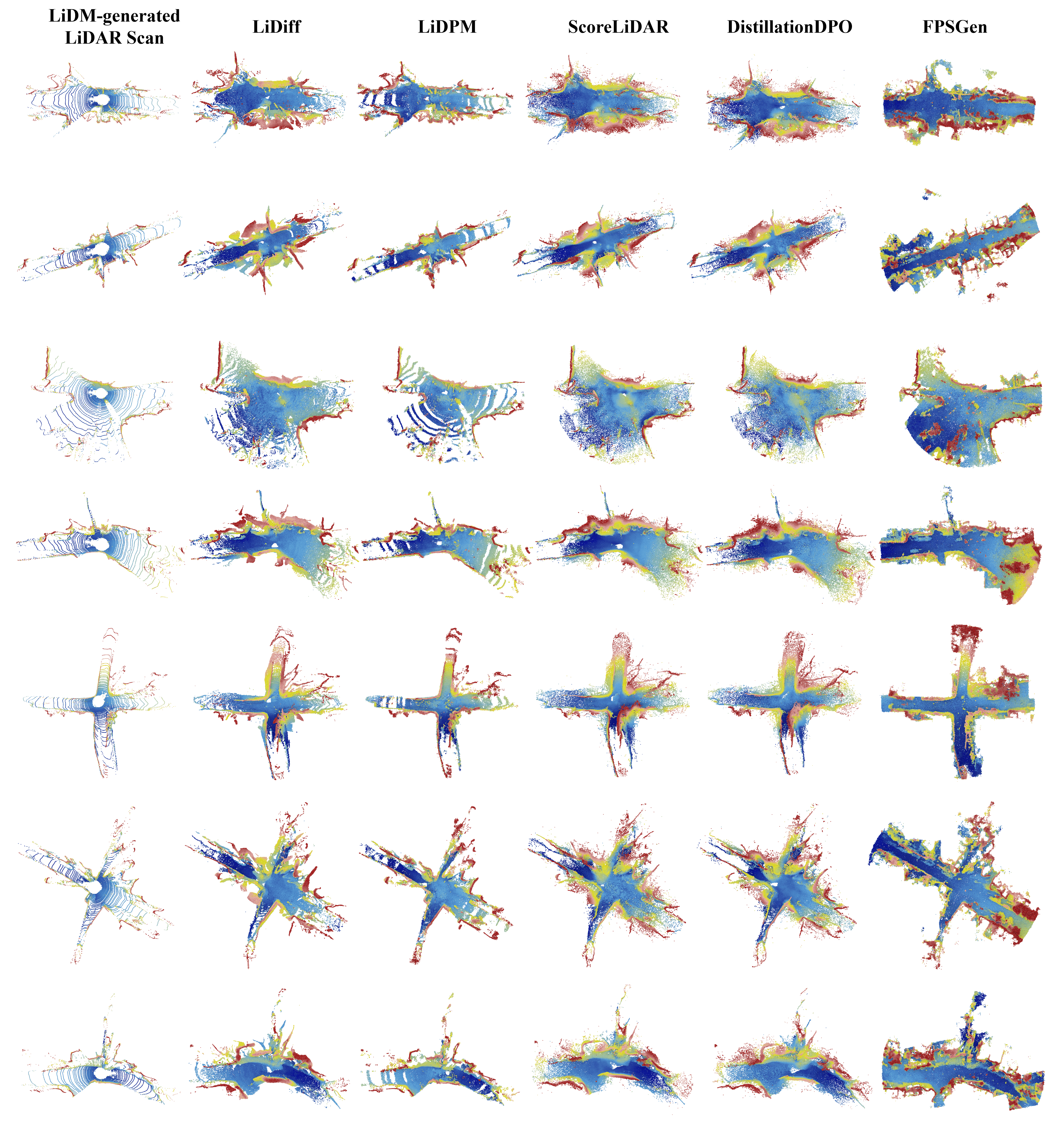}
\caption{LiDM \cite{Lidardiffusioncvpr2024} seeded generation pipeline visualization. LiDM provides the sparse generated scan, while different completion pipelines expand the same sparse input into complete point clouds. The comparison highlights whether the completion stage preserves global scene support instead of only densifying the local observed structure.}
\label{fig:qual_lidm_seeded_supp}
\end{figure*}

We further visualize the LiDM seeded generation diagnostic in Figure~\ref{fig:qual_lidm_seeded_supp}. In this protocol, LiDM \cite{Lidardiffusioncvpr2024} first generates a sparse LiDAR scan, and each completion pipeline then converts this sparse scan into a complete scene. We use LiDM as an established scene level LiDAR generation baseline whose sparse output provides a common stress test for completion-based generation pipelines. The visual comparison separates the LiDM generated input from the ability of each completion pipeline to expand that input into a complete point cloud. The comparison includes available outputs from completion baselines and FPSGen under the same rendering setup.

\section{Limitations and Future Work}
\label{sec:limitations}

\textbf{Temporal consistency and condition scope.}
FPSGen improves flexible-condition LiDAR scene generation by combining BEV-level support generation with point-level transport, but several limitations remain. Our current evaluation focuses on static scene geometry. Dynamic temporal consistency across generated sequences is not explicitly modeled, so extending the framework from independent frames to temporally coherent driving clips is an important direction. Although FPSGen supports multiple condition types, the current experiments mainly use LiDAR, vehicle masks, and road masks. Richer map elements, object-level controls, text prompts, and multi-agent traffic constraints would make the generator more useful for simulation and data augmentation.

\textbf{BEV anisotropy.}
The BEV representation is especially well matched to autonomous driving scenes, where the spatial extent is dominated by the horizontal ground plane and the vertical range is comparatively limited \cite{PointPillars}. This anisotropic structure makes BEV support generation efficient and provides a useful inductive bias for road layouts, object placement, and LiDAR point allocation. However, the same factorization can become restrictive in scenes with richer vertical structure, such as indoor environments, multi-level buildings, or other settings where the \(z\) axis variation is comparable to the horizontal extent. Extending FPSGen to such domains may require volumetric, multiplane, or adaptive 3D support representations rather than a single BEV prior with density, height, and mask channels.

\textbf{Approx OT characterization.}
Our Approx OT construction currently uses a teacher network as a practical amortized surrogate for large-scale scene matching, in contrast to explicit or large-pool OT approximations for point cloud flow models \cite{hui2025notsooptimal}. This avoids explicit OT over hundreds of thousands of points, but the teacher-estimated coupling is still only an approximation and its optimality is not deeply characterized. A more systematic study of teacher matching quality, error propagation from teacher-estimated clean endpoints to student flow fields, and stronger correspondence objectives could further clarify when the learned coupling is reliable.

\textbf{Point flow scalability.}
The point flow stage still relies on sparse convolutional processing \cite{choy20194d} and nearest neighbor matching between BEV and sparse coordinates. This is effective for scene-scale point clouds, but it can be computationally expensive when scaling to denser targets or longer-range sensors. Future work could explore hierarchical tokenization, adaptive point budgets, or streaming generation to reduce sampling cost.

\textbf{Downstream utility.}
While DCD \cite{wu2021densityaware} and occupancy-based metrics reveal important density and support differences, they do not fully measure downstream utility. A natural next step is to evaluate generated scenes by training or stress testing perception models, where controllable rare case generation may be more important than marginal improvements in geometric distances.

\bibliography{aaai2027}


\end{document}